\pdfoutput=1
\documentclass{article}

\newif\ificml
\icmlfalse

\usepackage{lib/arxiv}
\usepackage{amsmath}

\usepackage[utf8]{inputenc} %
\usepackage[T1]{fontenc}    %
\usepackage{url}            %
\usepackage{booktabs}
\usepackage{amsfonts}       %
\usepackage{nicefrac}       %
\usepackage{microtype}      %
\usepackage{lipsum}         %
\usepackage{graphicx}
\usepackage[comma,authoryear,round]{natbib}
\usepackage{doi}

\usepackage{amssymb}
\usepackage{ dsfont }
\usepackage{float}
\usepackage{multicol}
\usepackage{float}
\usepackage{titlesec}

\usepackage{xspace}
\usepackage{pmboxdraw}
\usepackage{adjustbox}
\usepackage[capitalize, noabbrev]{cleveref}
\usepackage{cuted}
\usepackage[most]{tcolorbox}
\usepackage[shortlabels,inline]{enumitem}
\usepackage{pythonhighlight} 

\usepackage{subcaption}

\usepackage{booktabs}

\usepackage{graphicx}
\usepackage{subcaption} 

\usepackage{tikz} 
\usetikzlibrary{bayesnet} 
\usetikzlibrary{arrows} 
\usepackage{wrapfig}
\usetikzlibrary{trees} 

\usepackage{multirow} 
\usepackage{booktabs} 

\usepackage{xspace} 

\usepackage{amsmath}
\usepackage{enumitem}

\usepackage{hyperref}

\usepackage{fontawesome5}
\usepackage{hyperref}

\makeatletter
\newcommand{\github}[1]{%
   \href{#1}{\faGithubSquare}%
}
\makeatother

\DeclareMathOperator*{\argmax}{arg\,max}

\newcommand{\numds}{five}
\newcommand{\numobjs}{five}

\newcommand{\llike}{$\log$-likelihood\xspace}
\newcommand{\lprior}{$\log$-prior\xspace}
\newcommand{\lpost}{$\log$-posterior\xspace}

\newcommand{\lprobs}{$\log$-probabilities\xspace}
\newcommand{\trprune}{$\texttt{Train}$-prune}
\newcommand{\genprune}{$\texttt{Gen}$-prune}
\newcommand{\spprune}{$\texttt{Train} \backslash \texttt{Gen}$-prune}
\newcommand{\cfg}{\ensuremath{\texttt{CFG}}}
\newcommand{\reg}{\ensuremath{\texttt{Reg}}}
\newcommand{\flt}{\ensuremath{\texttt{Flat}}}
\newcommand{\onest}{\ensuremath{\texttt{One-State}}}

\newcommand{\cfgd}{\ensuremath{\mathcal{D}_{\texttt{CFG}}}}
\newcommand{\regd}{\ensuremath{\mathcal{D}_{\texttt{Reg}}}}
\newcommand{\dtrain}{\ensuremath{\mathcal{D}_{\mathrm{train}}}}
\newcommand{\cfgs}{\ensuremath{\texttt{CFG-S}}}
\newcommand{\regs}{\ensuremath{\texttt{Reg-S}}}
\newcommand{\cfgl}{\ensuremath{\texttt{CFG-L}}}
\newcommand{\regl}{\ensuremath{\texttt{Reg-L}}}

\newcommand{\cfgds}{\ensuremath{\mathcal{D}_{\texttt{CFG-S}}}}
\newcommand{\regds}{\ensuremath{\mathcal{D}_{\texttt{Reg-S}}}}
\newcommand{\cfgdl}{\ensuremath{\mathcal{D}_{\texttt{CFG-L}}}}
\newcommand{\regdl}{\ensuremath{\mathcal{D}_{\texttt{Reg-L}}}}
\newcommand{\dtrains}{\ensuremath{\mathcal{D}_{\mathrm{train-S}}}}
\newcommand{\dtrainl}{\ensuremath{\mathcal{D}_{\mathrm{train-L}}}}

\newcommand{\dtestcfgs}{\ensuremath{\mathcal{D}_{\mathrm{test-S}}^{\texttt{Hier}}}}
\newcommand{\dtestcfgl}{\ensuremath{\mathcal{D}_{\mathrm{test-L}}^{\texttt{Hier}}}}

\newcommand{\dtestcfg}{\ensuremath{\mathcal{D}_{\mathrm{test}}^{\texttt{Hier}}}}
\newcommand{\dtestreg}{\ensuremath{\mathcal{D}_{\mathrm{test}}^{\texttt{Lin}}}}

\newcommand{\dtestregs}{\ensuremath{\mathcal{D}_{\mathrm{test-S}}^{\texttt{Lin}}}}
\newcommand{\dtestregl}{\ensuremath{\mathcal{D}_{\mathrm{test-L}}^{\texttt{Lin}}}}

\definecolor{darkblue}{rgb}{0, 0, 0.5}
\hypersetup{colorlinks=true, citecolor=darkblue, linkcolor=darkblue, urlcolor=darkblue}

\newcommand{\cbbbold}[1]{\textcolor{cbblue}{\textbf{#1}}}
\newcommand{\cbrbold}[1]{\textcolor{cbred}{\textbf{#1}}}
\definecolor{unmellowyellow}{rgb}{1.0, 1.0, 0.4}
\definecolor{pastelyellow}{HTML}{FFFFAD}

\hypersetup{
    colorlinks,
    linkcolor={red!50!black},
    citecolor={blue!80!black},
    urlcolor={blue!80!black}
}

\definecolor{brandblue}{rgb}{0.34, 0.7, 1}

\tcbset {
  base/.style={
    arc=0.0mm, 
    bottomtitle=0mm,
    boxrule=0mm,
    colbacktitle=black!10!white, %
    coltitle=black, 
    colback=white,
    left=2.5mm,
    leftrule=0.4mm,
    bottomrule=0.0mm,
    right=3.5mm,
    title={#1},
    toptitle=0.75mm, 
  }
}

\newtcolorbox{mainbox}[1]{
  colframe=tfcolor, %
  base={#1}
}

\usepackage{algorithm}
\usepackage{algorithmic}

\date{}

\usepackage{authblk}

\author{{Kabir Ahuja}\textsuperscript{1} \hspace{8mm} {Vidhisha Balachandran}\textsuperscript{2} \hspace{8mm}  {Madhur Panwar}\textsuperscript{3}   \hspace{8mm} {Tianxing He}\textsuperscript{1}  \hspace{8mm} {Noah A. Smith}\textsuperscript{1,4}  \hspace{8mm} {Navin Goyal}\textsuperscript{3}  \hspace{8mm} {Yulia Tsvetkov}\textsuperscript{1} \\
\vspace{0.1em}
\textsuperscript{1} University of Washington \hspace{8mm}
\textsuperscript{2} Carnegie Mellon University \hspace{8mm}
\textsuperscript{3} Microsoft Research\\
\textsuperscript{4} Allen Institute for AI\\
\vspace{0.1em}
\texttt{kahuja@cs.washington.edu}
}

\renewcommand{\headeright}{}
\renewcommand{\undertitle}{}
\title{Learning Syntax Without Planting Trees: Understanding Hierarchical Generalization in Transformers}
\renewcommand{\shorttitle}{Learning Syntax Without Planting Trees: Understanding Hierarchical Generalization in Transformers}
\hypersetup{
pdftitle={Learning Syntax Without Planting Trees},
pdfsubject={machine learning, nlp, deep neural networks},
pdfauthor={Kabir Ahuja, Vidhisha Balachandran, Madhur Panwar, Tianxing He, Noah A. Smith, Navin Goyal, Yulia Tsvetkov },
pdfkeywords={hierarchical generalization, language modeling, transformers},
}

\newif\ifcomments
\commentstrue
\commentsfalse

\ifcomments
    \newcommand{\kabir}[1]{\textcolor{red}{#1 --Kabir}}
    \newcommand{\vidhisha}[1]{\textcolor{blue}{#1 --Vidhisha}}
    \newcommand{\yulia}[1]{\textcolor{teal}{#1 --Yulia}}
    \newcommand{\navin}[1]{\textcolor{brown}{#1 --Navin}}
    \newcommand{\madhur}[1]{\textcolor{green}{#1 --Madhur}}
    \newcommand{\nascomment}[1]{\textcolor{violet}{#1 --Noah}}
    \newcommand{\tianxing}[1]{\textcolor{olive}{#1 --Tianxing}}
\else
    \newcommand{\kabir}[1]{}
        \newcommand{\vidhisha}[1]{\textcolor{blue}{}}
    \newcommand{\yulia}[1]{\textcolor{teal}{}}
    \newcommand{\navin}[1]{\textcolor{brown}{}}
    \newcommand{\madhur}[1]{\textcolor{green}{}}
    \newcommand{\nascomment}[1]{\textcolor{violet}{}}
    \newcommand{\tianxing}[1]{\textcolor{olive}{}}
\fi

\definecolor{orange}{HTML}{fc8d62}
\definecolor{mauve}{HTML}{8da0cb}
\definecolor{seagreen}{HTML}{66c2a5}
\definecolor{cbred}{HTML}{D7191C}
\definecolor{cbblue}{HTML}{2C7BB6}

\begin{document}
\maketitle

\begin{abstract}
Transformers trained on natural language data have been shown to exhibit hierarchical generalization without explicitly encoding any structural bias. In this work, we investigate sources of inductive bias in transformer models and their training that could cause such preference for hierarchical generalization. We extensively experiment with transformers trained on five synthetic, controlled datasets using several training objectives and show that, while objectives such as sequence-to-sequence modeling, classification, etc., often fail to lead to hierarchical generalization, the language modeling objective consistently leads to transformers generalizing hierarchically. We then study how different generalization behaviors emerge during the training by conducting pruning experiments that reveal the joint existence of subnetworks within the model implementing different generalizations. Finally, we take a Bayesian perspective to understand transformers' preference for hierarchical generalization: we establish a correlation between whether transformers generalize hierarchically on a dataset and if the simplest explanation of that dataset is provided by a hierarchical grammar compared to regular grammars exhibiting linear generalization. Overall, our work presents new insights on the origins of hierarchical generalization in transformers and provides a theoretical framework for studying generalization in language models. 


\hspace{.5em}\includegraphics[width=1.25em,height=1.25em]{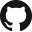}\hspace{.75em}\parbox{\dimexpr\linewidth-2\fboxsep-2\fboxrule}{\url{https://github.com/kabirahuja2431/transformers-hg}}

\end{abstract}
\section{Introduction}
\label{sec:intro}

Natural language is structured hierarchically: words are grouped into phrases or constituents, which can be further grouped to form higher-level phrases up to the full sentence. How well do the neural network models trained on language data learn this phrase structure of human language has been a subject of great interest. A flurry of past work have shown that syntax trees can be recovered from  recurrent neural network (RNN) and transformer-based models trained on large-scale language corpora \citep{tenney2018what,peters-etal-2018-dissecting,lin-etal-2019-open,wu-etal-2020-perturbed}. While these studies provide useful evidence of the aforementioned phenomenon, they do not shed light on the architectural choices, training paradigms or dataset characteristics that lead  models to learn the phrase structure of language. \kabir{I have changed the first paragraph of the intro, as there seemed to be a disconnect before between the generalization connundrum / simplicity bias in the first paragraph and hierarchical generalization in the second.}


A useful tool to understand these model and dataset specific properties is through the test for hierarchical generalization, i.e.,
evaluating the capability of a model to generalize to novel syntactic forms, which were unseen during training. A classic problem to test for hierarchical generalization is \textit{question formation}, where given a declarative sentence, e.g., \textit{My walrus does move the dogs that do wait.}, the task is to transform it into a question: \textit{Does my walrus move the dogs that do wait?} The task is accomplished by moving one auxiliary verb to the front.  The correct choice to move \emph{does} in this example (rather than \emph{do}), is predicted both by a \emph{hierarchical rule} based on the phrase-structure syntax of the sentence, and by a \emph{linear rule} that says to move the \emph{first} auxiliary. 
Hence, as a test for hierarchical generalization, we can ask,  for neural networks trained from scratch on data that is consistent with both hierarchical and linear rules (i.e., ambiguous data), do they learn to generalize hierarchically or do they learn a linear rule (e.g., moving the first auxiliary to the beginning of the sentence)?



This question has been well-studied in  past work for different neural network architectures. In particular, \citet{mccoy-etal-2020-syntax} showed that recurrent neural networks fail to generalize hierarchically when trained on ambiguous data, and only using tree-structured networks \citep{chen-etal-2017-improved, chen-etal-2018-tree}, which use explicit parses as inputs, leads to hierarchical generalization. \cite{petty_transformers_2021} and \cite{mueller-etal-2022-coloring} observed the same for transformers. However, \citet{murty-etal-2023-grokking} showed that, surprisingly, when trained for a long time after attaining perfect training accuracy, transformers  \emph{do} start to generalize hierarchically. They named this phenomenon \textit{Structural Grokking}, because it resembles  ``grokking''
as observed by \cite{Power2022GrokkingGB} (where neural networks start to generalize long after they have overfit the training data).

While the results of \cite{murty-etal-2023-grokking} suggest that transformers are capable of exhibiting hierarchical generalization despite being trained on ambiguous data, it remains unclear why they exhibit such a preference. In our work, we ask, \emph{why do transformers show hierarchical generalization, despite lacking architectural biases towards hierarchical structure?} We first explore if the choice of training objective can influence hierarchical generalization in transformers. Specifically, we consider \numobjs{} objectives in our study --  language modeling, sequence-to-sequence modeling, prefix language modeling, sequence classification, and cloze completion, and compare the hierarchical generalization exhibited by transformers under these objectives. As a test for hierarchical generalization, in addition to the English question formation task described above, we also include German question formation \citep{mueller-etal-2022-coloring}; tense-reinflection \citep{mccoy-etal-2020-syntax}, which converts a sentence in the past tense to the present; passivization \citep{mueller-etal-2022-coloring}, i.e., converting a sentence in active voice to passive; and simple agreement, a synthetic task that we construct to check if the model can predict correct agreement between the verb and subject in a declarative sentence. To better understand how different generalization behaviors are implemented within the trained networks, we propose two new attention head pruning strategies to discover subnetworks corresponding to different generalizations (hierarchical and linear rules).

Finally, to understand why language modeling results in bias towards hierarchical structure, we utilize the Bayesian framework from \citet{PERFORS2011306} and consider generative probabilistic grammars (PCFGs) modeling the simple agreement task. Specifically, we construct hierarchical grammars (consistent with the hierarchical rule) as well as regular grammars that generate the data linearly and hence are consistent with the linear rule. We then compare the posterior probabilities of the two grammars, to understand which grammar has a better trade-off for the goodness of fit (measured using the likelihood) and simplicity (by calculating the prior on grammars), thereby explaining the preference of transformers for hierarchical or linear generalization.

Since our aim is to understand hierarchical generalization in transformers in isolation, following \citealp{mccoy-etal-2020-syntax, murty-etal-2023-grokking}, we train transformer models from scratch, without any pretraining, eliminating the possibility of these models having bias towards hierarchical generalization due to having been trained on language data before \citep{mueller-etal-2022-coloring}. For the same reason, we also use synthetic datasets for training and evaluation that exclusively measure the inductive biases of these models towards hierarchical or linear generalization. 
Due to the controlled nature of our setup, the transformer models that we train are small (e.g. 6 layers and 512 hidden size).

\paragraph{Our contributions:}

\begin{itemize}[leftmargin=*]
    \item We discover that \emph{the choice of the training objective affects hierarchical generalization in transformers}. Among \numobjs{} training objectives and \numds{} datasets, we find that only the language modeling objective consistently obtains strong hierarchical generalization across different tasks. 

    \item We find that \textit{different types of generalizations consistent with the training data} (e.g., hierarchical and linear rules) \textit{can be discovered as subnetworks in the trained model}, and these subnetworks continue to coexist over the course of training, despite the overall model performing closer to one kind of generalization over the other. Further, we find these disparate subnetworks exist due to the ambiguity in the training data, as we find different subnetworks to disappear upon adding disambiguating examples (i.e.,  only consistent with the hierarchical rule).

    \item Finally, utilizing the Bayesian framework from \cite{PERFORS2011306}, we show a correlation between transformer LMs generalizing hierarchically and hierarchical grammars having higher posterior, compared to regular grammars that follow the linear rule. This suggests that transformers generalize hierarchically because the \textit{hierarchical grammars that fit the data are often ``simpler'' compared to regular grammars}. We also identify a case where this does not hold i.e. regular grammars have a higher posterior than hierarchical grammars, and show that transformers fail to generalize hierarchically in this case.
\end{itemize}

To the best of our knowledge, the present work is the first to show that the language modeling objective is a source of inductive bias for hierarchical generalization and to use the Bayesian perspective to explain hierarchical generalization in \nascomment{can we strengthen this and just say ``in language models''?} language models. 
Our work takes steps towards understanding hierarchical generalization in language models, and we hope that our analysis method will be useful to study other forms of generalization in these models.



\section{Background}



\paragraph{Hierarchical generalization.} Hierarchical generalization is a form of systematic generalization, where given  instances generated from a hierarchical grammar, we evaluate the capability of a model to generalize to unseen syntactic forms. For example, consider the task of converting a declarative English sentence to a question:

\begin{enumerate}[leftmargin=*]
    \item \label{ls:simp_pair}
    \begin{enumerate}
        \item \textbf{Input}: My walrus does move . \label{ls:simp_decl}
        \item \textbf{Output}: Does my walrus move ? \label{ls:simp_quest}
    \end{enumerate}
    \item \label{ls:amb_pair}
    \begin{enumerate}
        \item \textbf{Input}: My walrus does move the dogs that do wait . \label{ls:amb_decl}
        \item \textbf{Output}: Does my walrus move the dogs that do wait ? \label{ls:amb_quest}
    \end{enumerate}
\end{enumerate}

Notice that the task can be accomplished by moving one auxiliary verb to the front of the sentence. While for sentences of type \ref{ls:simp_decl}, with only a single auxiliary this is trivial, for sentences of type \ref{ls:amb_decl}, as English speakers we know that the auxiliary to move is the one associated with the head verb in the sentence (i.e., \emph{does}, which is associated with \emph{move}, not \emph{do}, which is associated with \emph{wait}). Modeling this rule requires understanding the phrase structure of the language. We call this \textit{Hierarchical Rule}. One can alternatively consider a much simpler explanation, the \textit{Linear Rule}: moving the first auxiliary in the sentence to the beginning. This linear rule is independent of the hierarchical  structure of the sentence. However, consider sentences of type 3 below:

\begin{enumerate}[leftmargin=*]\addtocounter{enumi}{2}
    \item \label{ls:gen}
    \begin{enumerate}
        \item \textbf{Input}: My walrus who doesn't sing does move . \label{ls:gen_decl}
        \item \textbf{Linear rule output}: \textcolor{cbred}{Doesn't my walrus who sing does move ?} \label{ls:gen_ord_quest}
        \item \textbf{Hierarchical rule output}: \textcolor{cbblue}{Does my walrus who doesn't sing move ?} \label{ls:gen_hier_quest}
    \end{enumerate}
\end{enumerate}
First, notice that sentence \ref{ls:gen_decl} has a different syntactic structure compared to the sentence \ref{ls:amb_decl}, as the relative clause \textit{who doesn't sing} accompanies the subject, unlike in example \ref{ls:amb_pair} where \emph{that do wait} modified the object (\emph{the dogs}). In this case, using the linear rule to form question will result in an ungrammatical sentence, i.e., outputting sentence \ref{ls:gen_ord_quest} instead of sentence \ref{ls:gen_hier_quest}. In this work, we study the following question: Consider neural networks trained from scratch on data consistent with both hierarchical and linear rules (e.g., examples \ref{ls:simp_pair}, \ref{ls:amb_pair}). When presented with sentences such as \ref{ls:gen_decl} do they generalize hierarchically (predicting \ref{ls:gen_hier_quest}) or do they learn a linear rule (predicting \ref{ls:gen_ord_quest})? 

\paragraph{Tasks and datasets.} In our study, we consider \numds{} tasks, including the question formation task above. Examples from all the tasks (excluding English question formation) are provided in Table \ref{tab:dataset_examples}.  All the tasks follow a common recipe: the training dataset has examples that are consistent with both hierarchical and linear rules. For evaluation, two variants of the test data are considered: an in-distribution test set, which follows the same distribution as the training data (i.e., has the same syntactic forms and is also ambiguous with respect to the correct rule); and  a generalization test set, which consists of examples which are only consistent with the hierarchical rule. Below we provide the details of the \numds{} tasks.

\begin{table}[!htbp]
    \centering
    \small
    \caption{Examples from the different tasks we study in our work. \colorbox{pastelyellow}{highlighted} text indicates examples in the generalization set. \nascomment{some examples add whitespace before punctuation, others don't - be consistent.  usggest changing highlighted gray to a lighter gray or something easier to read through like a pale yellow}}
    \resizebox{0.9\textwidth}{!}{
    \begin{tabular}{p{0.12\linewidth}|p{0.8\linewidth}}
         \toprule
         Task & Examples \\
         \midrule
         \multirow{4}{0.10\linewidth}{QF (German)} & unsere Papageien \textcolor{cbblue}{\textbf{können}} meinen Papagei , der gewartet \textcolor{cbred}{\textbf{hat}} , akzeptieren .  \\
         & $\to$ \textcolor{cbblue}{\textbf{können}} unsere Papageien meinen Papagei , der gewartet \textcolor{cbred}{\textbf{hat}} , akzeptieren ? \\
         & \colorbox{pastelyellow}{ihr Molch , der gegessen \textcolor{cbred}{\textbf{hat}} , \textcolor{cbblue}{\textbf{kann}} lächeln .}\\
         & \colorbox{pastelyellow}{$\to$ \textcolor{cbblue}{\textbf{kann}} ihr Molch , der gegessen \textcolor{cbred}{\textbf{hat}} , lächeln ?}\\
         \midrule

         \multirow{4}{0.10\linewidth}{Passivization} & some tyrannosaurus entertained your \textcolor{cbblue}{\textbf{quail}} behind your \textcolor{cbred}{\textbf{newt}} .\\
         & $\to$ your \cbbbold{quail} behind your \cbrbold{newt} was entertained by some tyrannosaurus .\\
         & \colorbox{pastelyellow}{the zebra upon the \cbrbold{yak} confused your \cbbbold{orangutans} .}\\
         &\colorbox{pastelyellow}{ $\to$ your \cbbbold{orangutans} were confused by the zebra upon the \cbrbold{yak} .}\\
        \midrule
         \multirow{4}{0.10\linewidth}{Tense reinflection} & my \cbbbold{zebra} by the \cbrbold{yak} \underline{swam} . \\
         & $\to$ my \cbbbold{zebra} by the \cbrbold{yak} \underline{swims} . \\
         & \colorbox{pastelyellow}{my \cbbbold{zebras} by the \cbrbold{yak} \underline{swam} .}\\
         & \colorbox{pastelyellow}{$\to$ my \cbbbold{zebras} by the \cbrbold{yak} \underline{swim} .}\\
         \midrule
         \multirow{2}{0.10\linewidth}{Simple Agreement}& my \cbbbold{zebra} by the \cbrbold{yak} $\to$ swims\\
         & \colorbox{pastelyellow}{my \cbbbold{zebras} by the \cbrbold{yak} $\to$ swim}\\
         \bottomrule
    \end{tabular}}
    \label{tab:dataset_examples}
\end{table}

\begin{enumerate}[leftmargin=*]
    \item \textbf{Question formation.} As described above, the task is to transform a declarative sentence into a question. We use the dataset from \cite{mccoy-etal-2020-syntax} for this task, which was constructed from a context-free grammar (CFG) with three sentence types varying in the existence and position of the relative clause (RC) in the sentence: (i) no RC, e.g., sentence \ref{ls:simp_decl}; (ii) RC attached to the object, e.g., sentence \ref{ls:amb_decl}; and (iii) RC attached to the subject, e.g., sentence \ref{ls:gen_decl}. The training data includes (a 50-50 split) (i) declarative-question pairs where the task is to take a declarative sentence and generate a question as output and (ii) auxiliary identity pairs where the task requires copying an input declarative sentence. The declarative-question pairs in the training set only contain sentences without any RC or with RC attached to the object.  As such, the training dataset is consistent with both the hierarchical and linear rules and ambiguous in terms of which rule is applicable. Importantly, the auxiliary identity pairs in the training data also include sentences with RC \emph{on the subject}, to expose the model to sentences of this type \citep{mccoy-etal-2020-syntax}. During training a token \texttt{quest} or \texttt{decl} is added to specify whether to perform question formation task or the copy task.\footnote{Notice that the dataset consists of sentences with explicit auxiliary verbs -- \textit{my walrus \underline{does} move}, instead of the less marked \textit{my walrus moves}. This choice was made by \cite{mccoy-etal-2020-syntax}, to study the problem as auxiliary fronting, i.e., moving the auxiliary verb to the beginning, and thereby studying the two rules (hierarchical and linear).} Following \cite{mccoy-etal-2020-syntax} and \cite{murty-etal-2023-grokking}, we evaluate the model on the \textit{first-word accuracy}, i.e., given the declarative sentence as the input, we evaluate whether the model predicts the correct auxiliary for the first word in the question in its generation.

    \item \textbf{Question formation (German).} This is the same task as above, but the sentences are in German instead of English. We use the dataset from \citet{mueller-etal-2022-coloring}, consisting of sentences with the modals \emph{können/kann} (can) or auxiliaries \emph{haben/hat} (have/has), together with infinitival or past participle main verbs as appropriate, which can be moved to the front similar to English to form questions.\footnote{Note that in German, negation is represented using another word \textit{nicht} which is not fronted with the auxiliary (\textit{can't} becomes \textit{kann nicht}), hence \citet{mueller-etal-2022-coloring} do not use the auxiliaries with negation for German, like we have for the English version.} Here again, the linear rule is to move the first modal or auxiliary to the front, and the hierarchical rule requires moving the token associated with the main verb. The dataset construction and evaluation metrics remain identical to the English version. 
 
    \item \textbf{Passivization.} The task here is to transform an active sentence to passive. The dataset from \citet{muller2022transformers} is constructed such that it contains active sentences of three types: (i) without any prepositional phrase (PP), (ii) with a PP on the object, and (iii) with a PP on the subject. Similar to question formation dataset, the active-passive pairs in the training dataset are constructed only using the sentences of type (i) and (ii). These two sentence types are again compatible with both rules: the hierarchical rule which involves identifying the object in the sentence and moving it to the front, and the linear rule that moves the second noun in the sentence to front. Like question formation, the training data is augmented with identity active-active pairs which consist of sentences of all the three types. For evaluation, following \citet{mueller-etal-2022-coloring}, we consider \textit{object noun accuracy}, which measures whether the correct noun was moved to the subject position.

    \item \textbf{Tense reinflection.} In tense reinflection, we are given a sentence in the past tense, and the task is to transform it into present tense. While performing the transformation to present tense, the model has to figure out from the context whether each verb should be singular or plural (\textit{-s} suffix) in the present tense. In this case, the hierarchical rule requires each verb to agree with the hierarchically-determined subject and the linear rule requires a verb to agree with the most recent noun in the sequence. We use the same dataset as \cite{mccoy-etal-2020-syntax}, where, similar to question formation, the training dataset contains tense reinflection pairs (past-present) that are consistent with both rules, and identity pairs (past-past) for copying that include past-form of sentences whose present form can only be generated using the hierarchical rule. The models are evaluated using \textit{main-verb accuracy}, which is calculated as the fraction of examples in the test set for which the generated present tense sentence has the correct main verb.

    \item \textbf{Simple agreement.} We also introduce a simplified version of the tense reinflection task. Unlike other tasks, simple agreement is a single-sentence task where only the \emph{present}-tense sentences from the tense-inflection are used for training. In this task ,we evaluate the model's ability to generate the correct inflection of the verb at the end of the sentence. E.g., when given the prefix\textit{ my \textbf{zebra} by the yaks} as input, does the model assign higher likelihood to the singular verb form \textit{swims} (correct) or the plural form \textit{swim} (incorrect), as the continuation. The hierarchical and linear rules are defined in the same way as tense reinflection. For evaluation, since from the context it is no longer clear what should be the correct verb, we use \textit{main-verb contrastive accuracy}, which is calculated by considering each sentence in the test dataset (e.g., \textit{my \textbf{zebra} by the yaks swims}), forming the prefix (\textit{my \textbf{zebra} by the yaks}) and checking if the model assigns the higher probability to the correct inflection of the main verb in the original sentence (\textit{swims} vs.~\textit{swim}).
    
\end{enumerate}

 For all tasks excluding simple agreement, there are 100k training examples (50k transformation pairs and 50k identity pairs) and 1k and 10k examples in in-distribution and generalization test sets respectively. For simple agreement, we generate 50k training examples (and 1k/10k for test datasets).

\section{How the Training Objective Influences Hierarchical Generalization}



We now discuss how the choice of training objective can influence hierarchical generalization in transformers. Prior work by \cite{mccoy-etal-2020-syntax}, \cite{petty_transformers_2021}, and \cite{mueller-etal-2022-coloring} used a sequence-to-sequence training objective to train encoder-decoder models and found that RNNs and transformers do not exhibit hierarchical generalization. More recently, \cite{murty-etal-2023-grokking} used a language modeling objective to train a decoder-only transformer, which they found \emph{did} generalize hierarchically when trained for a sufficiently large number of epochs -- well beyond the point of achieving perfect training task accuracy. To the best of our knowledge, this distinction isn't called out by  prior work. Hence we conduct a systematic study to understand what effect the training objective has on hierarchical generalization. 


%

\subsection{Training Objectives}
\label{sec:objectives}
We consider the following five training objectives in our study:

\paragraph{Language modeling.} Given a sequence of tokens,
the language modeling objective  trains the model to predict each token in a sequence given the preceding tokens.
The model is optimized to minimize the  negative log-likelihood of the sequences in the training data. For transformers, the language modeling objective is typically associated with decoder-only models like GPT \citep{brown2020language}, and the loss is computed over \emph{all} tokens in the sequence.  For the question formation task and the declarative-question pair from the introduction, if $\boldsymbol{s} = \langle s_1, s_2, \ldots, s_{21}\rangle = \langle$\emph{my, walrus, does, move, the, dogs, that, do, wait, ., \emph{quest}, does, my, walrus, move, the, dogs, that, do, wait, ?}$\rangle$, the cross-entropy loss is computed over $s_1$ through $s_{21}$, each given the preceding tokens:  
\begin{equation}
- \log p(\boldsymbol{s}) =    - \sum_{i=1}^{21} \log p(s_i \mid s_1, \ldots, s_{i-1}). \label{eq:lmo}
\end{equation}


\paragraph{Sequence-to-sequence modeling.} The sequence-to-sequence (seq2seq) modeling objective \citep{sutskever2014sequence}, is used to train the model to generate a target sequence (e.g., from the example above, $\langle s_{12}, \ldots, s_{21}\rangle$) \emph{given} an input sequence ($\langle s_1, \ldots, s_{11}\rangle$).  \nascomment{check that I got the indices right}
This objective, which includes only the terms from $i = 12$ to $21$ in equation~\ref{eq:lmo}, is typically associated with an encoder-decoder model as used in the original transformer architecture \citep{vaswani_attention_2017}. 
Note that the seq2seq objective is more suited for tasks with an explicit input and output (like question formation and tense inflection), but is not suitable for the simple agreement task. Hence, we do not evaluate the seq2seq objective for simple agreement.

\paragraph{Prefix language modeling.} In the prefix language modeling objective \citep{dang-et-al-2019-unified}, we again generate the output text given the input (or ``prefix''), but we use a single transformer decoder (similar to language modeling) instead of an encoder-decoder model. Differing from the original language modeling objective, here the loss is only computed over the output text and does not include the prefix.  One modification that we make to how the prefix-LM objective is typically used, is that we use a causal mask for the prefix tokens as well instead of having bi-directional attention over the prefix tokens, since we found the latter to perform subpar in our initial experiments  (we compare the two in detail in \textsection \ref{sec:obj_results}).

\paragraph{Sequence classification.} In the sequence classification objective, the model is trained to map the entire sequence to a discrete label. E.g., for question formation the model is given the input declarative sentence and trained to predict the correct auxiliary from the set of auxiliary verbs (\emph{do, does, don't, doesn't}) that should occur at the start of the question, i.e., a four-way classification task.

\paragraph{Cloze completion.} In the cloze completion setting, the model is given a sequence of tokens with some tokens masked and trained to predict the masked tokens.
E.g., for the question formation task, we consider the declarative-interrogative pair and mask out tokens in the interrogative sentence at all positions where the auxiliaries could be present. Specifically, we have mask tokens where (i) the auxiliary is present in the interrogative sentence or (ii) the auxiliary was present in the original declarative sentence. The model is trained to  predict the correct auxiliary at these \nascomment{do you mean at ``these'' positions?  ``right'' makes me think of the right side of the sentence} positions and \texttt{$<$EMPTY$>$} if an auxiliary is not present at a particular position.
Note that this objective is similar to masked language modeling as in \cite{devlin-etal-2019-bert}; however, instead of masking tokens randomly, we mask the specific tokens as described above.\footnote{Our initial experiments with random-masking resulted in subpar performance, even on in-distribution test sets.}  For the passivization task, we do not evaluate the cloze completion objective, because (unlike other tasks) the output sequence is significantly different from the input and not just in terms of one or two tokens, which makes defining the masking strategy in this case non-trivial.

Please refer to \textsection \ref{sec:obj_full_details} for full details of each objective for all of the \numds{} tasks.

\subsection{Experimental Setup}
\label{sec:expt_set}

We train transformer models from scratch for all of our experiments. We use transformer models with 8 heads and  embedding dimension 512 for all datasets and objectives. Following \citet{murty-etal-2023-grokking}, for question formation and tense reinflection, we train transformer models with 6 layers for the former (18M parameters) and 4 layers  (12M parameters) for the latter task, for all objectives excluding seq2seq. For the remaining tasks, we use 6-layer transformer encoder/decoder layers (18M parameters) depending on the training objective. For the seq2seq objective, we use a 6-layer encoder/6-layer decoder model (25M parameters) for all tasks.\
For all the tasks, tokenization is performed at the word level.
We use the Adam optimizer \citep{kingma-ba-2015-adam} for training the model with a learning rate of $0.0001$, following \citet{murty-etal-2023-grokking}. We use batch size of 8, and train for 300k steps (24 epochs) for all tasks excluding simple agreement, which we train for 200k steps (32 epochs), since the dataset is half the size of others (recall that we have 100k training examples for all the tasks except simple agreement for which we have 50k). We run each experiment with 5 seeds and report the average performance. 


\paragraph{Baselines.}  By design of the test datasets, a model following  the linear rule will obtain 100\% in-distribution accuracy and  0\% generalization accuracy.  Only a model consistent with the hierarchical rule will obtain 100\% accuracy on both test sets for all the tasks.

\subsection{Results}
\label{sec:obj_results}
\begin{figure}[htbp]
    \centering
    \includegraphics[width=0.9\textwidth]{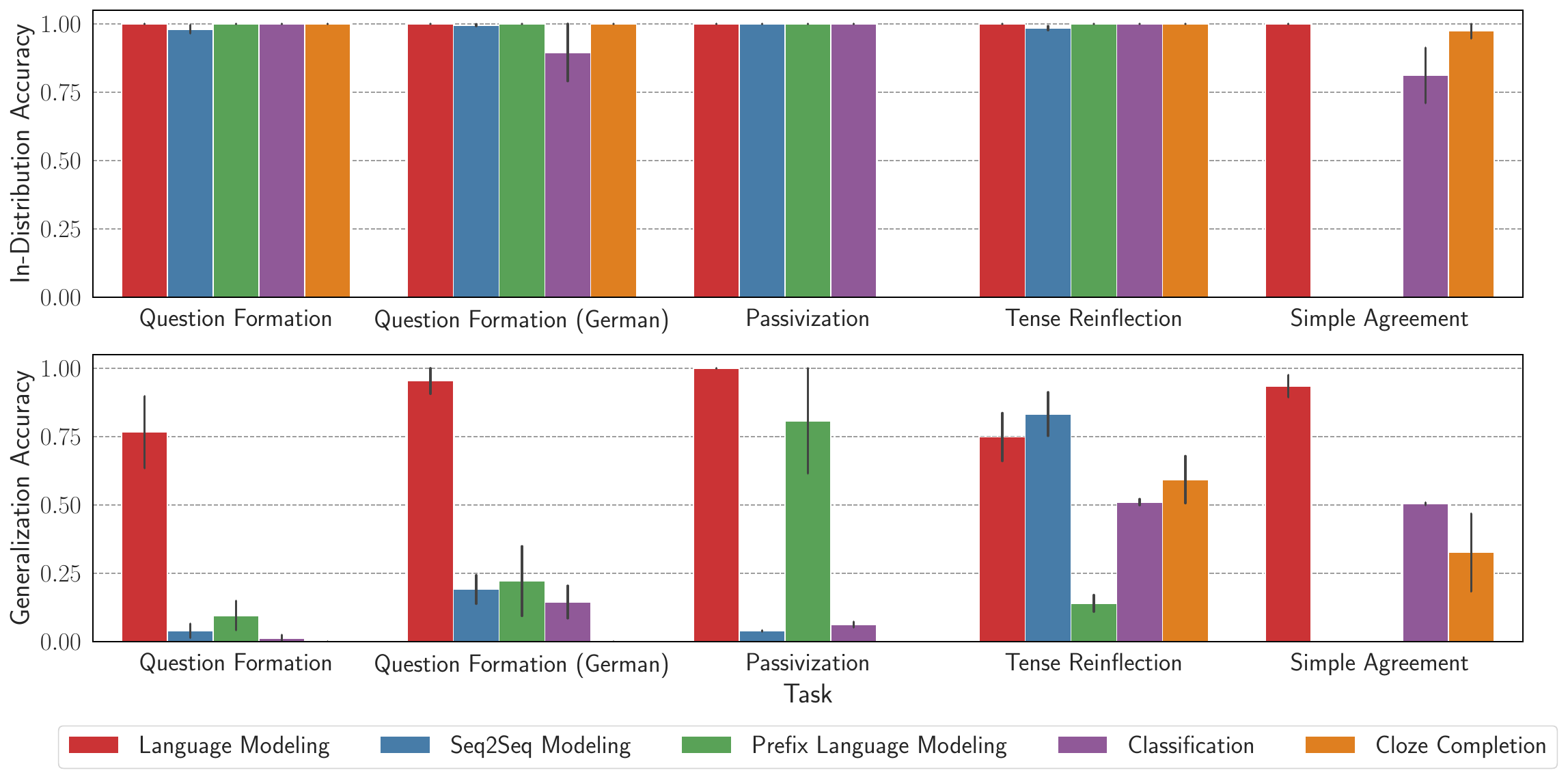}
    \caption{Effect of training objective on hierarchical generalization in transformers. The error bars correspond to the standard errors across 5 random seeds. Only the language modeling objective consistently obtains high generalization accuracy on all tasks.}
    \label{fig:obj_results}
\end{figure}

 We compare the \numobjs{} objectives for the \numds{} tasks and show the results in Figure \ref{fig:obj_results}. Notice that while all the objectives obtain close to 100\% accuracy on the in-distribution test sets (except sequence classification which performs slightly worse for simple agreement), there is a lot of variation in the \emph{generalization} accuracy.  Particularly, we observe that only the language modeling objective consistently obtains high generalization accuracy on all \numds{} tasks, while models trained with other objectives often struggle. While seq2seq and prefix LM perform well on tense reinflection\footnote{We suspect that the seq2seq model for tense reinflection might not actually be generalizing hierarchically; see discussion in Appendix \textsection \ref{sec:tiseq2seq}.} and passivization respectively, they perform much worse on the other tasks.  Thus, the choice of objective is likely the reason behind the discrepancy in the results of \citet{murty-etal-2023-grokking} showing that transformer models with the language modeling objective generalize hierarchically, and the result of \citet{petty_transformers_2021, mueller_-context_2023} showing that transformer models with  seq2seq objective do not generalize hierarchically. 


We also provide the training curves depicting the in-distribution and generalization performance of different objectives over the course of training in Figure \ref{fig:obj_tc_results} in Appendix. Consistent with the findings of \cite{murty-etal-2023-grokking}, for all the tasks, we find a delay in generalization for the LMs -- models typically obtain 100\% in-distribution accuracy much earlier than achieving high generalization performance. One might also notice that while LM objective consistently achieves high generalization performance, it is not perfect, like in the case of question formation and tense reinflection, where its average performance is roughly 75\%. Recall that these reported numbers are averaged across 5 seeds. For all the tasks we find that there are seeds for which LM models achieve 100\% generalization accuracy, and  there are also others with lower accuracy.  Apart from the two exceptions discussed above, this is not the case for other objectives, where none of the five seeds get 100\% generalization accuracy.

Interestingly, we observe (in Figure \ref{fig:obj_results}) that transformer LMs on average perform better on German question formation than the English version of the same task. We suspect this might be because the grammar used for generating German dataset is structurally more rich compared to the English grammar, as it also consists of both infinitival and past participle (depending on if the sentence consists of auxiliaries or modals) forms of the main verbs, while only infinitival forms  are included in the English version. As noted in \citet{McCoy2018RevisitingTP}, presence of rich hierarchical cues in the data can aid in hierarchical generalization.

\begin{figure}[!htbp]
\begin{center}    \includegraphics[width=0.5\linewidth]{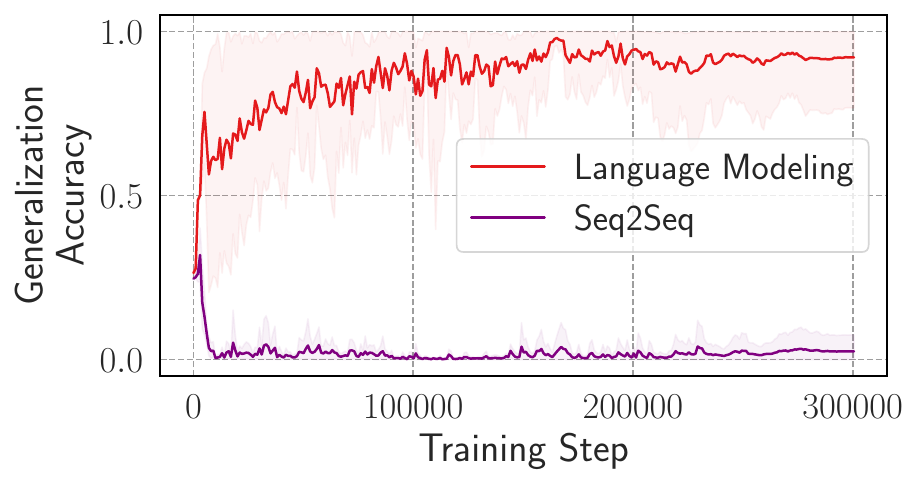}\end{center}
    \caption{Training an RNN  (GRU) using language modeling and seq2seq objectives on the question formation task. 300k training steps correspond to 24 epochs (or passes through the training data).}
    \label{fig:gru_gen}
\end{figure}

\paragraph{Revisiting \cite{mccoy-etal-2020-syntax}'s results for RNNs} Our results above suggest that the language modeling  objective imposes bias towards hierarchical generalization in transformers. Based on this, we revisit hierarchical generalization in RNNs, as the experiments of \cite{mccoy-etal-2020-syntax} were conducted using the seq2seq objective (and found to \emph{not} generalize when not using attention mechanism). We train 2-layer GRU  \citep{cho-etal-2014-learning} models using a learning rate of 0.001 using both LM and seq2seq objectives for the question formation task. As shown in Figure \ref{fig:gru_gen}, like transformers, RNNs also generalize hierarchically on the question formation task when trained using the language modeling objective, but that's not the case for the seq2seq objective. This further strengthens the evidence that language modeling acts as a source of inductive bias for hierarchical generalization. 

\paragraph{Robustness of Negative Results.} We next check how robust are our negative results for non-language modeling objectives not exhibiting hierarchical generalization across different choices of hyperparameters. We consider model depth $\in \{2, 4, 6, 8, 10, 12, 16\}$, number of attention heads per layer $\in \{2, 4, 8, 16, 32\}$, and embedding dimension $\in \{64, 128, 256, 512, 1024\}$ for all the four non-LM objectives. We vary one hyperparameter at a time while keeping the other two fixed to the default values. Additionally, for prefix-LM we consider both the variants with causal attention and bidirectional attention (the original formulation from \citet{dong-et-al-2019-unified}) on the input tokens, as we find this choice to influence the model's generalization capabilities. We provide results for the question formation task across different hyperparameter settings in Appendix Figure \ref{fig:hyperparams}. As we can see none of the hyperparameter settings results in matching the generalization performance of language modeling objective, which was 76\% on average. The closest we get is 60\% average generalization performance for prefix-LM with causal attention when using 12 layers, where 3 out of 5 seeds do end up with perfect generalization (not the case with any other objectives where none of the seeds succeed). We find this phenomenon to also hold for question formation German and passivization tasks (Figure \ref{fig:hyperparams_qfdepassiv} in Appendix), where seq2seq, prefix-LM (with bidirectional attention), classification, and cloze completion objectives fail to exhibit hierarchical generalization. However, for these tasks we do see prefix-LM with causal attention to perform on-par with language modeling. Overall, our results still show most consistent trend for hierarchical generalization when using language modeling objective, and only using prefix-LM with causal attention, which is the most similar to language modeling out of all the objectives we study, to perform on par for some hyperparameter settings (typically with larger depth). The only difference between LM objective and prefix-LM objective with causal attention, is the computation of loss over all tokens for the former, while only for the output tokens for the latter. Our results suggest that modeling loss over partial number of tokens in the sequence (e.g. only the output tokens and not the inputs) might be sufficient for hierarchical generalization in some cases as long as we perform autoregressive modeling. We leave further examination of this phenomenon for future work.

\paragraph{Takeaways.} Overall, \emph{our experiments implicate language modeling as a source of inductive bias for the neural models to generalise hierarchically}. We hypothesize that the reason LMs (approximately) learn the hierarchical rule rather than the linear rule is that, when viewing the full training objective as modeling the distribution of all the tokens in the sentence pairs and not just moving of the auxiliary to the beginning, it is the combined simplicity of the hierarchical rule and the data is greater than the linear rule and the data.  Perhaps modeling the hierarchical phrase structure is beneficial for modeling the distribution over full sequences. We will explore this hypothesis in more depth in \textsection \ref{sec:bor}. 


\section{Discovering Subnetworks with Different Generalization Behaviors}
\label{sec:subnet}
The results from 
\cite{murty-etal-2023-grokking}, and from \textsection\ref{sec:obj_results} show that the transformer  LM obtains perfect in-domain accuracy much earlier during the training, while generalization comes later. This implies that the model might be implementing something akin to the linear rule in the beginning of training and eventually generalizes to the hierarchical rule. In this section, we explore whether these rules are implemented as subnetworks in the model and ask how  these subnetworks evolve over the course of training.

\subsection{Finding Subnetworks}

Following \cite{merrill2023a} we use pruning to find the existence of subnetworks or circuits corresponding to different generalizations. In particular, we use the attention head pruning method from \citet{voita-etal-2019-analyzing}, which introduces learnable gates for each attention head of a trained transformer model. This introduces $\texttt{number of heads} * \texttt{number of layers}$ learnable parameters, which is typically equal to 48 in our experiments.  Pruning is then performed by training these learnable gates (while freezing the original model parameters) to minimize negative \llike{} objective, but also adding an  $L_0$-penalty as regularization to ensure sparsity. Since $L_0$-norm is nondifferentiable, a stochastic relaxation is used, which considers the gates as random variables drawn from head-specific hard concrete distributions \citep{louizos2018learning}. \kabir{Do we need more details about pruning algorithm here?}  \nascomment{not if you're following earlier work closely, though I think there was a piece missing, which I added, above} After completion of pruning, all the gates are either fully open or closed, and a closed gate implies that the output of the corresponding head is zeroed-out in the computation of multi-head self-attention. In the case that all heads in a layer are zeroed-out, that particular layer is skipped: inputs to the layer pass through unchanged to the next layer due to the residual connections.

Thus the pruning procedure does not modify any weights of the original model and merely performs  subset selection on  attention heads of the model. To find subnetworks consistent with different generalizations (linear-rule and hierarchical rule) we introduce three pruning strategies which differ in the data used for pruning:

\noindent \textbf{1. \trprune} uses the original ambiguous training dataset to prune the attention heads. The subnetwork thus found is likely to be a compressed version of the full model.

\noindent \textbf{2. \genprune } uses a small fraction of the generalization set (1\% or 100 examples) to prune the attention heads. If successful, this pruning would yield a subnetwork consistent with hierarchical generalization---obtaining close to $100\%$ generalization accuracy. 

\noindent \textbf{3. \spprune } is minimizing the (negative \llike) loss on the training data and \emph{maximizing} it for the (1\%) generalization data. In this case, successful pruning should yield a subnetwork that exhibits generalization consistent with the linear rule, i.e., obtains $0\%$ generalization accuracy but obtains $100\%$ in-distribution accuracy.

\paragraph{Experimental setup.} Unless specified otherwise, for pruning, we use a learning rate of $0.05$, the $L_0$ regularization penalty coefficient as $0.015$, and train for 10k steps, which we found to work well across different pruning settings.  Here we report the experiments for the question formation task and discuss the others in Appendix \textsection \ref{sec:subnetworks_tisa}, for which we also obtain consistent results. Note that since we are interested in discovering subnetworks implementing hierarchical and linear rules, while pruning, we only use the negative \llike{} of the first auxiliary in the question for computing the loss. 
To make sure that the discovered subnetworks are not just a by-product of the pruning procedure, we also consider control groups, which are obtained by pruning randomly initialized networks. 

\subsection{Results}

\begin{figure}[!htbp]
    \centering
    \includegraphics[width=0.95\linewidth]{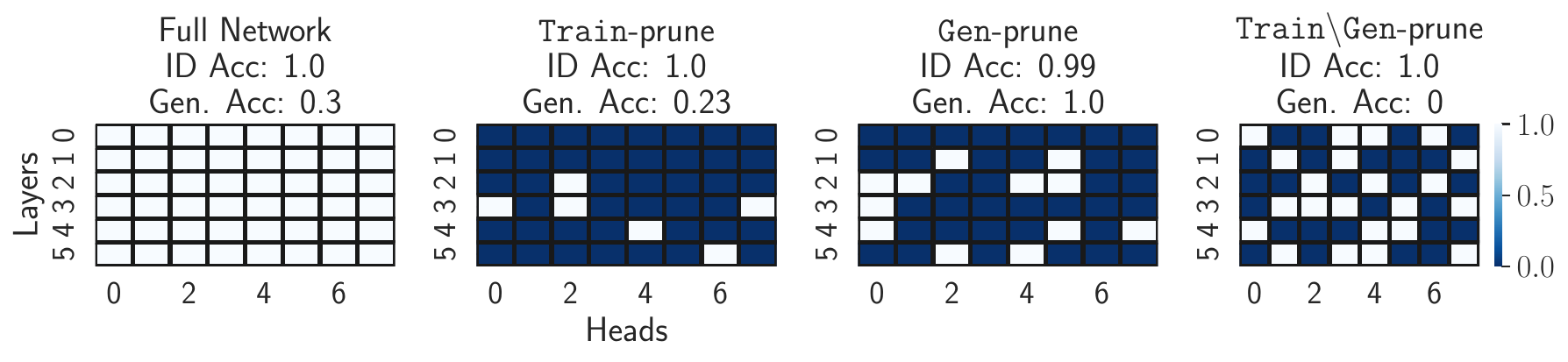}
    \caption{Pruning a transformer LM trained for 15000 steps
using the three methods. Dark blocks mean that the head is pruned and light means it is kept. 
} 
    \label{fig:head_prune}
\end{figure}

In Figure \ref{fig:head_prune}, we show the effect of different pruning methods on an intermediate model checkpoint, which does not yet generalize hierarchically
(the model before pruning has a generalization accuracy of 30\%).
After \trprune{}, roughly ~80\% heads of the full model are removed and in-distribution performance is conserved, though there is a drop in generalization performance (30\% to 23\%). 
After \genprune{}, we are able to find a subnetwork that achieves 100\% generalization accuracy.  This is striking, because the full network performed much worse. 
After \spprune{}, we find a subnetwork that achieves $0\%$ generalization accuracy while having $100\%$ in-distribution performance; this subnetwork is behaviorally equivalent to the linear rule. Hence, these pruning experiments reveal the existence of subnetworks implementing different generalization behaviors.

\begin{figure}[!htbp]
    \centering
    \includegraphics[width=0.95\linewidth]{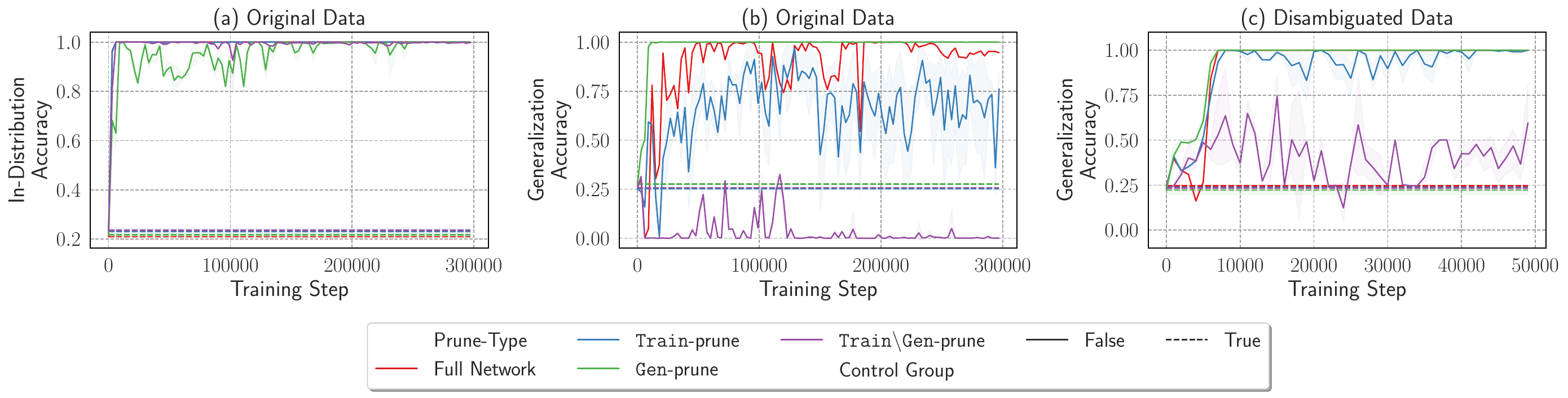}
    \caption{Tracking training dynamics with respect to the three pruning methods's subnetworks and the full network. (a) and (b): in-distribution and generalization accuracies of the LMs trained on the original ambiguous question formation data after pruning using the three methods, (c): generalization accuracy after pruning the model trained on disambiguated data. For models trained with original data, we can discover sub-networks consistent with hierarchical rule as well as the linear rule, while for the models trained with disambiguated data, linear rule subnetwork is not found (indicated by the curve corresponding to \spprune{} never approaching 0\% generalization accuracy).\nascomment{can you make these a bit bigger?  also sugest making y-axis go farther, so I can see the lines that are at the top (esp in (a)).  maybe also add a sentence summarizing what we take away from these figures, for skimming readers} \kabir{added.}}
    \label{fig:dynamics_all}
\end{figure}

\paragraph{Training dynamics.}  We now consider how these different subnetworks evolve over the course of the model training. To this end, we save checkpoints after every 1000 steps of training and perform the three kinds of pruning on those checkpoints. As shown in Figure \ref{fig:dynamics_all}({b}), the ``linear-rule'' subnetwork becomes discoverable through pruning \nascomment{check wording} at roughly 6000 training steps -- indicated by the 0\% generalization performance for \spprune{} curve and 100\% in-distribution performance in Figure \ref{fig:dynamics_all}(a). Soon after, we see in Figure \ref{fig:dynamics_all}(b), the formation of a (hierarchical) generalization subnetwork, at around 9000 training steps, where the subnetwork obtained using \genprune{} obtains 100\% generalization accuracy (the full network at this point only obtains 5\% generalization performance). \nascomment{more references to the figures are needed, I'm not sure where to look} While there are occasional spikes \nascomment{do you mean drops?} \kabir{I meant spikes in the purple curve. I had incorrectly written \genprune{} in the following line before, fixed to \spprune{} now} in the generalization performance of subnetworks found using \spprune{} during the first-thirds of the training,  during majority of the training (especially the latter two-thirds) both the linear and hierarchical generalization subnetworks continue to be discoverable over the course of training -- as indicated by stable 100\% generalization accuracy of subnetworks found by \genprune{} and 0\% generalization accuracy for \spprune{} subnetworks in Figure \ref{fig:dynamics_all}(b) \nascomment{what figure?}. 

Our experiments reveal that, throughout training, there is  competition between the two sub-networks, and while the behavior of the aggregate model becomes closer to the hierarchical rule with training, the competing linear-rule subnetwork does not really disappear. All three pruning methods are unsuccessful on the control group (randomly initialized networks), providing further evidence that these subnetworks are not introduced by the pruning methods, and behaviors akin to the hierarchical and linear rules are implemented within the language model.


We hypothesize that the ambiguous training data (with two plausible generalizations, linear and hierarchical) is the reason for the existence of the subnetworks with very different generalization behaviors. To evaluate this hypothesis, we consider the case where the model is trained with \emph{disambiguated} data -- we augment the ambiguous training dataset with examples that are only consistent with the hierarchical rule.\footnote{We add 10k such examples to the existing dataset containing 100k examples.} We plot the training dynamics of the model trained with this disambiguated data in Figure \ref{fig:dynamics_all}(c). The full model without any pruning in this case, as expected, generalizes perfectly after a few thousand training steps (see \ref{fig:dynamics_all}(c)) without any noise, in contrast with generalization accuracy of the full-model trained on ambiguous data in Figure \ref{fig:dynamics_all}(b)\nascomment{refer to figure}. More interestingly, Figure \ref{fig:dynamics_all}(c) shows that \spprune{} fails to yield subnetworks that obtain 0\% generalization accuracy, in contrast to the ambiguous data case in figure \ref{fig:dynamics_all}(b). To make sure this is not due to the choice of our pruning hyperparameters, we conduct an extensive  hyperparameter search, consisting of 128 combinations of the pruning learning rate, regularization penalty, and pruning steps (using Bayesian optimization) and still fail to find the setting where the \spprune{} succeeds for the disambiguated data case (see Figure \ref{fig:spprune_disamb_hyptune} in Appendix). This strongly suggests that the ``linear-rule'' subnetwork is never formed in the language model when it doesn't have a reason to be learned -- when the alternate generalization behavior (linear rule) is no longer applicable to the entire training dataset. 

We also experiment with the opposite disambiguation setup, where we augment the training data with examples only consistent with the linear rule, and in line with our findings, we find that the \genprune{} fails to find a subnetwork with 100\% generalization accuracy -- no subnetwork consistent with hierarchical rule is formed (Figure \ref{fig:dynamics_disamb_lin} in Appendix). Hence, \emph{ambiguity in the training data appears to drive the joint existence of these contrasting subnetworks.}

\paragraph{Effect of Head Capacity.} We next investigate what effect does the number of attention heads in the network have towards the presence of the two subnetworks. In addition to our original models trained with 8 heads per layer, we also train models with 16 heads, 4 heads, and 1 head and analyse the training dynamics using the same procedure as above. For models with 16 and 4 heads per layer (Figures \ref{fig:dynamics_16heads} and \ref{fig:dynamics_4head} in Appendix) we find results consistent with 8 heads case i.e. the two subnetworks form early during the training and continue to co-exist throughout the training. 
Interestingly, for 1 head per layer case (Figure \ref{fig:dynamics_1head} in Appendix), we find that both \genprune{} and \spprune{} on average (across the 5 seeds) fail to discover their respective subnetworks. This is particularly noteworthy, because the full model with 1 head on average gets better generalization performance than the 4 heads per layer model (0.7 for the former and 0.6 for the latter). Characterizing the behavior of 1 head model remains an open question as we didn't find evidence of either of the two presupposed generalizations, which we leave for the future work.
Overall, these results show that the head capacity i.e. the number of attention heads might be crucial for the development of the subnetworks corresponding to distinct generalization behaviors.

\section{Why Do Transformer-Based LMs Generalize Hierarchically?}
\label{sec:bor}

A useful tool for understanding generalization in  neural networks has been ``simplicity bias'', where the inductive bias towards simpler functions \citep{de2019random} has been shown to explain why neural networks tend to generalize instead of overfitting the training data \citep{valle-perez2018deep, bhattamishra-etal-2023-simplicity}. In our case, we are not interested in comparing the learned behavior of the language models (hierarchical rule) with the overfit solution, but instead with an alternate generalization (linear rule). Can we explain through ``simplicity'' the preference of the model towards hierarchical generalization? This might sound counterintuitive, because at least on surface it appears that the linear-rule should be simpler compared to the hierarchical rule. Our main argument is that when considering transformers trained with the language modeling objective, since the underlying data-generation process to be modeled produces each token in the full sequence (not, for instance, just the first auxiliary in the question formation task), modeling the dependencies between the tokens hierarchically as opposed to learning a linear rule for each dependency, might be simpler.\footnote{Such an argument is implicit in the field of theoretical syntax where hierarchical representations are rife.} In this section, we present a study showing some evidence for the simplicity of the hierarchical
generalization over the linear-rule based to explain the preference for the former by transformer  LMs. We leverage the Bayesian framework of \cite{PERFORS2011306}, utilizing generative grammars to model data-generation processes corresponding to the hierarchical and linear rules, and operationalize the notion of simplicity and goodness of fit using the posterior probabilities of the grammars given the observed data. We then show that there is a correlation between transformers' ability to generalize hierarchically and the training dataset being better explained using a hierarchical grammar than a regular one (which models the linear rule) according to the posterior criterion. 


\subsection{Background}
\paragraph{Operationalizing the notion of simplicity.} We make use of \textit{Solomonoff's theory of inductive inference} \citep{SOLOMONOFF19641}, which formalizes Occam's razor -- when two hypotheses explain the data equally well, the simpler one of the two is likely to be the correct one. This notion is mathematically formalised in Solomonoff's theory using a Bayesian approach by computing the posterior probabilities of the competing hypotheses and selecting the one with higher posterior.
\begin{equation*}
    p(h \mid D) \propto p(D \mid h) \cdot p(h)
\end{equation*}
Here, $p(D \mid h)$ denotes the likelihood of the observed data $D$ based on the hypothesis $h$, i.e.,  how well $h$ explains the data $D$. $p(h)$ denotes the prior probability of $h$, which in Solomonoff's theory assigned higher values for simpler hypotheses $h$. 
In other words, a more complex hypothesis will entail making more choices (``high program length'') and hence have a lower prior probability. Hence, by computing the posterior $p(h \mid D)$, Bayesian inference balances the tradeoff between the goodness of fit of a hypothesis (likelihood) and its simplicity (prior). This is closely related to ``Bayesian Occam's razor'' and the Minimum Description Length principle \citep{RISSANEN1978465}. \kabir{@Navin, can you please check the soundness of this paragraph?} \nascomment{there's a lot of literature that pushes this line since the 60s and 70s, specifically for reasoning about structure in language.  not sure it's worth getting into but you can find some pre-2006 citations on p. 61 of my thesis (\url{https://homes.cs.washington.edu/~nasmith/papers/smith.thesis06.pdf})}


\paragraph{Probabilistic grammars.} We mentioned in the previous paragraph that computing the posterior over the competing hypotheses can help us choose the one which better balances the trade-off between goodness of fit and simplicity. But for our problem, what form should these hypotheses or ``programs" take to represent the linear and hierarchical rules? Note that since our training objective is language modeling, we need to consider the hypotheses that generate the entire sequence of tokens as represented in the training data. Following \cite{PERFORS2011306}, we use probabilistic generative grammars to model the data-generation process.

For the purposes of this work we consider probabilistic context-free grammars (PCFGs) that can be represented using a 5-tuple i.e., $G = \{V, \Sigma, R, S, P\}$. Here, $V$ denotes the set of nonterminal symbols that form phrases or constituents in a sentence, $\Sigma$ denotes the set of terminal symbols or words in the sentences, $R \in V \times \{V \cup \Sigma\}^*$ denotes the set of production rules mapping phrases to sub-phrases or words, $S \in V$ is the start symbol that represents the whole sentence, and $P$ denotes the probabilities on the production rules. Specifically, for a given non-terminal when there are multiple productions possible, $P$ assigns a probability to each possible production rule. To generate data from a PCFG, we start from the start symbol $S$ and for each terminal that arises we apply a production rule sampled according to $P$ and repeat the procedure till we are only left with terminals to generate sentences.  PCFGs are typically used to model the hierarchical phrase structure of a language. We can also apply some constraints to the form of production rules in $R$ to obtain special cases (subsets) of CFGs. For example, regular grammars form a subset of CFGs, whose production rules can be put into a right-linear form: $A \to bC$, where $A$ and $C$ are nonterminal symbols and $b$ is a terminal. 

\paragraph{A Bayesian view of language generation.} We can view the data-generation process that generates dataset $D$ using the probabalistic grammar $G$
Given the dataset $D$, we can compute the posterior $p(G \mid D) \propto p(D \mid G) \cdot p(G)$, where $p(D \mid G)$ is the likelihood of the data given the probabilistic grammar, and $p(G)$ measures the simplicity of $G$. To get an intuitive understanding of the prior probability of a grammar and how it encodes simplicity, recall that grammars that we consider are 5-tuples $\{V, \Sigma, R, S, P\}$. Hence choosing a grammar $G$ involves making choices like the number of nonterminal and terminal symbols, number of production rules from each nonterminal, nature of the production rule etc. By assigning probability distributions to each of these choices, we can compute the prior probability of a given grammar. We can choose the prior distribution that favours simpler grammars, e.g., following \cite{PERFORS2011306}, we use geometric distributions for the number of nonterminals and productions, hence a simple grammar with fewer nonterminals and productions will receive a higher probability compared to a more complex grammar. We can hence compute the posteriors for the grammars representing the competing generalization hypotheses (hierarchical and linear rule) to compare how each of these balances the tradeoff between goodness of fit and simplicity.

\subsection{Method}

We now discuss how we apply the Bayesian Occam's razor approach discussed above to explain why transformer language models generalize hierarchically. As an overview of our approach, we start by constructing a PCFG to model the hierarchical rule (denoted \cfg) and a regular grammar (\reg) that generates data based on the linear rule. We then generate data using both the grammars -- \cfgd{} from \cfg{} and \regd{} from \reg{}. The intersection of the two datasets, \cfgd$ \cap $ \regd, is comprised of  ambiguous examples  consistent with both the linear rule and hierarchical rule. We  will use this as our training corpus \dtrain{}. We then compute the posterior probabilities for both \cfg{} and \reg{} given \dtrain{} and select the one with the higher posterior: $G^* = \argmax_{G \in \{\cfg, \reg\}}p(G \mid \dtrain)$. We then train a transformer language model on \dtrain{}, and check if it generalizes according to $G^*$. Specifically, if $G^* = \cfg$, does the transformer follow the hierarchical rule, and if $G^* = \reg$, does the transformer follow the linear rule?  The selection of $G^*$ is intended to simulate ``idealized'' Bayesian learning, and to the extent that the transformer's learning behavior matches $G^\ast$ across different scenarios, we find support for a simplicity bias in the transformer's training setup.
\kabir{Will it be better to have these as a list of steps instead?} \kabir{Any other relvant detail missing from the overview here?}  \nascomment{I think it's ok, I added a little more} In what follows, we provide details about each of these steps.

\paragraph{Task.}For the purposes of this study we consider the simple agreement task, as constructing hierarchical and linear grammars for its data is straightforward.\footnote{Question formation and tense reinflection involve pairs of sentences, where the second sentence is a transformed version of the first. Such sentence pairs would likely require more complex frameworks like synchronous grammars \citep{Aho1969SyntaxDT}, which we leave to future work.}

\paragraph{Constructing grammars for simple agreement.} Following \cite{PERFORS2011306}, we hand-construct the CFG and regular grammars. The CFG is constructed so that each verb agrees with the hierarchically connected subject, while the regular grammar is constructed to follow the linear rule (each verb in the sentence agrees with the most recent noun). The constructed grammars are assigned uniform probabilities for the production rules i.e., given a nonterminal, all the productions are equally likely. For an example of productions from both the grammars see Figures \ref{fig:cfg_prod_ex} and \ref{fig:lg_prod_ex} in the Appendix. For constructing CFG, we use Chomsky Normal Form for the productions: Each production rule is of the form $A \to BC$ or $A \to a$, where $A, B, C$ are nonterminals and $a$ is a terminal symbol. Similarly, for the regular grammar \reg{}, we use the right-linear form of productions: Every rule is of the form $A \to bC$ or $A \to a$. 

Following \cite{PERFORS2011306}, we adopt a type-based approach for constructing the grammars: terminal symbols $\Sigma$ instead of being the word tokens (e.g. \emph{walrus}, \emph{sing}) are syntactic categories  (e.g., determiner, singular-noun, intransitive-verb, etc.), so that we can use these grammars to strictly model abstract syntactic structures and not vocabulary-type frequencies, and it also gives us a manageable number of possible generations by the grammars. 

For both context-free and regular grammars we generate two variants, depending on the diversity of the sentence types generated by them:

\noindent \textbf{Small grammars \cfgs{} and \regs{}}: Here we construct CFG and regular grammars that only generate 18 sentence types. Recall that a sentence type is a sequence of syntactic categories, e.g., sentences like \textit{The walrus sings} can be represented by sentence type \textit{determiner singular-noun intransitive-verb}. Different sentence types in this case differ by the plurality of the nouns (singular or plural), type of verbs (transitive or intransitive), and presence or absence of prepositional phrases accompanying the nouns. The resulting hand-constructed \cfgs{} in this case has 15 nonterminals and 21 production rules and \regs{} has 14 nonterminals and 22 production rules. Both grammars have the same 8 terminals. Out of the 18 sentence types generated by both the grammars, 12 are common between the two (ambiguous) and 6 remaining in \cfgs{} that are only consistent with the hierarchical rule and 6 only consistent with linear rule in \regs{}.

\noindent \textbf{Large grammars \cfgl{} and  \regl{}}: In this case we consider larger grammars, which can generate much more diverse sentence types -- 180 sentence types. The major difference with the smaller grammars here is that they are allowed to generate relative clauses, which can be present at both the subject and object in the sentence. \cfgl{} has 25 nonterminals and 38 productions, while \regl{} has 41 nonterminals and 63 productions. Note that based on these numbers alone it is evident that we need much more complex regular grammars to generate diverse sentence types. Out of the 180 sentence types generated by each grammar, 120 are common between the two, and the remaining sentence types are only generated by the specific grammars (following either hierarchical or linear rule).

\paragraph{Generating datasets.} We generate the sentence types from each of the 4 grammars -- \cfgds{}, \regds{}, \cfgdl{}, and \regdl{}. As mentioned before, the training dataset is constructed by considering the sentence types common between the CFG and corresponding regular grammar.  We have $\dtrains{} = \cfgds{} \cap \regds{}$ for the small grammars, and $\dtrainl{} = \cfgdl{} \cap \regdl{}$ for the larger ones. Note that these are the datasets of sentence-types, and transformers are trained on sentences. To generate sentences from these type corpora, we repeatedly sample sentence types, and replace the syntactic categories with the allowed tokens for that category (e.g., \textit{determiner} can be replaced with \textit{the}, \textit{our}, \textit{my}, etc.). Using this procedure we generate a corpus of 50k sentences from $\dtrains{}$ and 50k sentences from $\dtrainl{}$. Note that the simple agreement experiments in \textsection \ref{sec:objectives}, were performed using the latter dataset derived from $\dtrainl{}$.

The generalization test sets are generated by considering the sentence types that are unique to a specific grammar. E.g., we can have the test set $\dtestcfgs = \cfgds  \setminus \regds$, which contains sentence types that are unique to \cfgds{} and hence only consistent with the hierarchical rule and not the linear rule. Similarly, $\dtestregs = \regds \setminus \cfgds$, consists of sentence types consistent only with the linear rule. We can equivalently define $\dtestcfgl$ and $\dtestregl$. While talking about the two datasets in general and not specifically about the small ($S$) or large ($L$) variants, we just use the notation $\dtestcfg$ and $\dtestreg$.

\paragraph{Computing the posterior for each grammar.} Now that we have the four grammars constructed, we can compute the posteriors for the grammars given the corresponding training datasets. Note that, since we are only interested in comparing the posteriors of CFG and regular grammars, we can estimate the posterior by computing the likelihood and prior and taking product of the two, i.e., $p(G | D) \propto p(D \mid G) \, p (G)$. Recall that the prior probability of a grammar can be computed by calculating the probability of each of the choices that goes into defining that grammar:

\begin{equation}
\label{eq:prior}
    p(G) = p(|V|)\prod_{k = 1}^{|V|}p(P_k) \, p(\theta_k)\prod_{i=1}^{P_k}p(R_{k,i}).
\end{equation}

Here, $|V|$ is the number of nonterminals, $P_k$ is the number of productions from the $k^{\mathrm{th}}$ nonterminal with the probabilities of each production given by $\theta_k \in [0,1]^{P_k}$,  and $R_{k,i}$ denotes the right hand side of the $i$th production rule from the $k$th nonterminal. Following, \citet{PERFORS2011306}, we use a geometric prior on $p(|V|)$ and $p(P_k)$. Recall that the geometric distribution is given by $p(n; p) = (1-p)^{n-1}p$, where $p$ is a parameter of the geometric distribution, often interpreted as the probability of success, and a geometric distribution models the probability of success after $n$ trials. Hence, choosing a geometric prior penalizes the grammars with a large number of nonterminals ($|V|$) and productions per nonterminal ($P_k$). In our experiments we use $p = 0.5$, following \citet{PERFORS2011306}, but we  conduct a sensitivity analysis on the choice of this parameter \nascomment{if this is in appendix, say so} \kabir{this is discussed in main text, only the plots are in Appendix which we explicitly refer in the main text later}.  For $\theta_k$, we use a flat (i.e., $\alpha = 1$) Dirichlet prior, a popular choice for modeling 
 probabilities for categorical distributions ($K - 1$ simplex). Note that since the Dirichlet is a continuous distribution, the probability of any specific $\theta_k$ is zero and we use the discrete relaxation from \cite{PERFORS2011306} to model $p(\theta_k)$.  The probability of the production rule $p(R_{k,i})$, depends on the type of grammar. For CFGs, since we consider them in CNF, the production rules are of the form $A \to BC$ or $A \to a$, hence the probability of the right hand side can be given by, $p(R_{k,i}) = \frac{1}{2} \frac{1}{|V|^2}\mathds{1}(|R_{k,i}|=2) + \frac{1}{2} \frac{1}{|\Sigma|}\mathds{1}(|R_{k,i}|=1)$. Since the regular grammars are in the right linear form i.e. productions of the form $A \to bC$ or $A \to a$, we can compute  $p(R_{k,i}) = \frac{1}{2} \frac{1}{|\Sigma|}\frac{1}{|V|} \mathds{1}(|R_{k,i}|=2) + \frac{1}{2} \frac{1}{|\Sigma|} \mathds{1}(|R_{k,i}|=1)$. One might notice that we are missing the probability of number of terminal symbols $p(\Sigma)$ in the prior equation. We ignore this because both the CFG and regular grammars have the same number of terminals in our experiments, and since we are interested in just comparing the probabilities, the inclusion or exclusion of $p(\Sigma)$ doesn't make a difference.\footnote{One might also notice that $p(G)$ allows some probability for generating the same rule more than once; it ``leaks'' probability mass.  No prior literature, to our knowledge, suggests that this should pose a problem to our analysis.}


The likelihood $p(D \mid G)$, measures the probability that the dataset $D$ is generated from the grammar $G$. For $m$ sentence types in the dataset $D$, the likelihood is given by

\begin{equation}
    p(D \mid G) = \prod_{i=1}^m p(S_i \mid G),
\end{equation}

where $S_i$'s denote the sentence types in $D$. $p(S_i \mid G)$ is computed by taking product of the probabilities of production rules used to derive $S_i$ using $G$ (including adding the probabilities when multiple parses are possible for $S_i$). Note that computing $p(S_i \mid G)$ requires estimating the production probabilities $\theta_k$ from each nonterminal. We use the Inside-Outside algorithm \citep{Baker1979TrainableGF}, to obtain an approximate maximum likelihood estimate of the production probabilities on the dataset $D$. Hence, having computed both the prior $p(G)$ and $p(D \mid G)$, we can compute the posterior $p(G \mid D)$.

\paragraph{Other choices of grammars.} Given our generated training datasets ($\dtrains, \dtrainl$), there can be grammars other than the four we constructed that can generate these datasets.  In their analysis, \citet{PERFORS2011306} also consider two subsets of the regular grammars:  Flat and  One-state.  Flat grammars have production rules which are  the list of memorized sentences, i.e., of the form $S \to a_1 a_2\cdots a_n$. Here $a_i's$ are terminal symbols and there are no nonterminals other than $S$. Hence, flat grammars can be used to model memorization without generalization. One-state grammars are equivalent to finite state automata  with a single state and hence permit any terminal symbol to follow any other. We also include these two grammars in our analysis.

Further, even among the class of context-free and regular grammars, there might exist grammars with better posteriors on the training datasets $\dtrains$ and $\dtrainl$ than the ones that we hand-construct. To remedy this, we also experiment with applying local search on our constructed grammars, using Bayesian model merging \citep{stolcke1994inducing} to minimize the grammars while improving the posterior on the respective training datasets. While in the main text we discuss the results for hand-constructed grammars, we provide the details on the minimization algorithm and the corresponding results for minimized grammars in \textsection \ref{sec:grammar_appendix}. \nascomment{since you don't give names to these simpler grammars, it makes me think  this is a side thing and they won't show up in the experiments.  but it's not actually clear whether the table below is using this approach or not! be explicit}

\paragraph{Explaining generalization in transformers.} Recall that our goal has been to quantify the notion of simplicity of the two competing hypotheses (hierarchical and linear rule), which are consistent with the training data used to train transformer-based LMs. The posterior probabilities of the two types of grammars are a way to measure which grammar better balances the trade-off between the goodness of fit and simplicity. Our aim is to check whether the trained transformer LM exhibits generalization consistent with choosing the simpler (i.e., larger posterior) grammar.  We evaluate this by comparing the negative log-likelihood (NLL) assigned by the transformer LM to the test sets corresponding to the two generalizations. E.g. for the transformer model trained using data derived from $\dtrainl$, we evaluate its NLL on the generalization test sets derived from the two grammars $\dtestcfgl$ and $\dtestregl$, and check if it assigns a lower NLL to the test data coming from the simpler grammar. Note that we compute the average NLL overall all examples in a test set for the final value. For a more intuitive metric, we also compute the main-verb accuracy -- fraction of test examples for which the verb predicted by the model agrees with the hierarchically associated noun in the sentence. A main-verb accuracy of 1 indicates that the model's generalization is consistent with the hierarchical rule and an accuracy of 0 when it is consistent with the linear rule.

\subsection{Results}

\paragraph{Comparing posteriors.} The $\log$-probabilities for all the hand-constructed grammars on the two datasets is provided in 
Table \ref{tab:posteriors}. On both datasets, the one-state grammar gets the highest \emph{prior}, which is expected as it is the simplest grammar that we study. However, the one-state grammar also fits the data the worst which is indicated by the lowest \llike (for both  datasets). The flat grammars fit both the datasets the best and have the highest \llike,  which is also expected since a flat grammar  memorizes the training data. But it can come at a cost of increased complexity, especially when the training data is diverse; and so the flat grammar has the lowest \lprior on the full dataset. 

For the high diversity dataset $\dtrainl$, we observe that the CFG best balances the tradeoff between the simplicity and goodness of fit, obtaining the highest posterior. This shows why it would be more beneficial to model this dataset using a hierarchical phrase structured grammar than a linear grammar. However, when we consider the low-diversity dataset $\dtrains$, while the CFG still obtains a better posterior than the regular grammar, it is the one-state grammar  obtains the highest posterior out of all the grammars. This is consistent with the findings of \citet{PERFORS2011306}, who found that for small corpora, one-state grammars often obtain higher posteriors than the context-free and regular grammars. In such cases, learning the distribution of syntactic category sequences, without abstract nonterminals, wins out on the Bayesian criterion.

We obtain consistent findings with some subtle differences for the grammars minimized using the Bayesian model merging algorithm, which we detail in \textsection \ref{sec:grammar_appendix}.


\paragraph{Sensitivity to prior.} Note that choosing the prior is subjective and can influence these results. Hence, to be extra careful, we conduct a sensitivity analysis by varying the values of the geometric distribution parameter $p$. We experiment with $p \in \{0.01, 0.1, 0.2, 0.3, \cdots, 0.9, 0.99\}$ for the probability distribution on the nonterminals ($p(|V|)$) and number of productions  ($p(P_k)$), and obtain findings consistent with those in Table \ref{tab:posteriors} (see Figure \ref{fig:prior_senst} in Appendix). We also experiment with having different values of $p$ parameter for $p(|V|)$ and $p(P_k)$, and try out 49 combinations ($\{0.01, 0.1, 0.3, 0.5,  0.7, 0.9, 0.99\} \times \{0.01, 0.1, 0.3, 0.5,  0.7, 0.9, 0.99\}$) . For each of these combinations, we find that for $\dtrains$ case, consistent with Table \ref{tab:posteriors} the $\cfgs$ always obtain a lower posterior compared to the $\onest{}$ grammar. Similarly for the $\cfgl$ and $\regl$, the findings are also consistent across all 49 combinations i.e. $\cfgl$ always obtain a higher posterior than $\regl$.

\nascomment{I think this sensitivity analysis is very important, but you're not giving the reader any confidence that you did this rigorously.  to make the strongest possible argument, you would show just how far you have to push the prior in favor of simple grammars before you get the counter-intuitive result} \kabir{I hope now its more clear and convincing?}


         

\begin{table}[!htbp]
    \centering
    \small
    \caption{Comparing the \lprobs for each of the 4 grammars given the training datasets $\dtrainl$ and $\dtrains$.}
    \resizebox{0.95\textwidth}{!}{
    \begin{tabular}{lrrr|rrr}
         \toprule
         \multirow{2}{*}{Grammar} & \multicolumn{3}{c|}{$\dtrainl$ (120 types)} & \multicolumn{3}{c}{$\dtrains$ (12 types)}  \\
         \cmidrule{2-7}
         & $\log$-Prior & $\log$-Likelihood & $\log$-Posterior & $\log$-Prior & $\log$-Likelihood & $\log$-Posterior\\
         \midrule
         \textbf{\cfg{}} & -367 & -639 & \textbf{-1006} & -169 & -34 & -203 \\
         \textbf{\reg{}} & -619 & -616 & -1235 & -190 & \textbf{-30} & -220 \\
         \textbf{\flt{}} & -4567 & \textbf{-574} & -5141 & -281 & \textbf{-30} & -311 \\
         \textbf{\onest{}} & \textbf{-58} & -2297 & -2355 & \textbf{-51} & -121 & \textbf{-172}\\
         \bottomrule
         
    \end{tabular}
    }
    \label{tab:posteriors}
\end{table}

\begin{figure}[!htbp]
    \centering
        \begin{subfigure}{0.25\textwidth}
    \centering
    \includegraphics[width=0.99\textwidth]{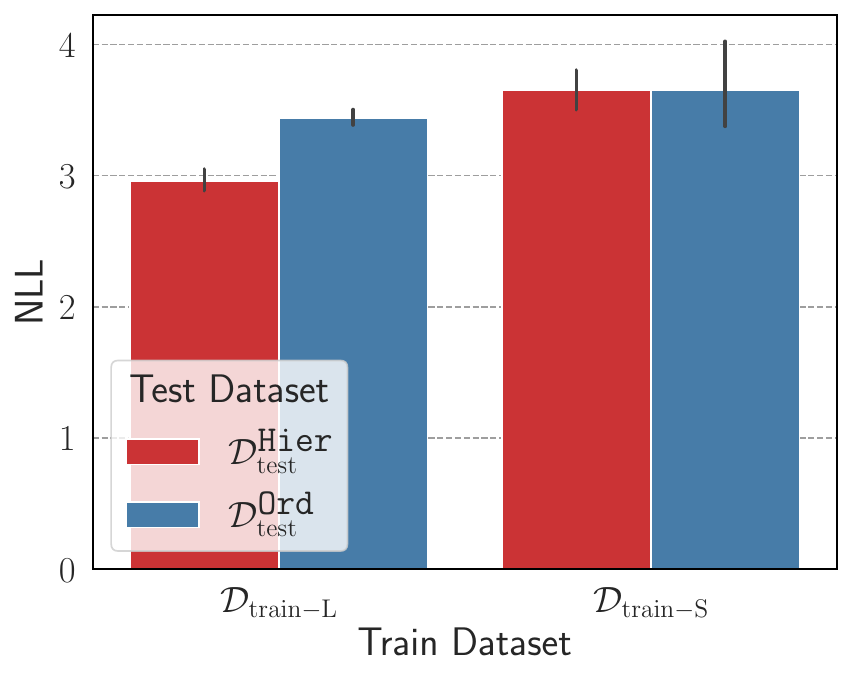}
    \caption{Negative log-likelihood evaluated on the hierarchical ($\dtestcfg$) and linear ($\dtestreg$) generalization test sets.}
    \label{fig:sa_bor_ppl}
    \end{subfigure}\hfill
    \begin{subfigure}{0.33\textwidth}
    \centering
    \includegraphics[width=\linewidth]{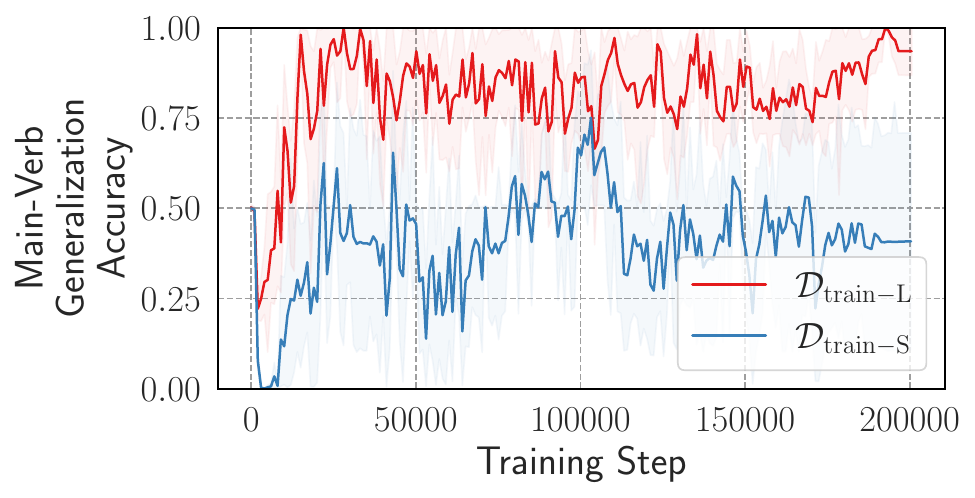}
    \caption{Main-verb generalization accuracy on the $\dtestcfg$ test set.}
    \label{fig:sa_bor_acc}
    \end{subfigure}\hfill
    \begin{subfigure}{0.33\textwidth}
    \centering
    \includegraphics[width=\linewidth]{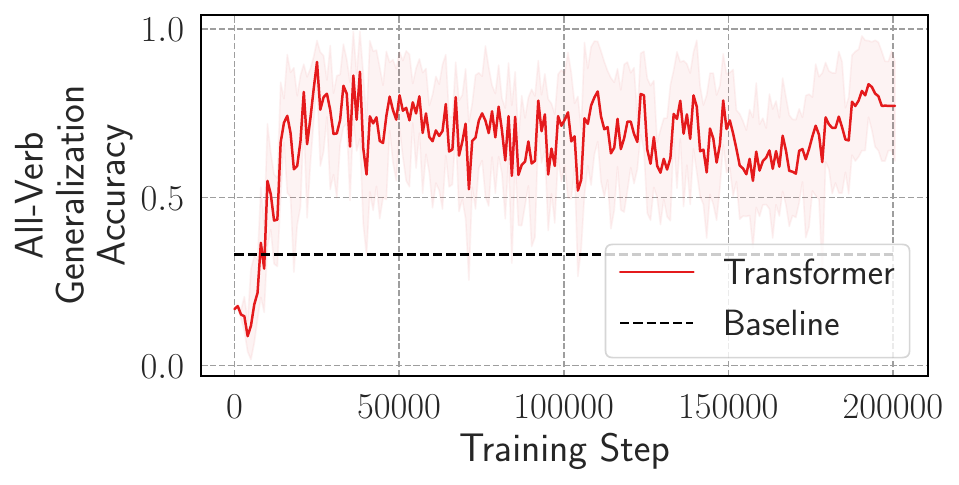}
    \caption{All-verb generalization accuracy on $\dtestcfg$ test for the model trained on the 120-types dataset.}
    \label{fig:sa_bor_all_verb}
    \end{subfigure}\hfill
    \caption{Performance of transformer models trained on the $\dtrainl$ and $\dtrains$ datasets.}
    \label{fig:sa_bor}
\end{figure}

\paragraph{Performance of transformer-based LMs.} We train the transformer-based LMs on the two datasets ($\dtrainl, \dtrains$) and evaluate their generalization based on the $\dtestcfg$ and $\dtestreg$ test sets. Note that for both datasets, 50k training examples are used. Recall that the two training datasets differ in their diversity (120 types in $\dtrainl$ vs.~12 in $\dtrains$). We use the same experimental setup as discussed in \textsection \ref{sec:expt_set}. 
In Figure \ref{fig:sa_bor_ppl}, we see for the models trained on the low-diversity dataset $\dtrains$ that the model obtains similar negative log-likelihood values on both test sets, implying that the model has no preference for generalizing according to the linear rule or the hierarchical rule. For this dataset, neither the CFG nor the regular grammar were optimal in terms of the posterior probabilities, so we observe that the transformer's learning behavior is consistent with the ``idealized'' setup above. For the models trained on the $\dtrainl$ dataset, however, we see that the model  learns to generalize hierarchically, with the NLL on the $\dtestcfg$ test set being significantly lower  than that on the $\dtestreg$ test set.   

Besides NLL, we also compute the main-verb accuracy, where we check whether the model assigns a higher probability to the main verb agreeing with the correct inflection form than the incorrect one. As we can see in Figure \ref{fig:sa_bor_acc}, the model trained on the $\dtrainl$ dataset obtains close to 100\% accuracy when evaluated on the $\dtestcfg$ test set. However, the model trained on the $\dtrains$ dataset obtains close to 50\% generalization accuracy, again showing no preference for a hierarchical or linear rule (the latter would lead to 0\% accuracy). We also verify that these results are not just a by-product of the choice of hyperparameters, and train transformer models with different layers (2, 4, 6, 8, 12), on the $\dtrains$ dataset, and in none of the cases did we observe the models exhibiting preference for hierarchical generalization (see results in Appendix Figure \ref{fig:bor_ld_diff_layers}).

We also consider a stricter metric: all-verb generalization accuracy which is obtained by checking whether \emph{all} predicted verbs (and not just the main verb) in the sentence have the correct inflection. The reason for considering this metric is that, for the agreement task, 100\% main-verb generalization accuracy can also be obtained without learning the hierarchical rule and simply agreeing with the first noun in the sentence.  Note that the all-verb accuracy is computed by feeding prefixes preceding each verb in the sentence and obtaining the model's predictions. We provide the results with all-verb generalization accuracy in Figure \ref{fig:sa_bor_all_verb}, where we show that a baseline that always selects the verb to agree with the first noun in the sentence obtains 33\% accuracy according to this metric, but as we can see transformers perform substantially better than this baseline, indicating that they do not trivially learn this heuristic to obtain high main-verb accuracy.


\paragraph{Takeaways.} Results from this study indicate that when transformers-based LMs exhibit hierarchical generalization, despite being trained on ambiguous data, under the tested conditions, the hierarchical grammar not only fits the data well but is also simpler compared to the regular grammar with linear agreements. For the cases where this condition does not hold, we observe that the trained transformers exhibit no preference for hierarchical generalization.

\noindent \textbf{Limitations.} David Marr's famous three levels of analysis of an information processing system (e.g. human brain or AI) includes the \textit{computational level} -- where we consider what does an ideal solution to the problem that the system is solving looks like, \textit{algorithmic level} -- what algorithm might solve the problem in question, and \textit{implementation level} -- where we study how the representation and algorithm are implemented physically \cite{marr-1982-vision}. In our work we only focused on the computational level analysis i.e. understanding the behavior of transformer LMs in terms of an ideal (Bayesian) learner. Our results only provide a correlation between transformers generalizing hierarchically and training data being explained more effectively (based on the posterior criterion) by a hierarchical grammar than a regular grammar, showing an agreement with the idealised learner. However, we do not make any claim about the algorithmic and implementation level details in transformer LMs i.e. whether these models internally learn the underlying grammars. There is some evidence in prior work showing transformer LMs to learn the probabilistic grammars when trained on data generated from them, e.g. \citet{AL2023-cfg}, show transformer LMs internally learning the underlying PCFG when trained on data generated from the same. 
While this provides some encouraging support for our Bayesian interpretation, further investigation of the internal mechanisms of transformer LMs under our experimental setup is needed to establish a causal connection, which we leave to explore in future work.  

\section{Related Work}

\paragraph{Language acquisition in humans.} The problem of question formation has been well studied in the linguistics and cognitive science literature, where it has been observed that children from a very young age can produce grammatical output consistent with the hierarchical rule. In particular, the poverty of stimulus argument \citep{Chomsky_1968, Piattelli-Palmarini1980-PIALAL} asserts that children are unlikely to observe sufficient data to rule out order rule during language acquisition, and hence knowledge of the language being phrase structured must be innately specified (\textit{linguistic nativism}). On the other hand, as a critique to the nativist argument, the \textit{empiricist} argument \citep{redington-etal-1998-distributional,lewis2001learnability,reali2004structure} states that distributional and statistical regularities in the data can be used to explain why children choose a hierarchical rule over an order-based rule. A critique of the empiricist argument is that it ignores the hierarchical phrase structured nature of natural language \citep{prefors2006poverty} \nascomment{add cite}. Works like \cite{prefors2006poverty, PERFORS2011306} address this critique to empiricist argument using a Bayesian perspective on grammar induction and show that given child-directed corpus, an ideal learner can infer that a hierarchical grammar is simpler and fits the data as well as a linear grammar, without having this knowledge specified innately. 

\paragraph{Hierarchical generalization in neural networks.} Studying hierarchical generalization in neural networks has had its roots in empiricist arguments for language acquisition. \citet{lewis2001learnability} showed that a simple recurrent network   language model trained on the CHILDES dataset \citep{childes2000} (designed  for studying language of and directed to young children), would assign higher probabilities to questions constructed using the hierarchical rule than the order rule. A critique of the above work has been that it doesn't model the relation between the declarative and the question, hence failing to fully address the original poverty of stimulus argument. \cite{frank2007transformational} trained simple recurrent networks on the transformational task (form a question from the declarative) and found some evidence of the networks generalizing hierarchically, though the performance was found to depend heavily on the auxiliaries. 

\citet{McCoy2018RevisitingTP} used the setup from \cite{frank2007transformational} and performed a more thorough study on hierarchical generalization in RNNs (trained as seq2seq models), finding that while these models exhibit limited generalization performance, using attention and training on data with additional syntactic cues can help improve the performance. \citet{mccoy-etal-2020-syntax} studied the architectural inductive biases in RNNs influencing hierarchical generalization, and found that only using a tree-structured model would consistently lead to hierarchical bias. \citet{petty_transformers_2021, mueller-etal-2022-coloring} corroborated these findings for transformers,  finding networks to generalize linearly instead of hierarchically. In contrast to these findings, recently \cite{murty-etal-2023-grokking}, showed that transformers, when trained for longer duration -- way beyond saturating in-distribution performance -- started exhibiting hierarchical generalization. 

While all of these works train neural network models from scratch, recently there has been work on understanding hierarchical generalization in transformer models pretrained on large amounts of naturalistic language data. \cite{mueller-linzen-2023-plant} found that pretraining  encoder-decoder transformers on corpora like Wikipedia or CHILDES results in hierarchical bias in these models, though training on CHILDES  \nascomment{meaning childes?  confusing} was found to be orders of magnitude more sample-efficient towards imparting this bias. \cite{mueller_-context_2023} studied hierarchical generalization during in-context learning in language models, finding large variance in performance across different models. They found this variance to be explained by the composition of training data and particularly found the models trained on code to generalize better.

\paragraph{Grokking.} One puzzle in deep learning generalization is the phenomenon of ``grokking,'' where neural network are observed to start generalizing long after having overfit the training data \citep{Power2022GrokkingGB}. Numerous efforts have been made to understand  grokking and why it occurs. \citet{millidge_grokking_nodate} conjecture that for overparameterized networks the optimal set (i.e., the set of all parameter values resulting in 0 training loss) corresponds to a manifold in parameter space and stochastic gradient descent  essentially acts as a random walk in this manifold, eventually hitting the parameters that generalize. The other explanations rely on \textit{simplicity bias}, hypothesizing that the solutions that generalize are simpler but slower to learn \citep{shah_[159]:_nodate, nanda_progress_2023, bhattamishra-etal-2023-simplicity, varma_explaining_2023}. \citet{thilak2022slingshot} explain grokking from an optimization standpoint and show it to happen at the onset of a phenomenon they call as ``slingshot mechanism,'' identified by spikes in the training loss which result in increased norm of the final-layer weights.
\cite{liu-etal-2022-understanding} attempt to explain grokking through the theory of representation learning, identifying four phases during training and grokking occurring in a "Goldilocks zone" between two of these phases.

\paragraph{Training dynamics and subnetwork generalization.} \citet{merrill2023a}, identify dense and sparse subnetworks in the transformer models trained on a sparse-parity task and found the model starting to generalize as the norm of the sparse subnetwork undergoes rapid norm growth. \citet{chen2024sudden} identify emergence of syntactic attention structure in transformer masked language models, \nascomment{incomplete sentence here} resulting from sudden drops in the loss, leading to the model subsequently acquiring different linguistic capabilities. In concurrent work, \citet{bhaskar2024heuristic} find, using pruning, and for BERT-based models finetuned on NLP tasks like natural language inference and paraphrase identification, the existence of subnetworks that exhibit same in-domain performance but very different out-of-distribution generalization performance. This finding is in line with our observations about the presence of subnetworks consistent with different generalization behaviors. However, due to the nature of our problem, we are further able to show what specific behaviors these subnetworks associate with, how each of these evolves over the course of  training, and suggest why these subnetworks co-exist during training.

\section{Conclusion}

We showed that language modeling training objective can act as a source of inductive bias towards hierarchical generalization, by comparing different training objectives on \numds{} tasks and  finding the LM objective to be the only one that consistently generalizes hierarchically across all of them. We also find that when the training data is consistent with two rules, we can find subnetworks in the transformer LM trained on this data corresponding to each of these rules, which continue to coexist over the course of training. Finally, we provided a Bayesian interpretation to explain why transformer LMs generalize hierarchically:  hierarchical grammars that fit sufficient diverse language data as well as regular grammars are often, in a sense, simpler.

There are multiple directions that can be explored in the future. While our results indicate language modeling as a source of hierarchical bias, it still remains unclear why hierarchical generalization is delayed. 
Further, \citet{murty-etal-2023-grokking} showed that deeper transformer LMs often fail to generalize hierarchically, which remains unexplored in our setting. While the experiments concerning our Bayesian interpretation only involved the simple agreement tasks for which it was possible to construct CFGs, in future it would be interesting to explore methods to model the simplicity and goodness of fit for competing hypotheses for tasks involving transformation of an input sentence to output sentence. In our work, we used the Bayesian interpretation to understand hierarchical generalization in transformers. However, the Bayesian interpretation has been useful to study other forms of generalization in humans as well, including (among others) word learning \citep{xu2007word}, concept learning \citep{goodman2008rational,lake-etal-2015-human}, pragmatics \citep{frank-goodman-2012-predicting}, and theory of mind \citep{baker2011bayesian} \kabir{try finding papers from research groups other than tenenbaum's as well (e.g. Princeton cognitive science group)}, and these capabilities have also been observed to some extent in transformer based LMs as well \cite{patel2023magnifico, hu-etal-2023-fine, shapira2023clever} \kabir{cite some papers}. How well these interpretations can be applied to explain such capabilities in transformers is another potentially interesting direction.

\bibliographystyle{plainnat}
\bibliography{references}  
\newpage
\appendix
\section{Appendix}
\subsection{Training Objectives and Hierarchical Generalization}
\subsubsection{Details about training objectives}
\label{sec:obj_full_details}
Here we detail the input-output structure for all objectives concerning the \numds{} tasks that we study.

\paragraph{Language modeling.}  As discussed in the main text, for the question formation task we simply consider the sequence $s$ as declarative-question pair (or declarative-declarative pair for copy task), e.g., $s = \{\text{my}, \text{walrus}, \cdots, \text{\texttt{quest}}, \text{does}, \cdots, \text{move}, \text{?}\}$. Similarly, for passivization it is the active-passive sentence pair (or active-active); for tense reinflection it is the pair of past and present tense sentence (or past-past), and for simple agreement it is simply the single input sentence.

\paragraph{Sequence-to-sequence modeling and Prefix language modeling.} The inputs for the two objectives are the declarative sentence (or active sentence for passivization and past tense sentence for tense reinflection) and the outputs sequences are the corresponding questions (or passive sentence/present tense sentence depending on the task). Note that all four tasks allow identity pairs, hence the outputs can be the same as the inputs when \texttt{decl} token is provided at the end of the input.

\paragraph{Sequence classification.} For question formation, the input is the declarative sentence, and the output is the four possible auxiliary tokens, $\{\mathrm{do}, \mathrm{does}, \mathrm{don't}, \mathrm{doesn't} \}$ for English and $\{\text{können}, \mathrm{kann}, \mathrm{haben}, \mathrm{hat} \}$ for German. For passivization task, the input is the sentence in active voice and the output is the subject of the passive sentence, which can be any of the 26 nouns in the datasets vocabulary. For tense reinflection, the input is the sentence in past tense and the output is the present tense form of the main-verb in the input sentence (18 classes corresponding to the verbs in dataset). For simple agreement, the input is the sequence of tokens until the main verb and predict the main-verb  as a multi-label (across vocabulary of 18 verbs) classification task. The classification head for all tasks excluding tense reinflection, is attached to the last token in the sequence. For tense reinflection it is attached to the main-verb in the input sentence as otherwise the linear-rule which uses the noun most recent to the main-verb might not be appropriate. We also use causal mask for all tasks, as we found the models to perform better on in-distribution test set in our initial experiments when using it. Also, note that due to the nature of the objective, identity pairs are not supported.

\paragraph{Cloze completion.} For the question formation task, we consider the declarative-interrogative pair and mask out tokens in the interrogative sentence at all positions where the auxiliaries could be present. Specifically, we have mask tokens where i) the auxiliary is present in the interrogative sentence or ii) the auxiliary was present in the original declarative sentence. The model is trained to predict the correct auxiliary at the right positions and \texttt{$<$EMPTY$>$} if an auxiliary is not present at a particular position. Similarly, for tense reinflection, we consider the past-present sentence pair, mask out all the verbs in the present tense sentence and train the model to predict the right form of the verbs. In the simple agreement task, we consider only the present tense sentence, mask out all the verbs and train the model to predict them. Here also we found using causal mask helps in better in-distribution performance and hence use it in all our experiments.

\subsubsection{Training curves}
\begin{figure}[htbp]
    \centering
    \begin{subfigure}{0.33\textwidth}
    \centering
    \includegraphics[width=0.99\textwidth]{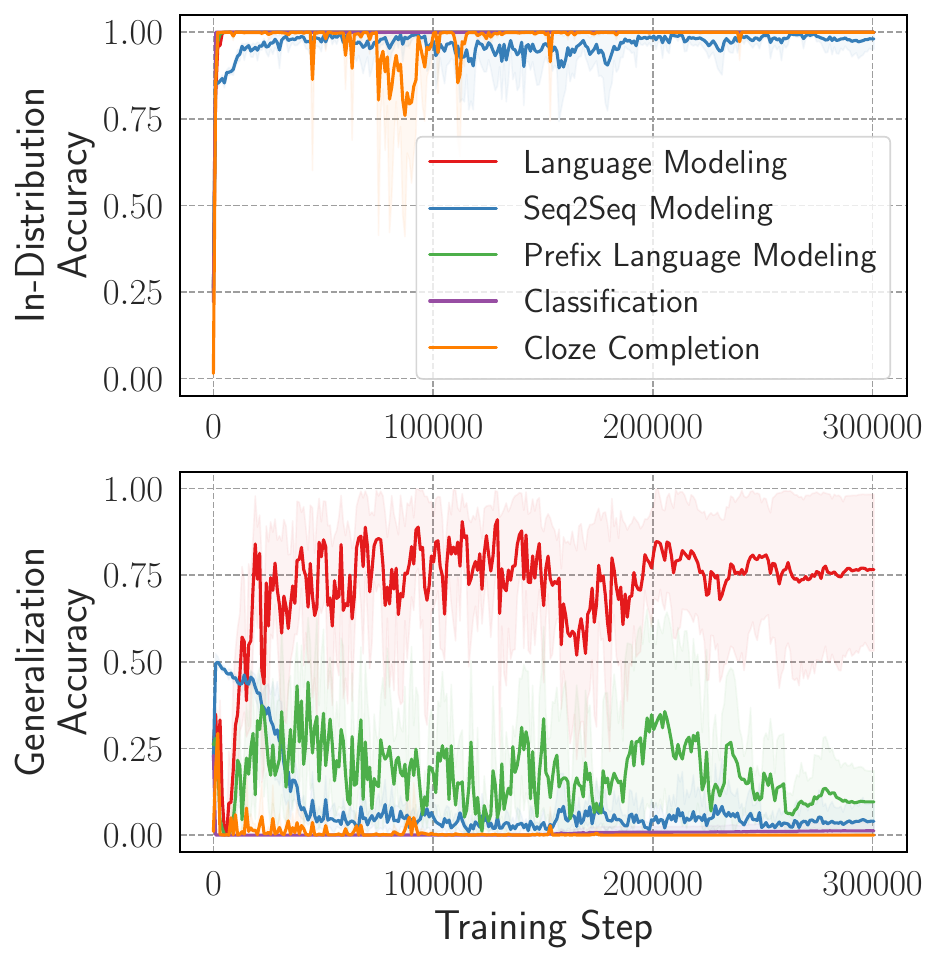}
    \caption{Question formation}
    \label{fig:qf_obj}
    \end{subfigure}\hfill
    \begin{subfigure}{0.33\textwidth}
    \centering
    \includegraphics[width=0.99\textwidth]{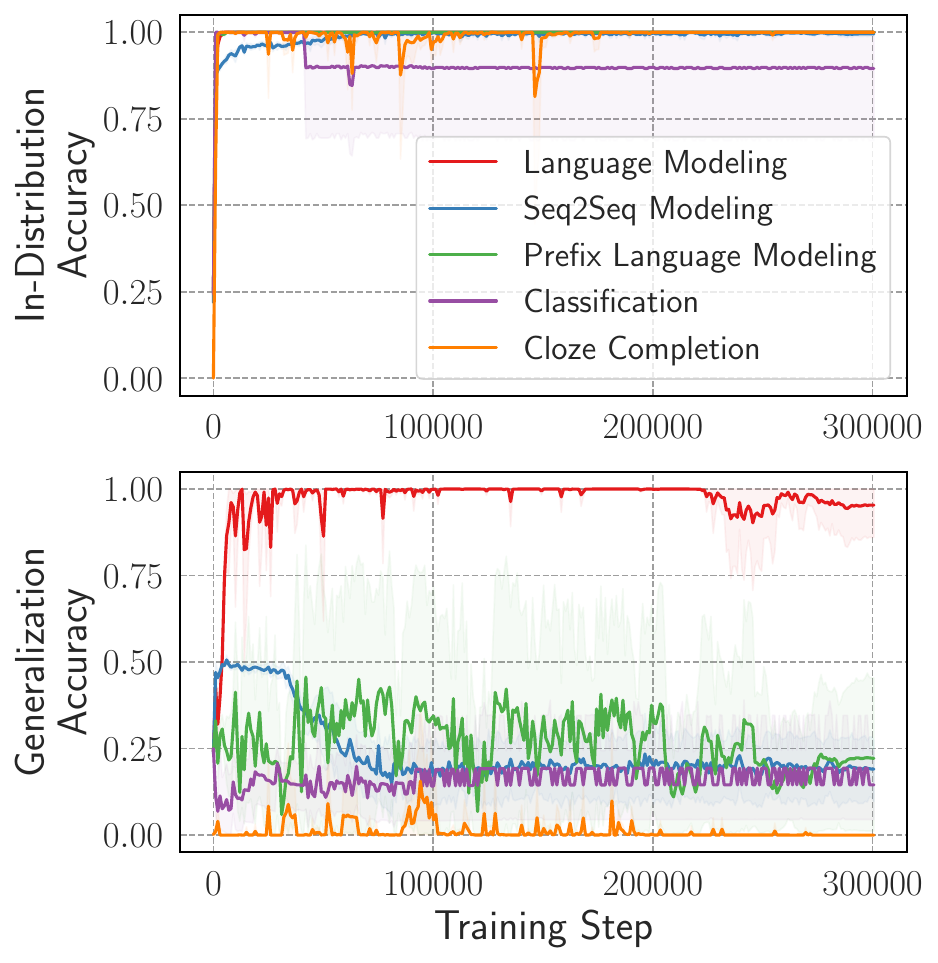}
    \caption{Question formation (German)}
    \label{fig:qf_de_obj}
    \end{subfigure}\hfill
    \begin{subfigure}{0.33\textwidth}
    \centering
    \includegraphics[width=0.99\textwidth]{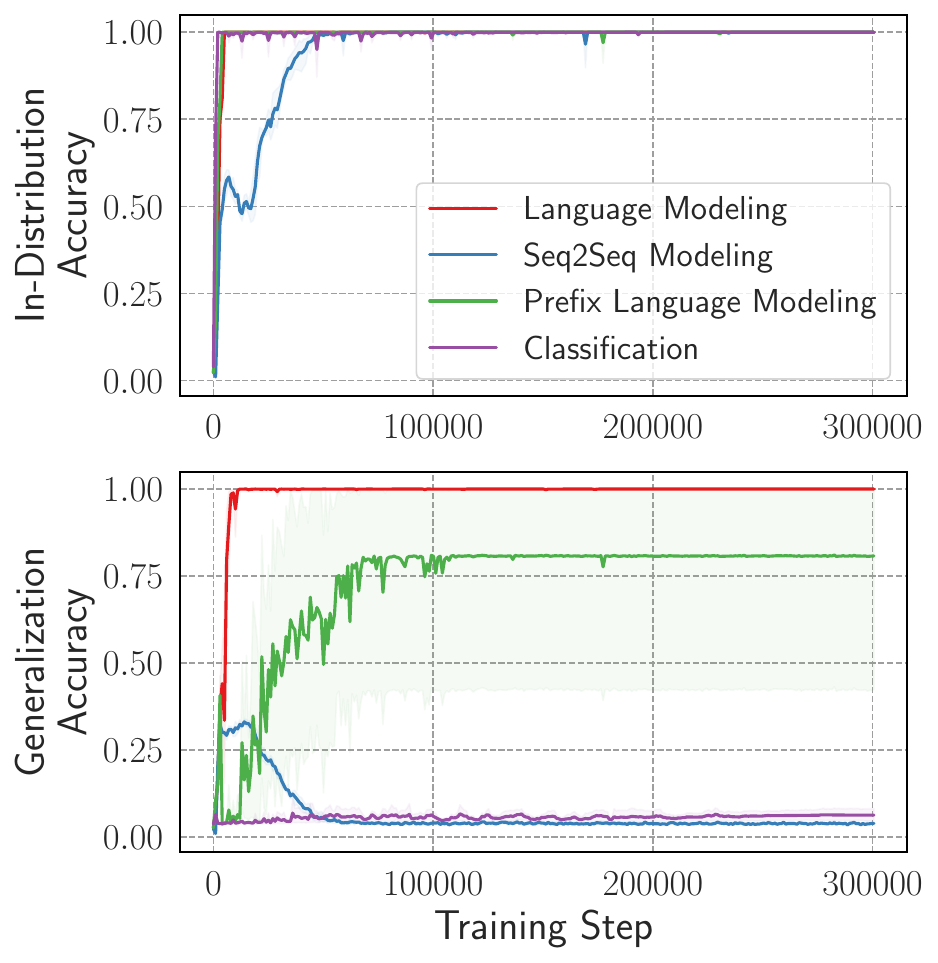}
    \caption{Passivization}
    \label{fig:passiv_obj}
    \end{subfigure}%

    \begin{subfigure}{0.33\textwidth}
    \centering
    \includegraphics[width=0.99\textwidth]{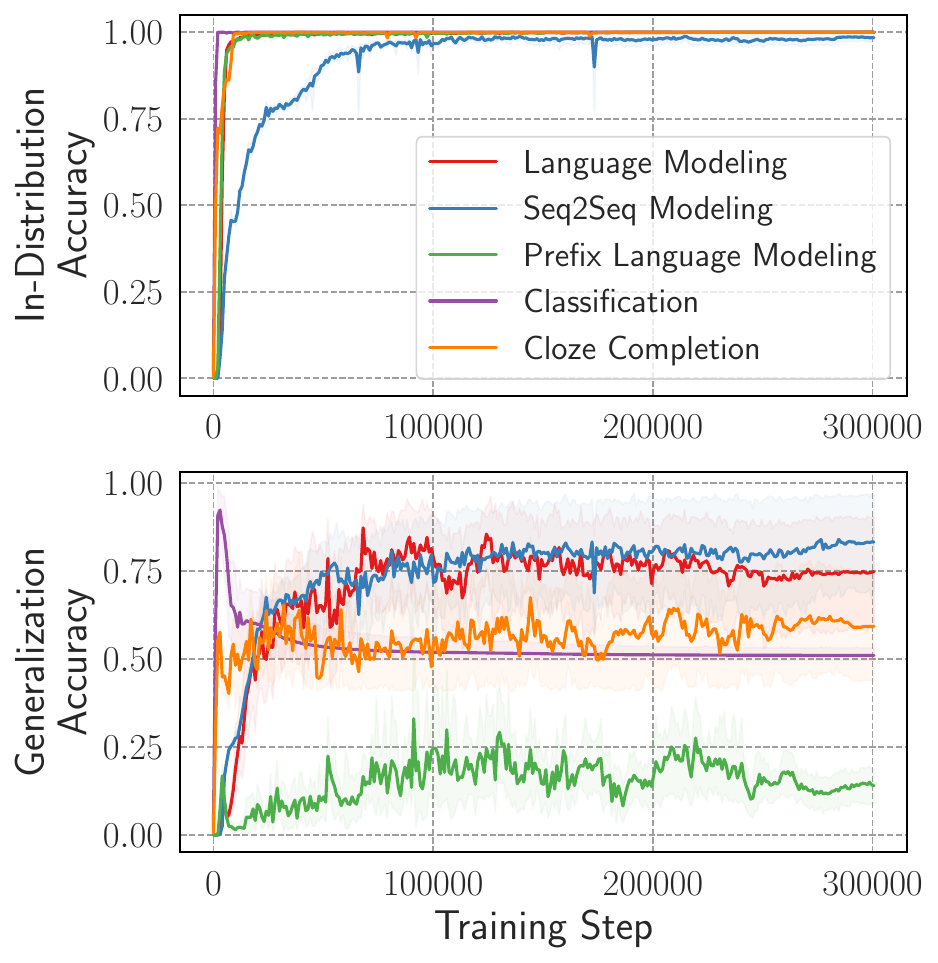}
    \caption{Tense reinflection}
    \label{fig:tense_obj}
    \end{subfigure}
    \begin{subfigure}{0.33\textwidth}
    \centering
    \includegraphics[width=0.99\textwidth]{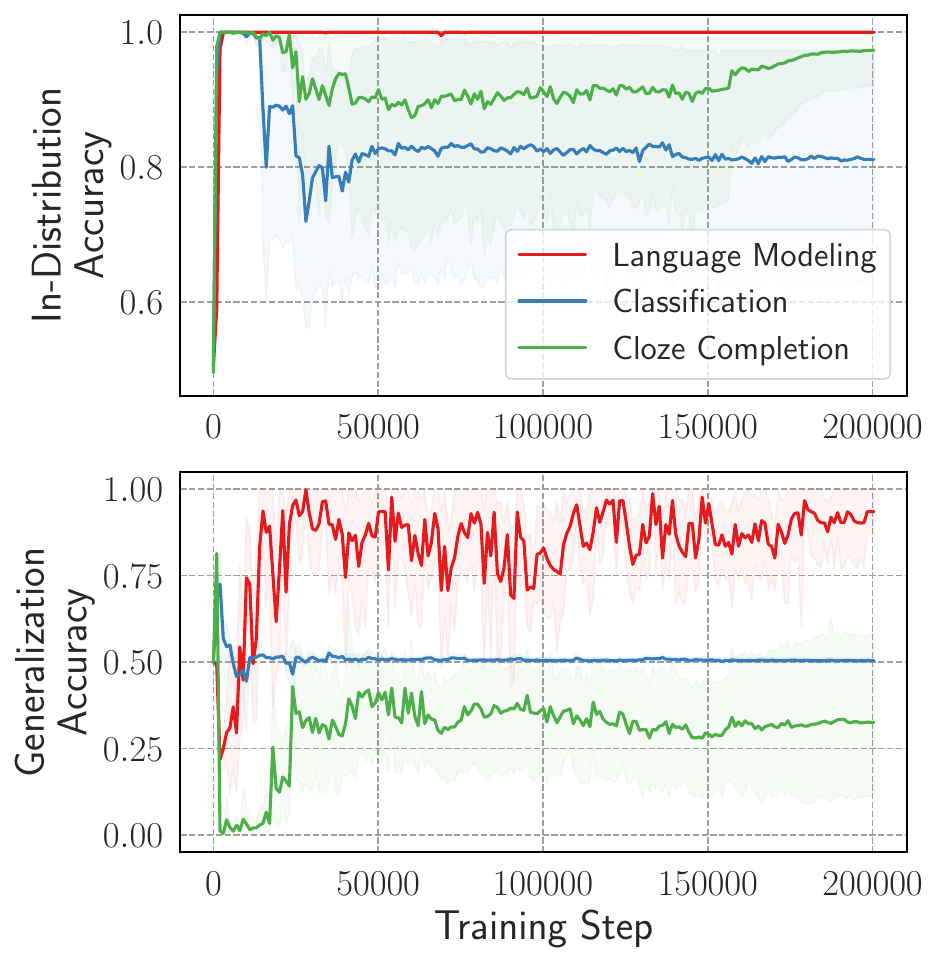}
    \caption{Simple agreement}
    \label{fig:sa_obj}
    \end{subfigure}\hfill
    
    \caption{ Training curves for different objectives on the \numds{} tasks.
    \vidhisha{I think you can put the legend at the top common to all graphs rather than repeat in each graph.}}
    \label{fig:obj_tc_results}
\end{figure}

The performance of the \numobjs{} objectives on the \numds{} datasets across model training is provided in Figure \ref{fig:obj_tc_results}.

\begin{figure}[!htb]

     \centering

     \begin{subfigure}{0.45\linewidth}
    \centering
    \includegraphics[width=0.99\textwidth]{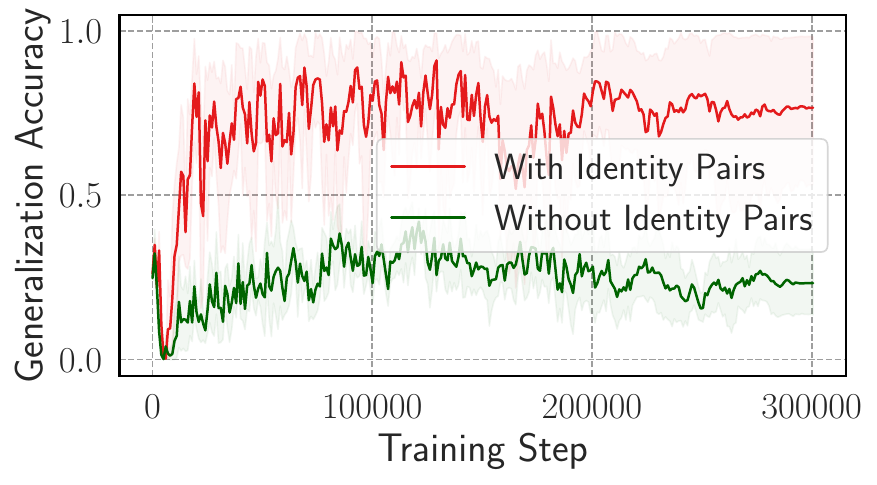}
    \caption{Question formation}
    \label{fig:qf_id_eff}
    \end{subfigure}\hfill
    \begin{subfigure}{0.45\linewidth}
    \centering
    \includegraphics[width=0.99\textwidth]{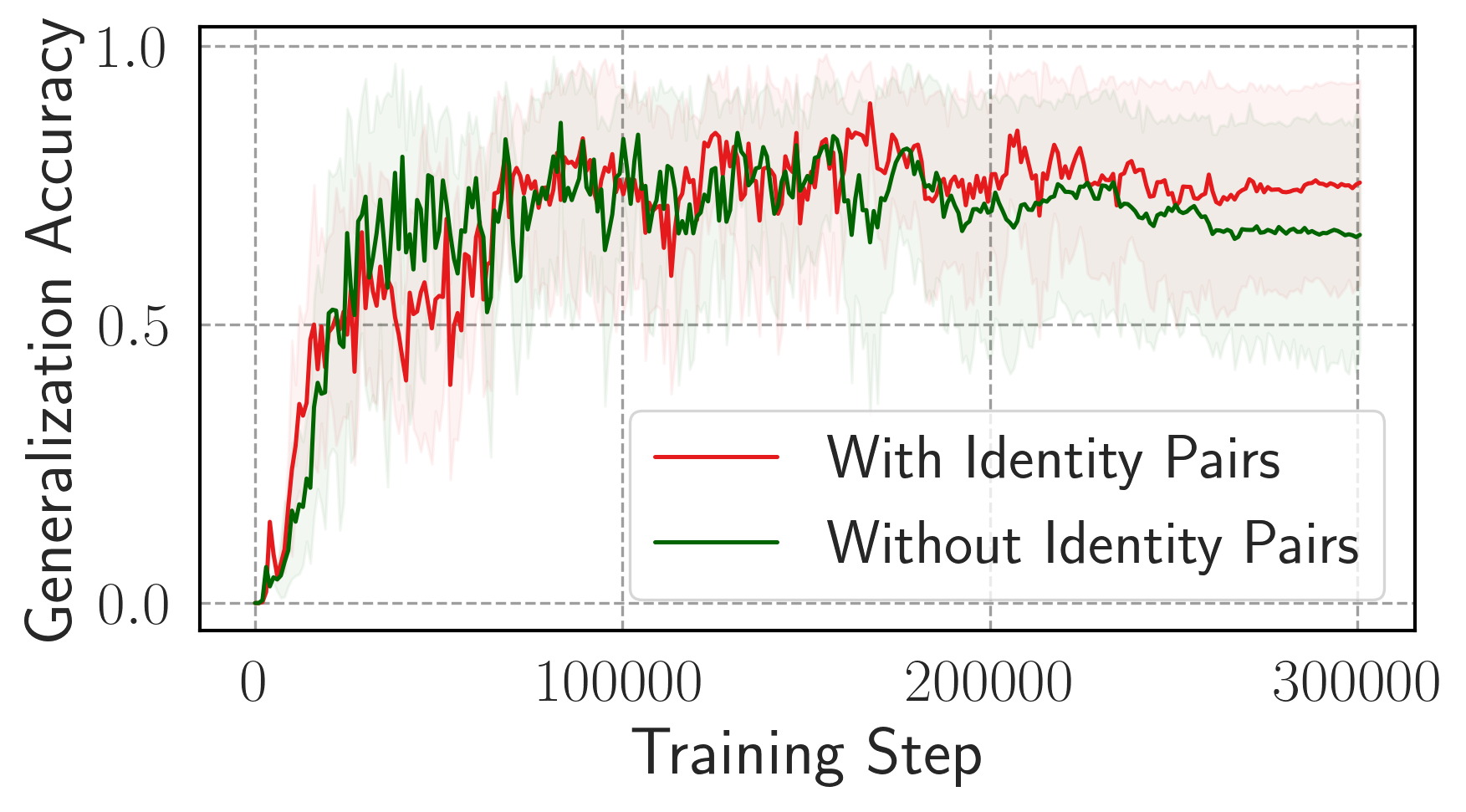}
    \caption{Tense reinflection}
    \label{fig:tense_id_eff}
    \end{subfigure}\hfill
    \caption{Effect of the presence of identity pairs (the auxiliary copy task) in training data on hierarchical generalization.}
    \label{fig:id_eff}

\end{figure}

\paragraph{The effect of identity pairs on hierarchical generalization.} As noted in \textsection \ref{sec:objectives}, the training datasets for tasks like the question formation and tense reinflection tasks from \cite{mccoy-etal-2020-syntax}, also used by \cite{murty-etal-2023-grokking}, include identity pairs of declaratives (for question formation) and past tense sentences (for tense reinflection) for the auxiliary copy task. This data also includes input sentences with the same syntax as the inputs in the generalization set, though the outputs are still unseen. Since the model is exposed to the form of sentences in the generalization set, we would like to explore if this auxiliary data (identity pairs) contributes to hierarchical generalization: Do models learn hierarchical rule without having seen the input-types from the generalization set? \nascomment{you need to explicitly state your hypothesis here.  it's not at all clear to me what led you to do this experiment}
As can be seen in Figure \ref{fig:id_eff}, for the question formation task removal of identity pairs results in a significant drop in generalization accuracy for transformers trained with language modeling objective. For tense reinflection, however, the generalization performance remains more or less similar. We suspect this might be due to the gap between the sentence types in the in-distribution and generalization set is arguably greater for question formation than tense reinflection. In question formation the generalization set has sentences with relative clauses attached to the subject and without the identity pairs, such sentences will be completely unseen during the training. On the other hand for tense reinflection, the training dataset has all nouns of the same plurality (i.e. either all singular or all plural), while in generalization the nouns have different plurality to break the ambiguity. The overall structure of the sentence does remain same for tense-reinflection, while that's not the case for question formation. \kabir{I think this part can be moved to appendix, taking a lot of space and not very important. Now that I think more about it, without identity pairs, it is probably unreasonable to expect generalization for the question formation task in the first place.}


\subsubsection{Do seq2seq trained transformers really generalize hierarchically for tense reinflection?}
\label{sec:tiseq2seq}

For tense reinflection, in Figure \ref{fig:obj_results} we observe that seq2seq objective performs close to language modeling in terms of generalization accuracy. We believe this might be due to the fact that it is possible to achieve high generalization accuracy on tense reinflection without learning the hierarchical rule. Notice that in both examples for tense reinflection in Table \ref{tab:dataset_examples}, there exists an alternate non-hierarchical rule: The main-verb of the sentence should agree with the first noun in the sentence, since the way \cite{mccoy-etal-2020-syntax}'s dataset is constructed ensures that the first noun in the sentence is always the subject hierarchically connected to the main-verb. To validate if these models truly learn the hierarchical rule and not a shortcut, we evaluate the tree-structuredness of the learned representations of the trained models using the tree projection method of \cite{murty2023characterizing}. Their method characterizes the extent to which computations in the transformer can be explained by an equivalent tree-structured neural network.

\begin{figure}[!htbp]
    \centering

     \begin{subfigure}{0.4\linewidth}
    \centering
    \includegraphics[width=0.99\linewidth]{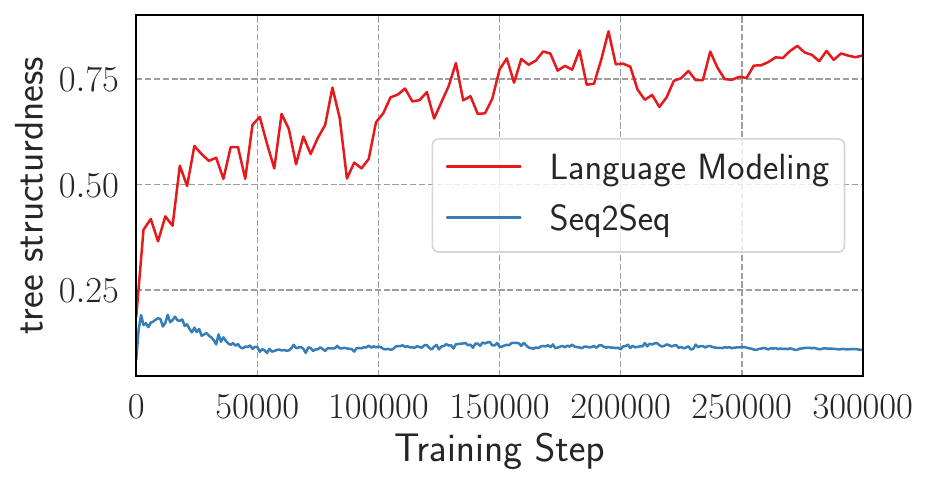}
    \caption{Tree-structuredness scores for models trained on tense reinflection task.}
    \label{fig:ts}
    \end{subfigure}\hfill
     \begin{subfigure}{0.48\linewidth}
    \centering
    \includegraphics[width=0.95\linewidth]{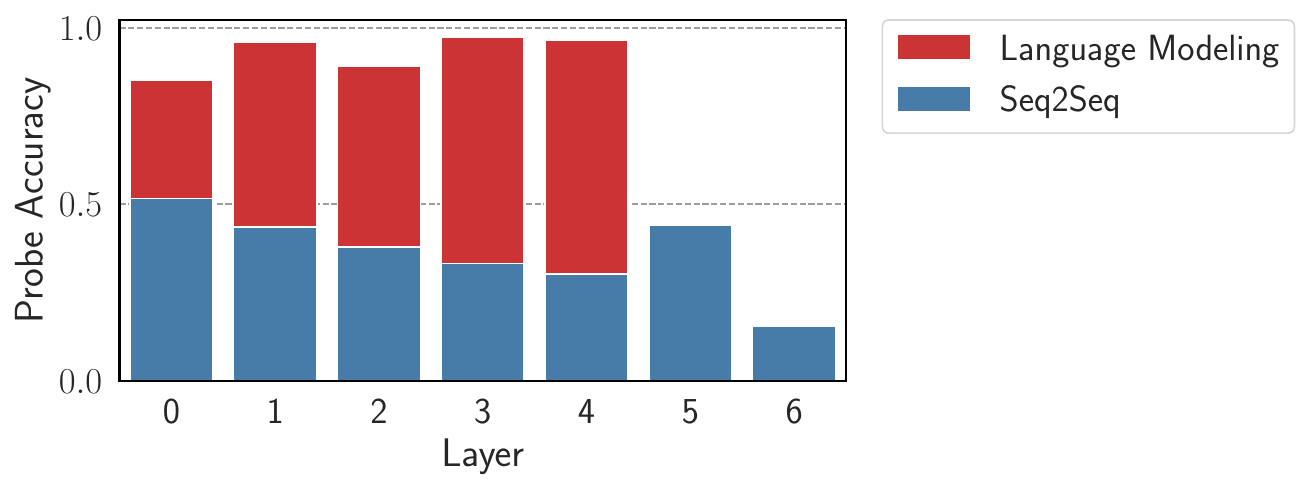}
    \caption{Linear probing on transformers trained for tense reinflection tasks using language modeling and Seq2Seq objectives. The missing red bars for layer 5 and 6 are because the transformer trained with langauge modeling objective only has 4 layers, while the Seq2Seq model has 6.}
    \label{fig:probe_tense}
    \end{subfigure}
    \caption{Investigating if Seq2Seq models really generalize hierarchically for tense reinflection.}
\end{figure}

In Figure \ref{fig:ts}, we compare the tree-structuredness scores for the transformer models trained using language modeling and seq2seq objectives. Consistent with \cite{murty-etal-2023-grokking}, we observe that for the language modeling objective the transformer representations continue to become more tree-structured as training progresses. However, for seq2seq model the tree-structuredness score remains low throughout the training. This indicates that while the seq2seq model obtains high generalization accuracy, it might not actually be generalizing hierarchically. To verify this further, we also performed linear-probing \citep{alain2017understanding} on the final layer representations of the two models, and found for language modeling-trained model, the syntactic categories of the words (e.g. singular noun is the category for \textit{walrus}) in the sentence can be recovered with a very high accuracy. By contrast, linear probes trained on seq2seq model are unable to extract the syntactic categories (50\% probing accuracy compared to 99\% for language model) (see figure \ref{fig:probe_tense}).

\begin{figure*}[!htb]
\begin{subfigure}{0.4\linewidth}
\centering
\includegraphics[width=0.99\textwidth]{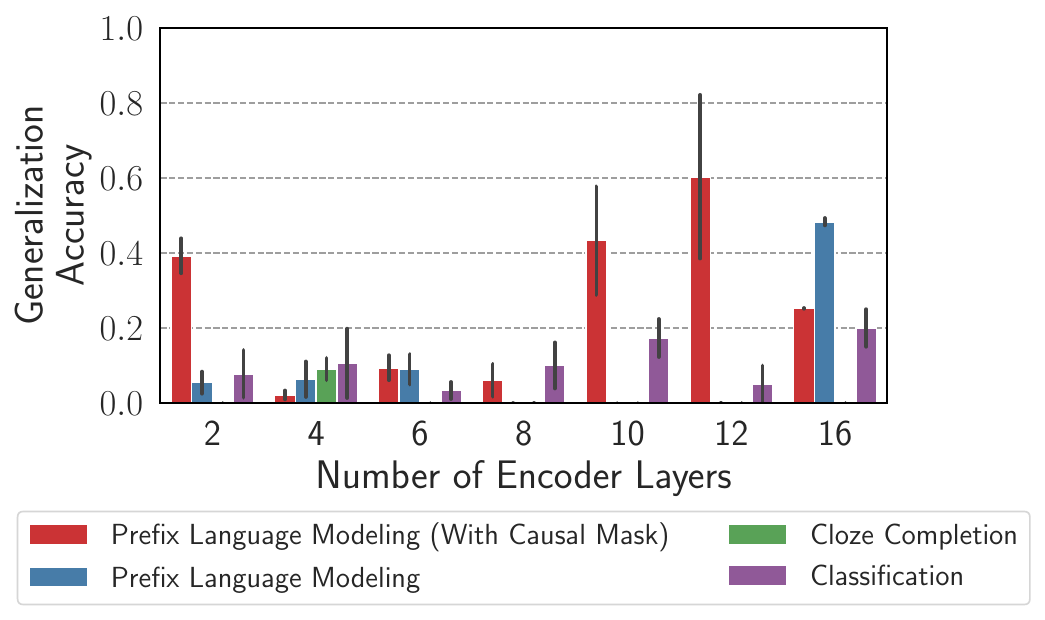}
\caption{Effect of model depth for PrefixLM and Cloze Completion training objectives}
\end{subfigure}\hfill
\begin{subfigure}{0.55\linewidth}
\centering
\includegraphics[width=0.99\textwidth]{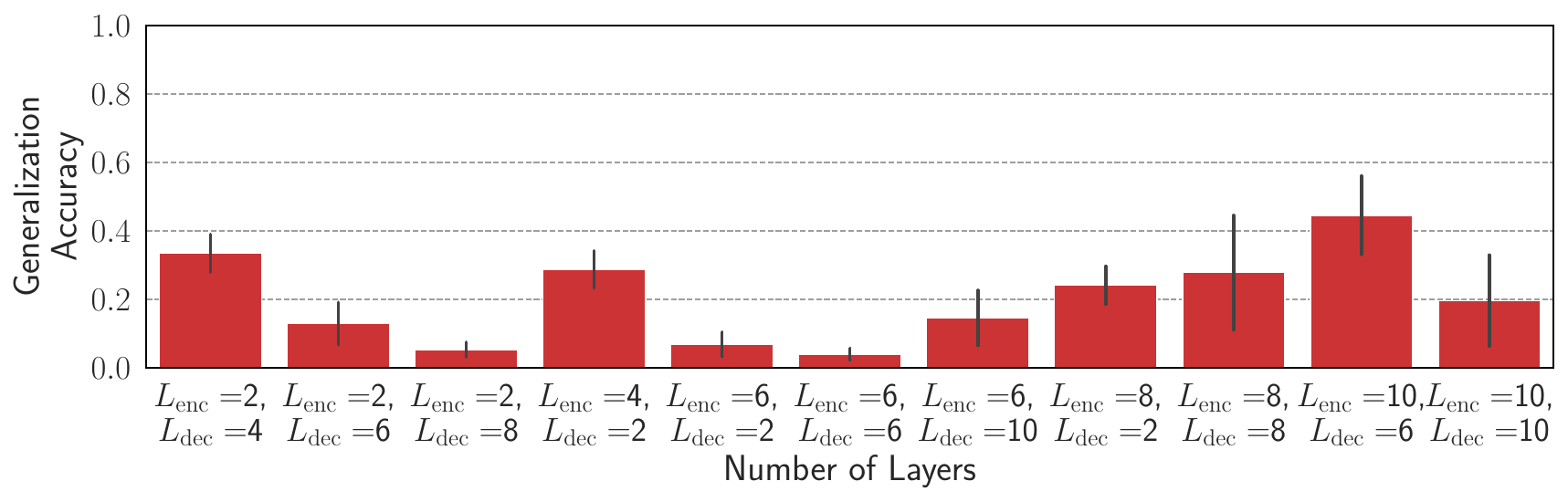}
\caption{Effect of model depth for Seq2Seq training objective.}
\end{subfigure}
\begin{subfigure}{0.45\linewidth}
\centering
\includegraphics[width=0.99\textwidth]{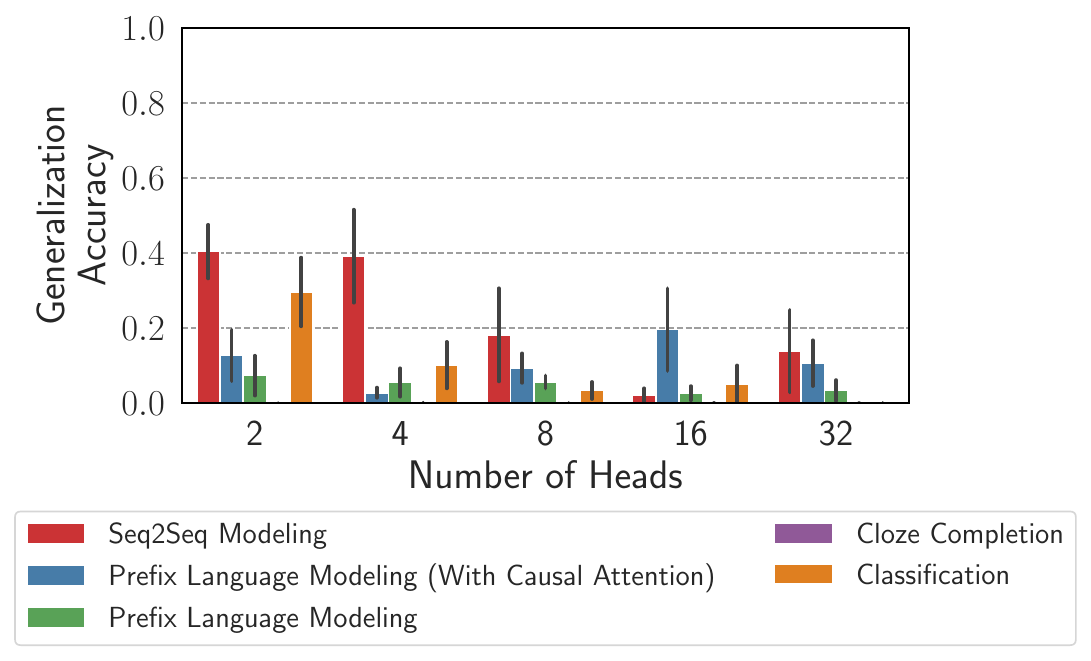}
\caption{Effect of number of heads for different training objectives}
\end{subfigure}\hfill
\begin{subfigure}{0.45\linewidth}
\centering
\includegraphics[width=0.99\textwidth]{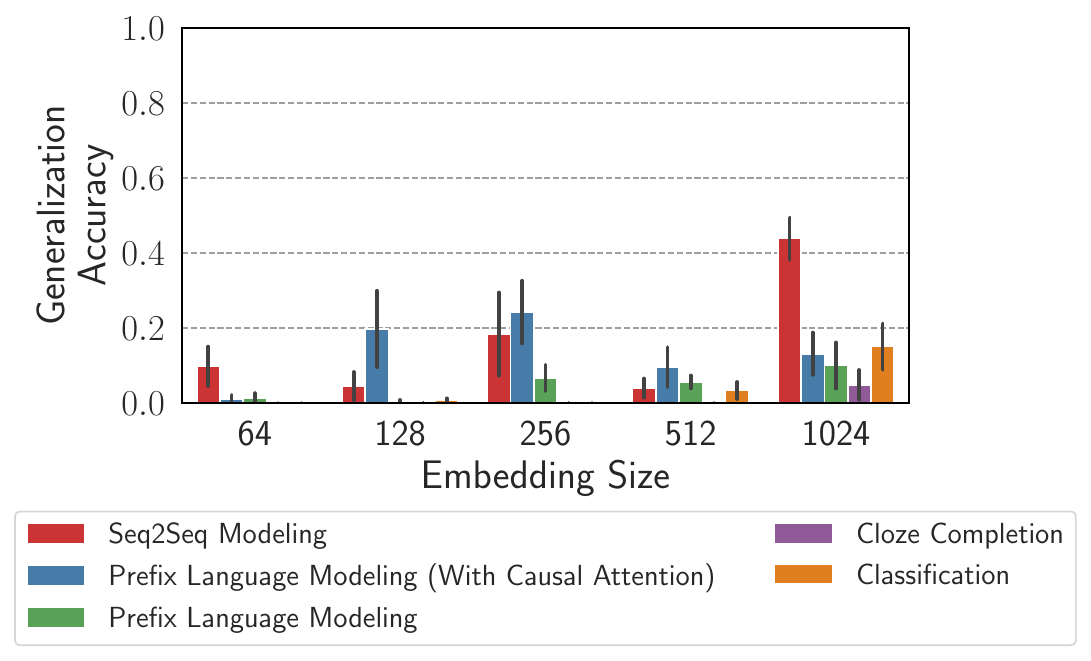}
\caption{Effect of embedding size for different training objectives}
\end{subfigure}
\caption{Robustness of negative results for non-language modeling objectives across different hyperparameter setting for the question formation task. For each experiment, we vary one hyperparameter while keeping the other two fixed to default values i.e. 6 layers, 8 heads, and 512 embedding dimension. Note that the language modeling objective achieves an average generalization accuracy of 0.76 with the default hyperparameters.}
\label{fig:hyperparams}
\end{figure*}

\begin{figure*}[!htb]
\tiny
\begin{subfigure}{0.4\linewidth}
\centering
\includegraphics[width=0.99\textwidth]{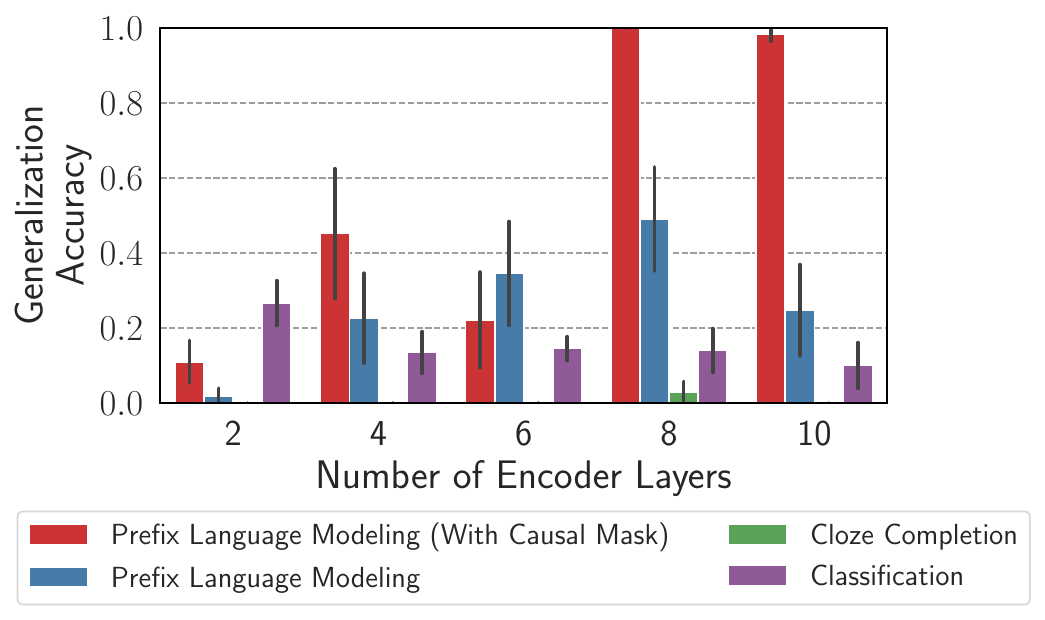}
\caption{Effect of model depth for PrefixLM, Classification, and Cloze Completion training objectives on the question formation German task.}
\end{subfigure}\hfill
\begin{subfigure}{0.55\linewidth}
\centering
\includegraphics[width=0.99\textwidth]{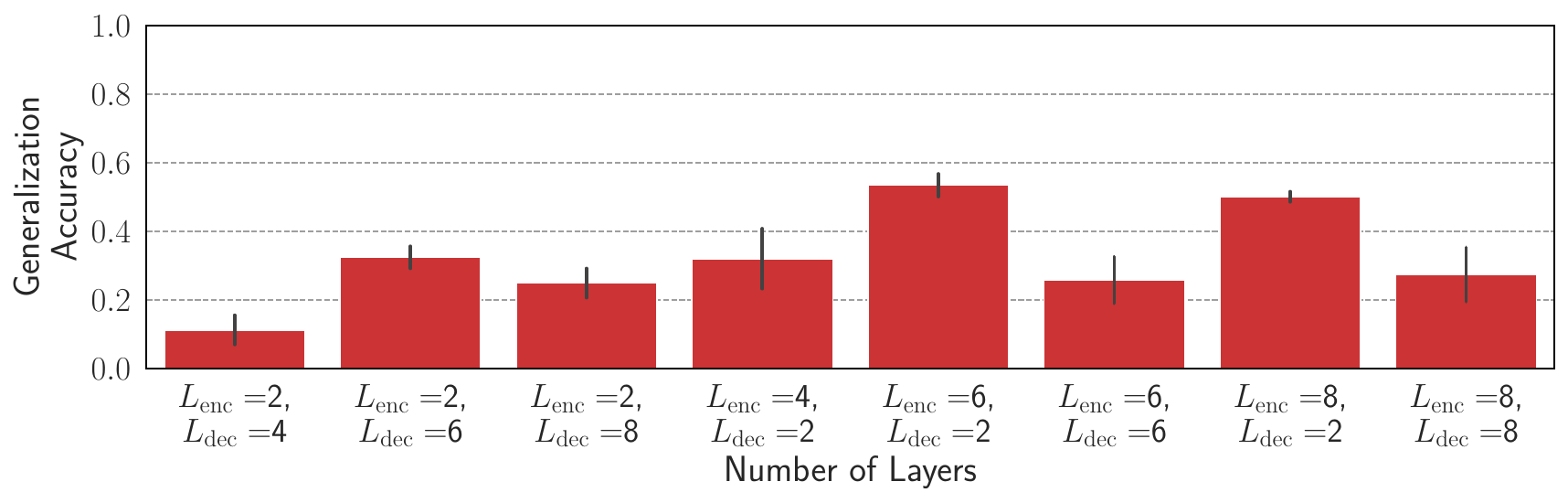}
\caption{Effect of model depth for Seq2Seq training objective on the question formation German task.}
\end{subfigure}

\begin{subfigure}{0.4\linewidth}
\centering
\includegraphics[width=0.99\textwidth]{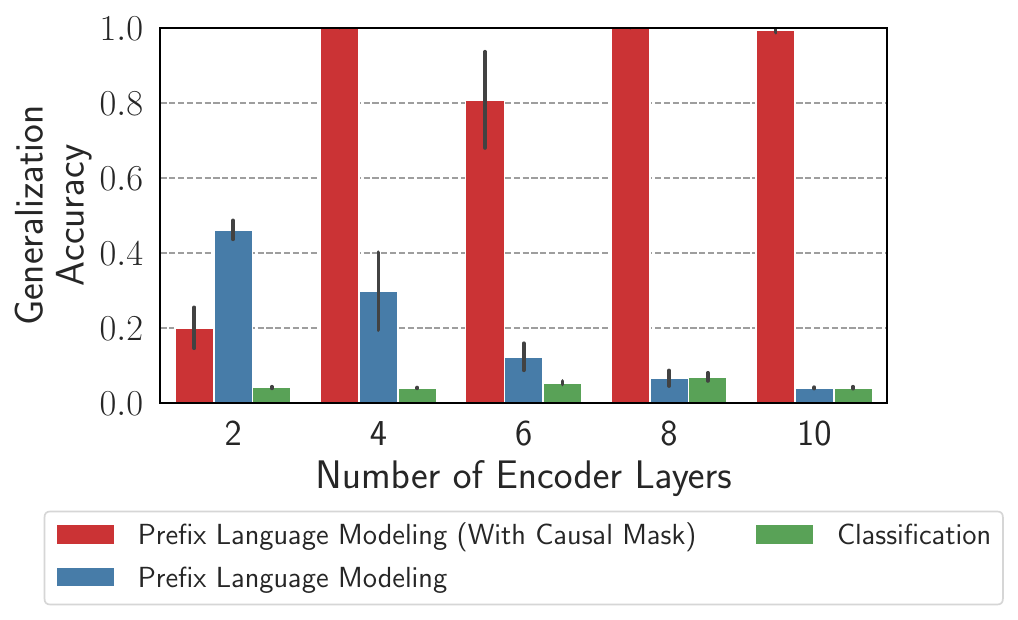}
\caption{Effect of model depth for PrefixLM, Classification, and Cloze Completion training objectives on the passivization task.}
\end{subfigure}\hfill
\begin{subfigure}{0.55\linewidth}
\centering
\includegraphics[width=0.99\textwidth]{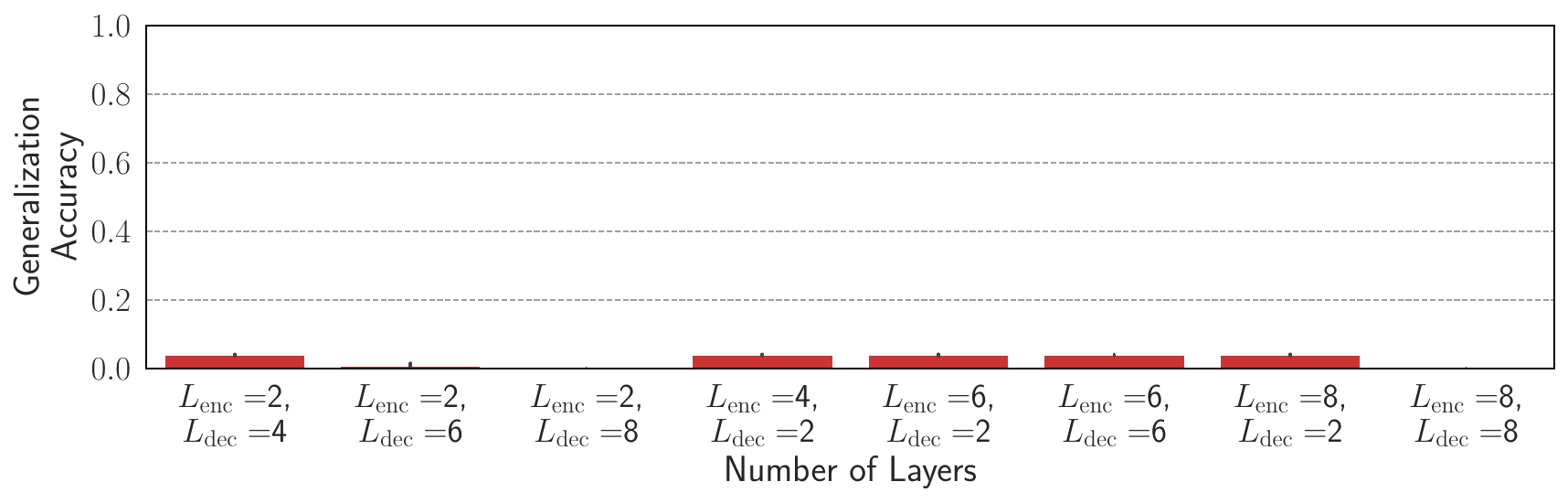}
\caption{Effect of model depth for Seq2Seq training objective on the passivization task.}
\end{subfigure}

\caption{Robustness of negative results for non-language modeling objectives across different model depths for question formation German and passivization tasks. Note that the language modeling objective achieves roughly an average generalization accuracy of 1.0 with the default hyperparameters. We omit plots for heads and embedding dimension in interest of space, but we obtain consistent results with what we observe here.}
\label{fig:hyperparams_qfdepassiv}
\end{figure*}
\subsection{Subnetworks with Different Generalization Behaviors}
\label{sec:subnetworks_tisa}
\paragraph{Tasks other than Question Formation.}

\begin{figure}[!htbp]
    \centering
    \begin{subfigure}{0.45\textwidth}
        \includegraphics[width=0.99\textwidth]{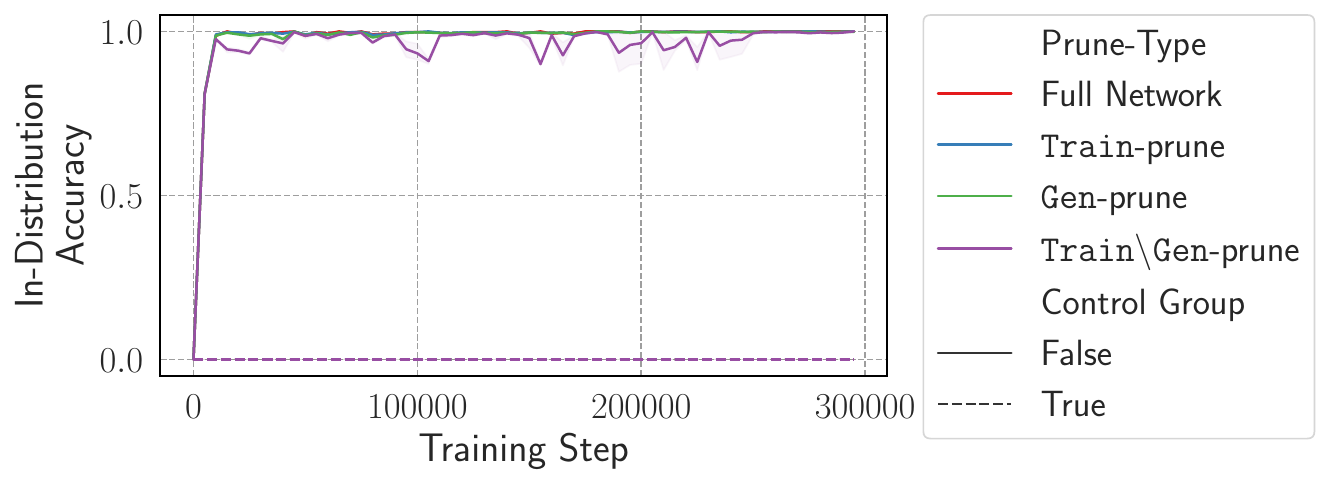}
        \caption{In-distribution accuracy for different pruning methods across training (original question formation training data)}
        \label{fig:tense_dynamics_val}
    \end{subfigure}\hfill
    \begin{subfigure}{0.45\textwidth}
        \includegraphics[width=0.99\textwidth]{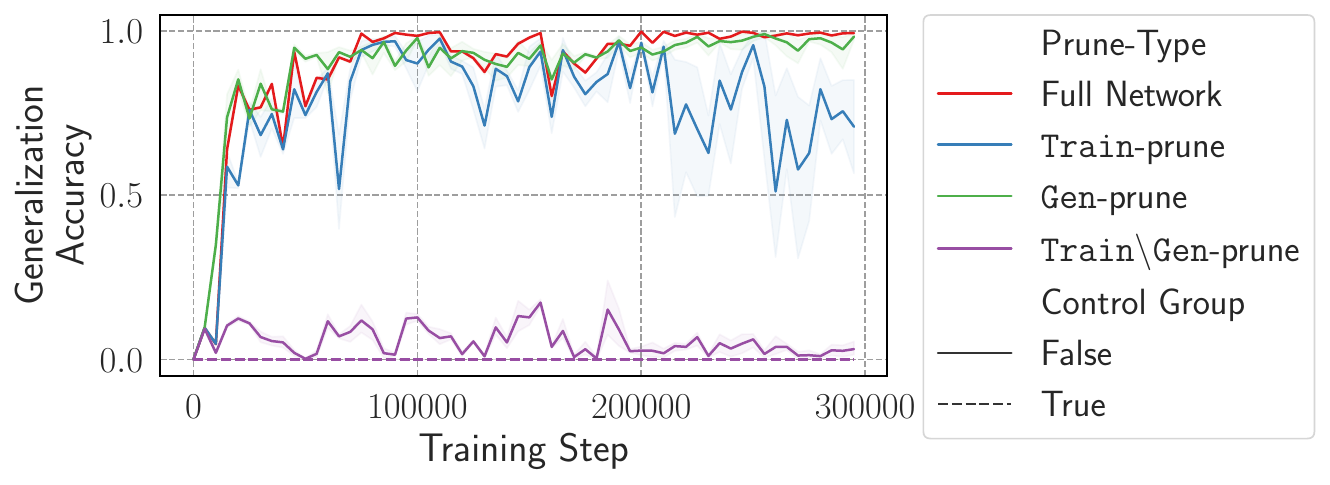}
        \caption{Generalization accuracy for different pruning methods across training (original question formation training data)}
        \label{fig:tense_dynamics_test}
    \end{subfigure}\hfill
    \caption{Tracking training dynamics w.r.t. to the three pruning methods for tense reinflection task}
    \label{fig:tense_dynamics}
\end{figure}

\begin{figure}[!htbp]
    \centering
    \begin{subfigure}{0.45\textwidth}
        \includegraphics[width=0.99\textwidth]{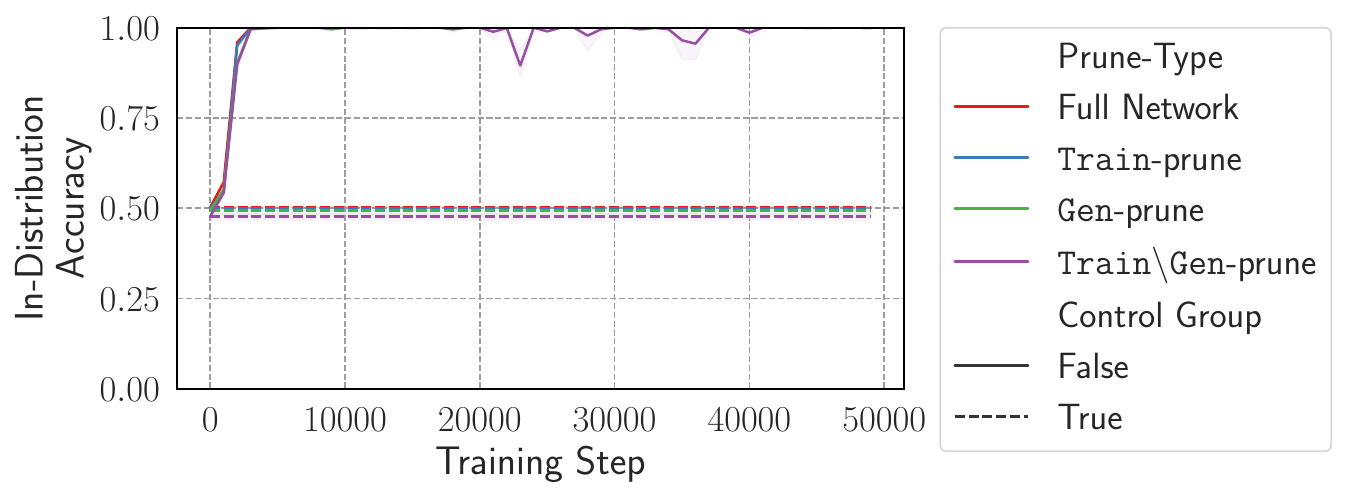}
        \caption{In-distribution accuracy for different pruning methods across training (original question formation training data)}
        \label{fig:tense_dynamics_val}
    \end{subfigure}\hfill
    \begin{subfigure}{0.45\textwidth}
        \includegraphics[width=0.99\textwidth]{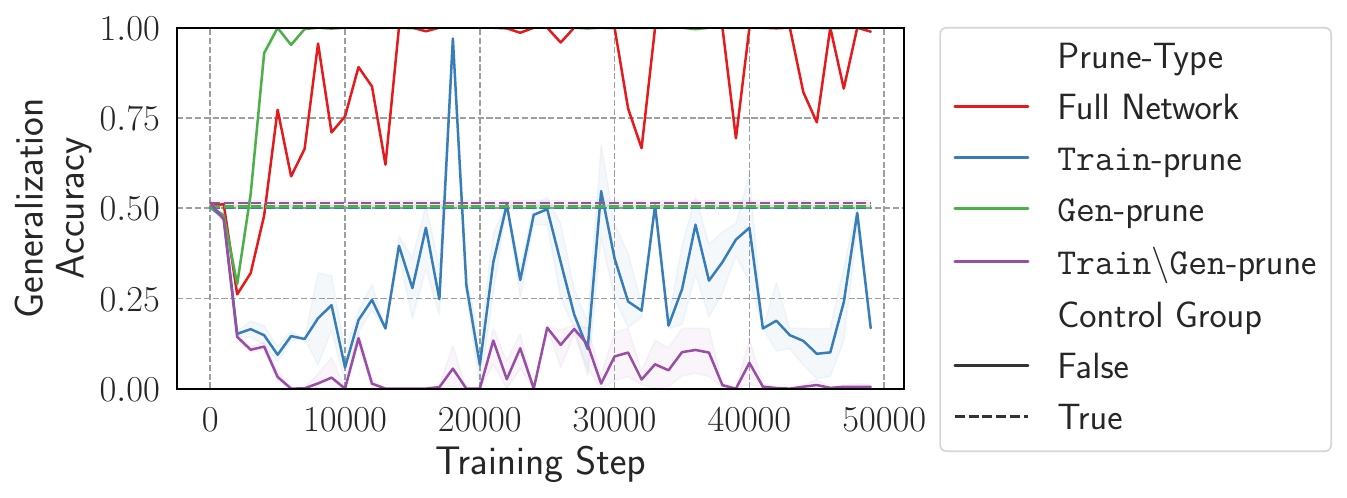}
        \caption{Generalization accuracy for different pruning methods across training (original question formation training data)}
        \label{fig:tense_dynamics_test}
    \end{subfigure}\hfill
    \caption{Tracking training dynamics w.r.t. to the three pruning methods for simple agreement task}
    \label{fig:sa_dynamics}
\end{figure}

In the main paper under \textsection \ref{sec:subnet} our results on the discovery of subnetworks with different generalization performances were performed on question formation task. Here, we provide the results for tense reinflection and simple agreement. For tense-reinflection, we slightly modify the pruning procedure. Since, for tense reinflection, we need to generate the entire present tense sequence to check if the predicted main verb is in the correct form, we compute the loss over all output tokens during pruning unlike question formation, where only the loss on the first auxiliary in the question was computed. Due to this, for \spprune{} training becomes highly unstable as the procedure involves minimizing training and maximizing the test loss. Hence, we propose an alternate \spprune{} procedure for this task, where we generate a ``linear-rule" version of the generalization set, where the sentence pairs are generated in such a way that they are only consistent with the linear-rule. Note that this can be done by simply taking a past tense sentence in the generalization set and flipping the inflection of the main-verb based on the agreement with the most recent noun preceding the verb.  Note that similar to \genprune{}, here also we only use 1\% of the total data from the ``linear-rule" generalization set for pruning to avoid the possibility of overfiting. For simple agreement the procedure remains same as question formation, with the only difference that the loss is computed on the main-verb in this case during pruning instead of the auxiliary. Pruning results for the two tasks are provided in Figures \ref{fig:tense_dynamics} and \ref{fig:sa_dynamics}. We find results consistent with our findings for question formation task here as well, where the ``linear-rule" and ``hierarchical-rule" subnetworks can be found using pruning and continue to co-exist over the course of training.


\begin{figure}
    \centering
    \includegraphics[width=0.6\textwidth]{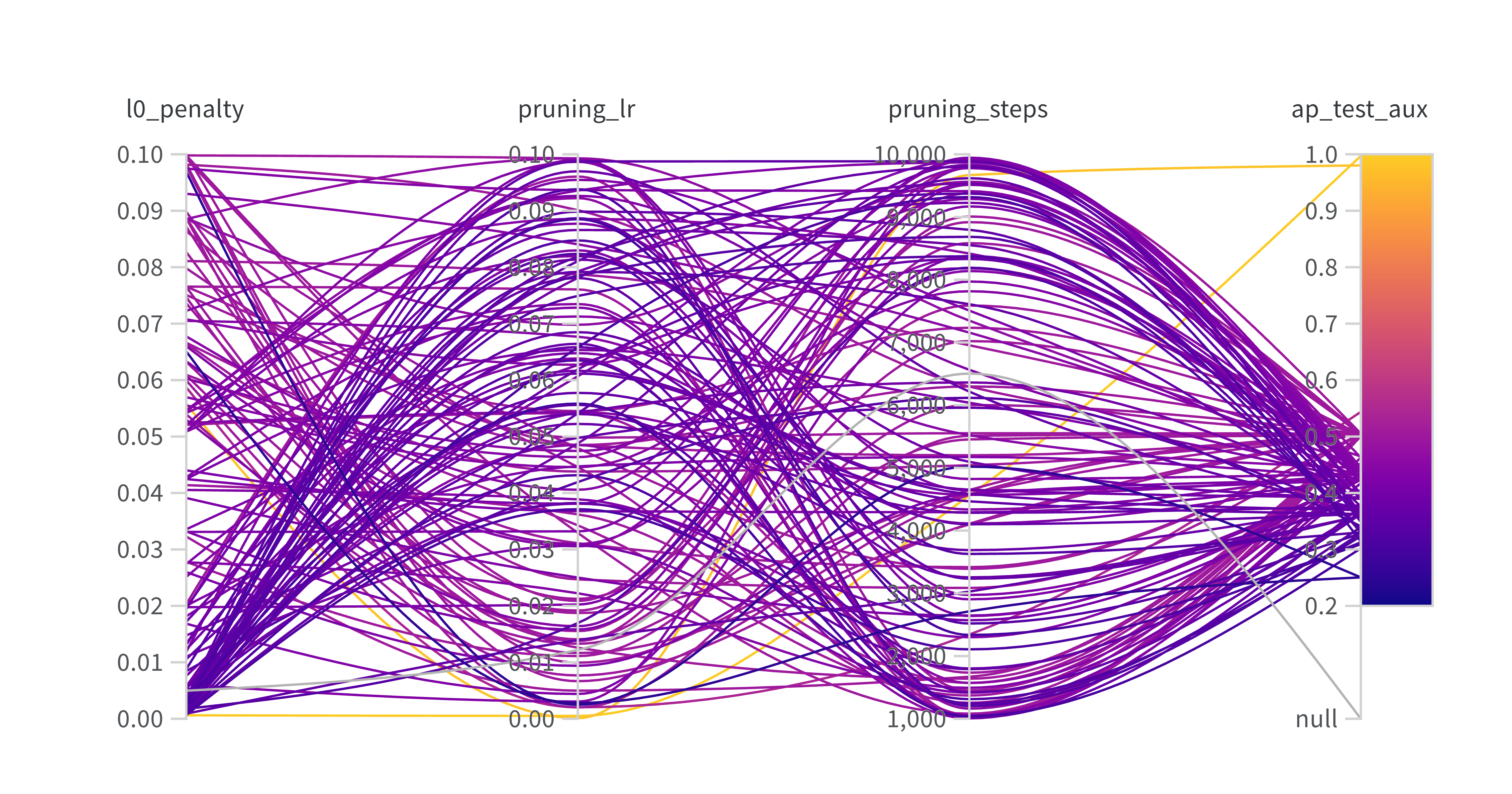}
    \caption{Performing \spprune{} using different pruning hyperparameters, $L_0$ regularization penalty, pruning learning rate, and pruning steps. \texttt{ap\_test\_aux} denotes the generalization accuracy after pruning. We try 128 combinations of these parameters using Bayesian optimization and in no case we find a subnetwork obtaining 0\% generalization accuracy. At best case we find subnetworks with 25\% accuracy which correspond to a random baseline for this task (since there are 4 choices of the auxiliary verbs).}
    \label{fig:spprune_disamb_hyptune}
\end{figure}

\begin{figure}
    \centering
    \includegraphics[width=0.95\textwidth]{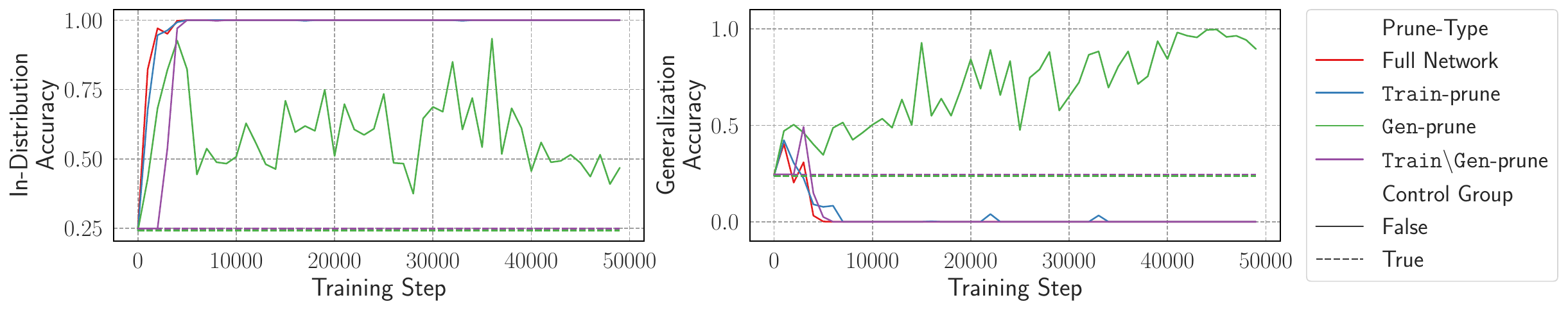}
    \caption{Training dynamics of transformer LM trained with question-formation data which is disambiguated with examples consistent only with the linear rule (by augmenting 10k such examples to the original 100k ambiguous examples). As can be seen, the full network in this case after a few thousand steps plateaus at 0\% generalization performance, which is expected since only the linear rule is applicable to the entire dataset and hence the model is more likely to learn linear generalization in this case. Further, even \genprune{} in this case fails to find subnetworks with 100\% in-distribution as well as 100\% generalization performance. While further during training, \genprune{} does find subnetworks with higher generalization performance, the in-distribution performance at these points is very low, meaning the subnetwork isn't actually consistent with the hierarchical rule.}
    \label{fig:dynamics_disamb_lin}
\end{figure}

\begin{figure*}[!htbp]
    \centering
    \begin{subfigure}{0.32\textwidth}
        \includegraphics[width=0.99\textwidth]{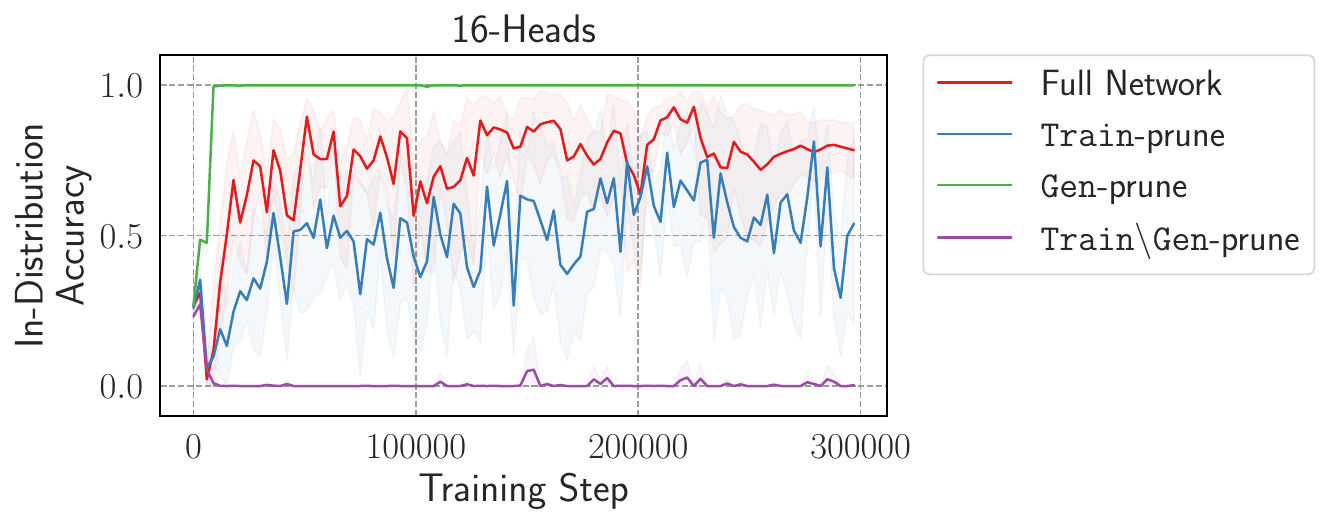}
        \caption{Training dynamics for Transformer LM with 16 heads per attention layer}
        \label{fig:dynamics_16heads}
    \end{subfigure}\hfill
    \begin{subfigure}{0.32\textwidth}
        \includegraphics[width=0.99\textwidth]{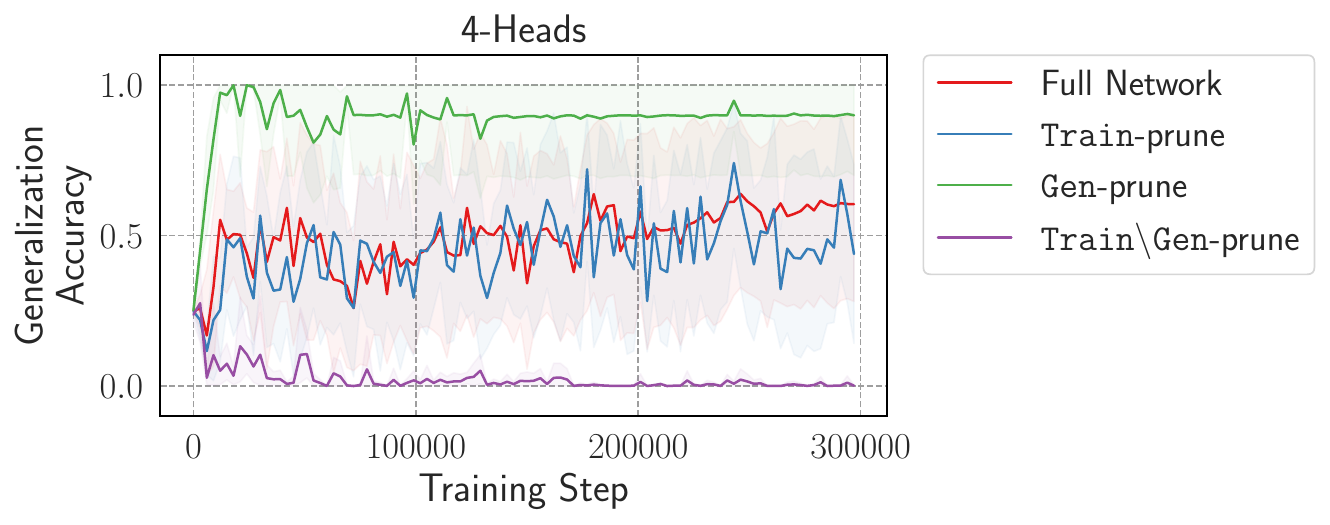}
        \caption{Training dynamics for Transformer LM with 4 heads per attention layer}
        \label{fig:dynamics_4head}
    \end{subfigure}\hfill
    \begin{subfigure}{0.32\textwidth}
        \includegraphics[width=0.99\textwidth]{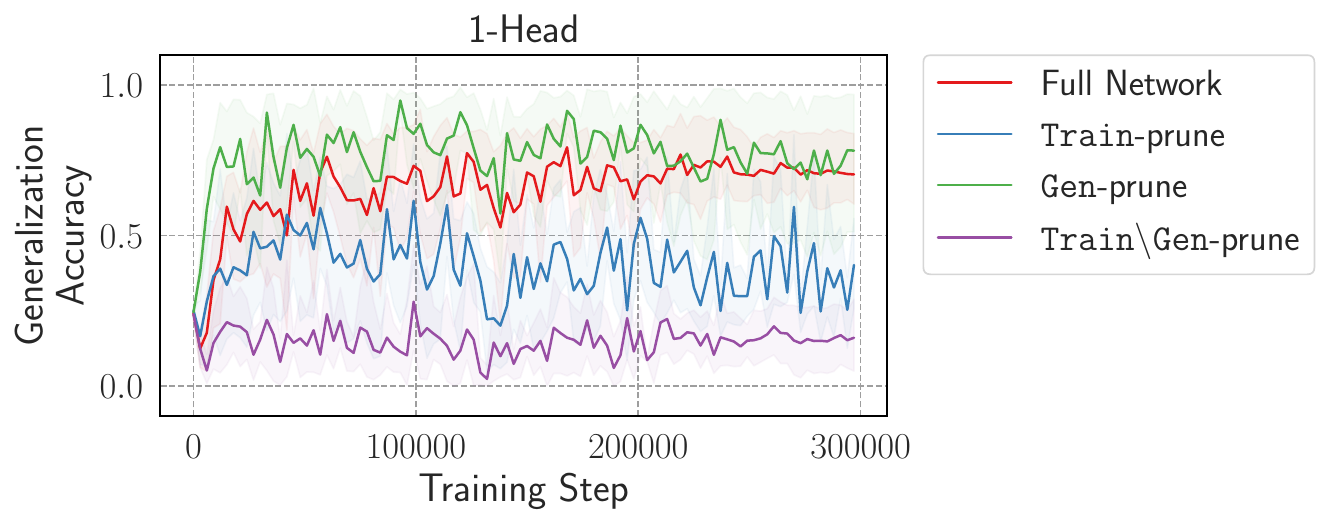}
        \caption{Training dynamics for Transformer LM with 1 head per attention layer}
        \label{fig:dynamics_1head}
    \end{subfigure}\hfill
    \caption{Effect of head capacity on the presence of hierarchical and linear generalization subnetworks. Runs are averaged over 5 seeds.}
    \label{fig:dynamics_heads}
\end{figure*}
\subsection{Grammar details}
\label{sec:grammar_appendix}

\paragraph{Local search using Bayesian model merging.}
The hand-constructed grammars that we consider in our study might not be optimal in terms of the posterior given the training data. E.g., there might be some redundant production rules or non-terminals, which can be removed by merging two or more non-terminals. We use the Bayesian Model Merging (BMM) algorithm from \citet{stolcke1994inducing, PERFORS2011306} to perform a local search for grammars with higher posteriors starting from the hand-constructed ones. The algorithm works as follows: We start from the initial grammar and iterate over all possible merges. A merge involves replacing two non-terminal symbol by a single new non-terminal and adding two new productions mapping the new non-terminal to the older ones. E.g. for production rules $A \to BC$, $A \to BD$, we can merge $C$ and $D$ to $F$, resulting the new production rules: $A \to BF$, $F \to C$, and $F \to D$. For each merge, we thus obtain a  new grammar, and compute its posterior. We then select the grammar with the highest posterior (greedy search) and repeat the procedure with this new grammar. If no merge results in a grammar with higher posterior than the initial grammar, we terminate the search. We denote the context free grammars after merge as $\cfg{}^{*}$ ($\cfgs^*$ and $\cfgl^{*}$) and regular grammars as $\reg{}^*$ ($\regs^*$ and $\regl^{*}$).


An important detail to note here is that while performing the merging algorithm, we use the ambiguous corpus $\dtrainl$ or $\dtrains$ for computing the posteriors and hence searching the right set of merges. The final grammar obtained, while should assign high likelihood to the ambiguous training data, it might no longer be consistent with the held out sentence types, e.g., $\dtestcfg$ or $\dtestreg$, and hence the final grammars obtained might not strictly model the linear or hierarchical rules. To check if such a situation arises in our case, we compare the set of all generations from a grammar before and after merging. If the two are same, it implies that the grammar continues to be consistent with both the ambiguous and unambiguous sentence types, and hence obey the linear or order rule of the original hand-constructed grammar. We find that for $\cfgs$, after applying the merging algorithm, the grammar obtained is no longer consistent with just the hierarchical rule and starts to also generate sentence types consistent with the linear rule. This implies that \textbf{for the low-diversity data case, even using a CFG it is better to avoid modeling the hierarchical rule given the ambiguous data}. For $\cfgl$, the grammar remains consistent with the hierarchical rule even after merging.

The \lprobs{} after applying BMM algorithm are provided in Table \ref{tab:posteriors_bmm}. For the $\dtrainl$ dataset, we find that our results remain consistent with those for hand-constructed grammars in Table \ref{tab:posteriors}: $\cfgl^*$ obtains a lower posterior than $\regl^{*}$. On the other hand for the $\dtrains$ dataset,  $\cfgs^*$ ends up with a higher posterior than the $\onest$ grammar. However, as noted above after minimization $\cfgs^*$ is no longer consistent with the hierarchical rule, i.e., doesn't generate sentences where verbs only agree with the hierarchically connected nouns. Hence, our observations that for the lower-diversity case, modeling the hierarchical rule is not optimal according the posterior criterion remains consistent here as well.



\begin{table}[!htbp]
    \centering
    \small
    \caption{Comparing the \lprobs for each of the 4 grammars after performing BMM on the \cfg{} and \reg{} grammars  given the training datasets $\dtrainl$ and $\dtrains$. The super-script $*$ symbol on \lpost{} for $\cfg^*$ on $\dtrains$ indicates that while the results show highest posterior for this grammar, after minimization the grammar no longer models the hierarchical rule.}
    \resizebox{0.95\textwidth}{!}{
    \begin{tabular}{lccc|ccc}
         \toprule
         \multirow{2}{*}{Grammar} & \multicolumn{3}{c|}{$\dtrainl$ (120 types)} & \multicolumn{3}{c}{$\dtrains$ (12 types)}  \\
         \cmidrule{2-7}
         & $\log$-Prior & $\log$-Likelihood & $\log$-Posterior & $\log$-Prior & $\log$-Likelihood & $\log$-Posterior\\
         \midrule
         \textbf{\cfg{}$^*$} & -345 & -639 & \textbf{-984} & -112 & -42 &\textbf{-155}$^*$ \\
         \textbf{\reg{}$^*$} & -393 & -658 & -1051 & -125 & -34 & -159 \\
         \textbf{\flt{}} & -4567 & \textbf{-574} & -5141 & -281 & \textbf{-30} & -311 \\
         \textbf{\onest{}} & \textbf{-58} & -2297 & -2355 & \textbf{-51} & -121 & -172\\
         \bottomrule
         
    \end{tabular}
    }
    \label{tab:posteriors_bmm}
\end{table}

\begin{figure}[!htbp]
    \centering
    \begin{subfigure}[t]{0.3\textwidth}
    \centering
    \includegraphics[width=0.99\textwidth]{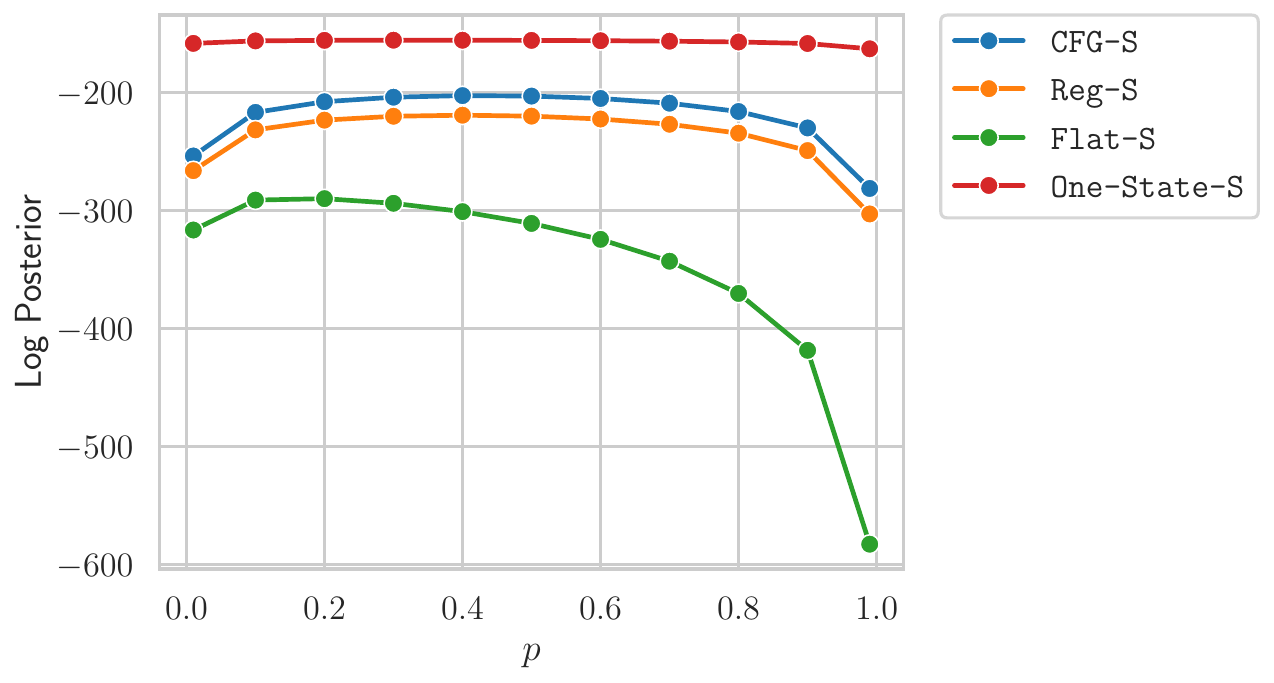}
    \caption{Low diversity data case -- $\dtrains$ for hand-constructed grammars}
    \end{subfigure}\hfill
    \begin{subfigure}[t]{0.3\textwidth}
    \centering
    \includegraphics[width=0.99\textwidth]{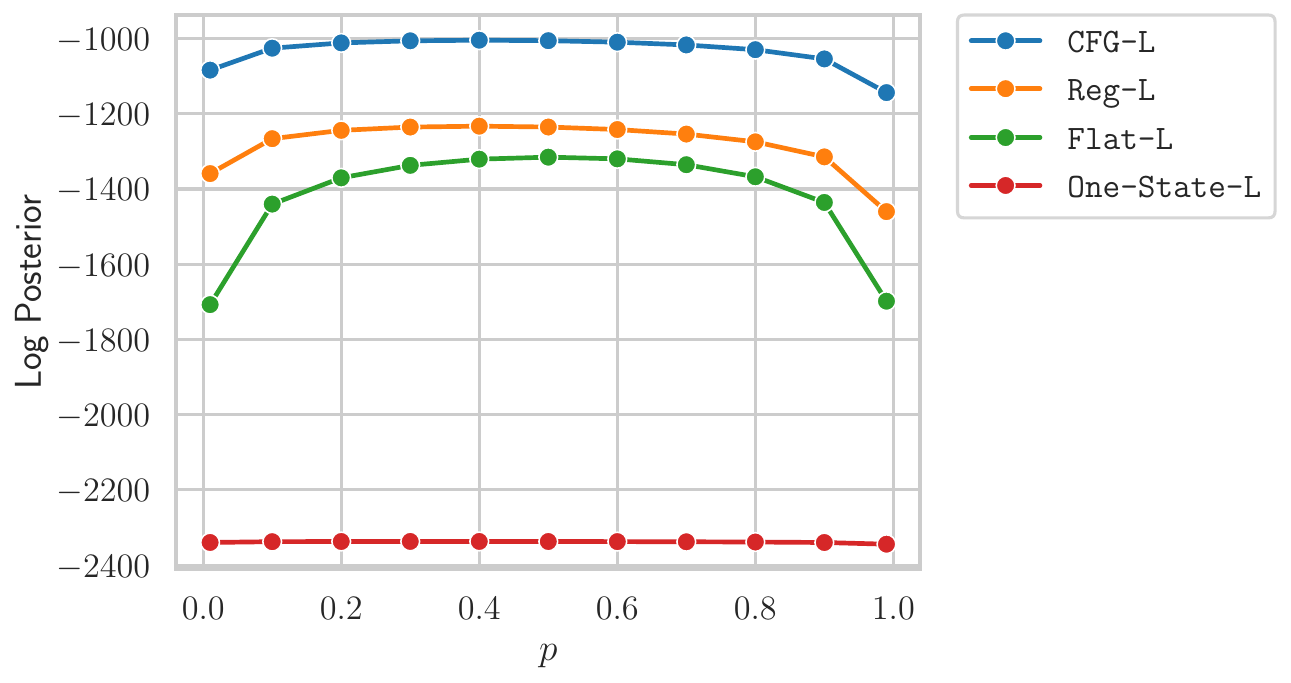}
    \caption{High diversity data case -- $\dtrainl$ for hand-constructed grammars}
    \end{subfigure}\hfill
    \begin{subfigure}[t]{0.3\textwidth}
    \centering
    \includegraphics[width=0.99\textwidth]{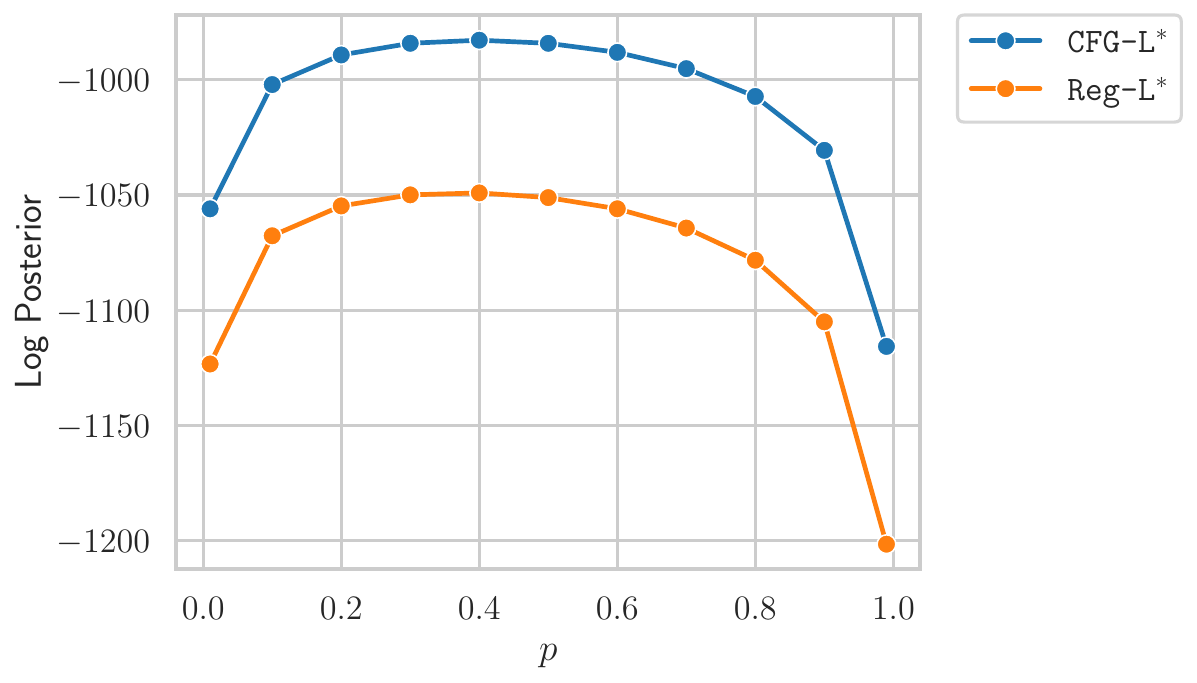}
    \caption{High diversity data case -- $\dtrainl$ for BMM minimized grammars}
    \end{subfigure}
    
    \caption{Sensitivity analysis on varying the geometric distribution parameter $p$. Note that the same $p$ is used for both $p(|V|)$ and $p(P_k)$ here. For the BMM minimized case (\textit{right}), we omit the Flat and one-state grammar to show the difference between the posteriors for $\cfgl^*$ and $\regl^*$ more clearly, as the other two grammars obtain much worse posteriors compared to the two (see the \textit{middle} figure)}.
    \label{fig:prior_senst}
\end{figure}

\begin{figure}[!htbp]
    \centering
    \begin{subfigure}[t]{0.33\textwidth}
    \centering
    \includegraphics[width=0.99\textwidth]{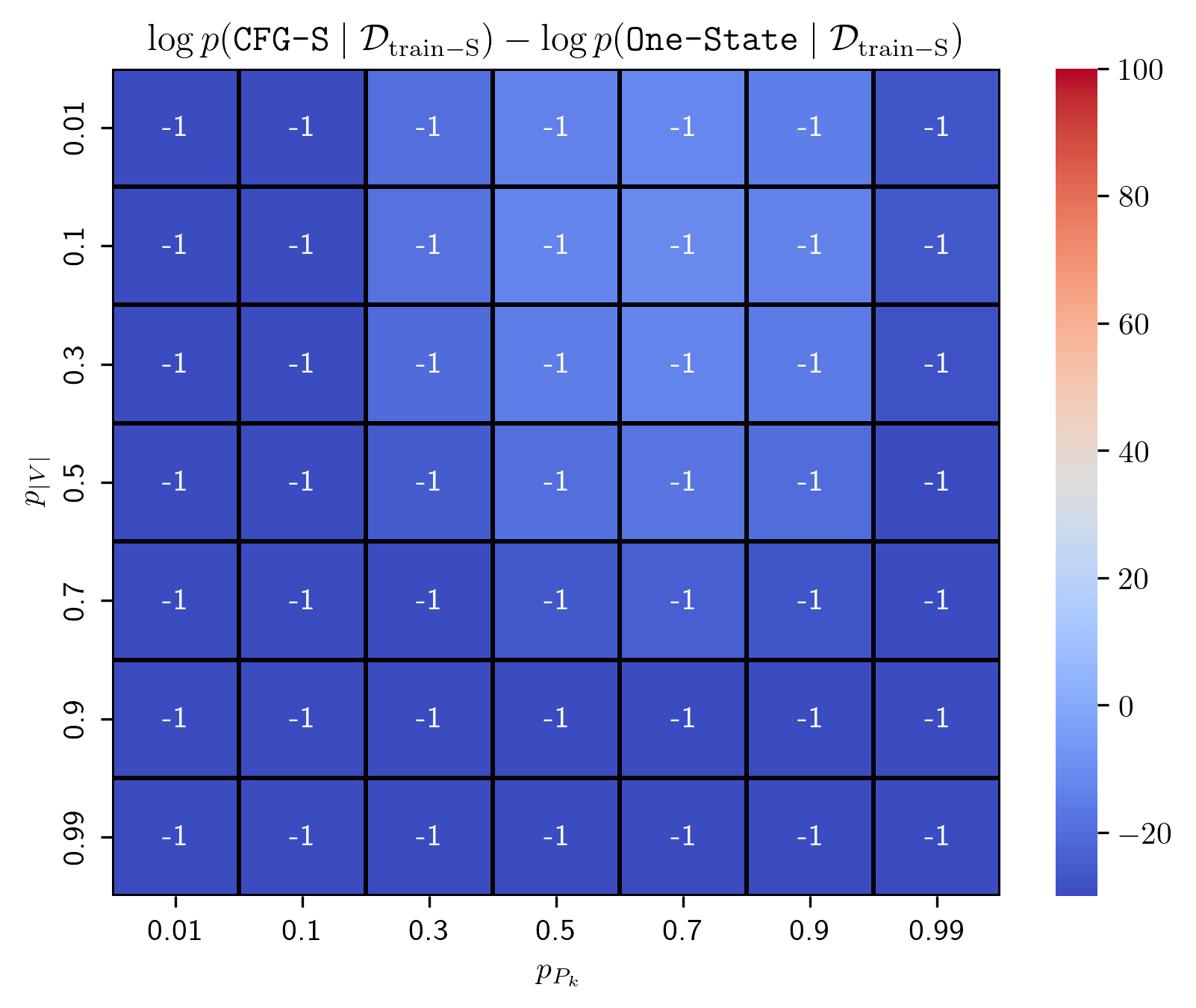}
    \caption{Low diversity data case -- $\dtrains$ for hand-constructed grammars.}
    \end{subfigure}\hfill
    \begin{subfigure}[t]{0.33\textwidth}
    \centering
    \includegraphics[width=0.99\textwidth]{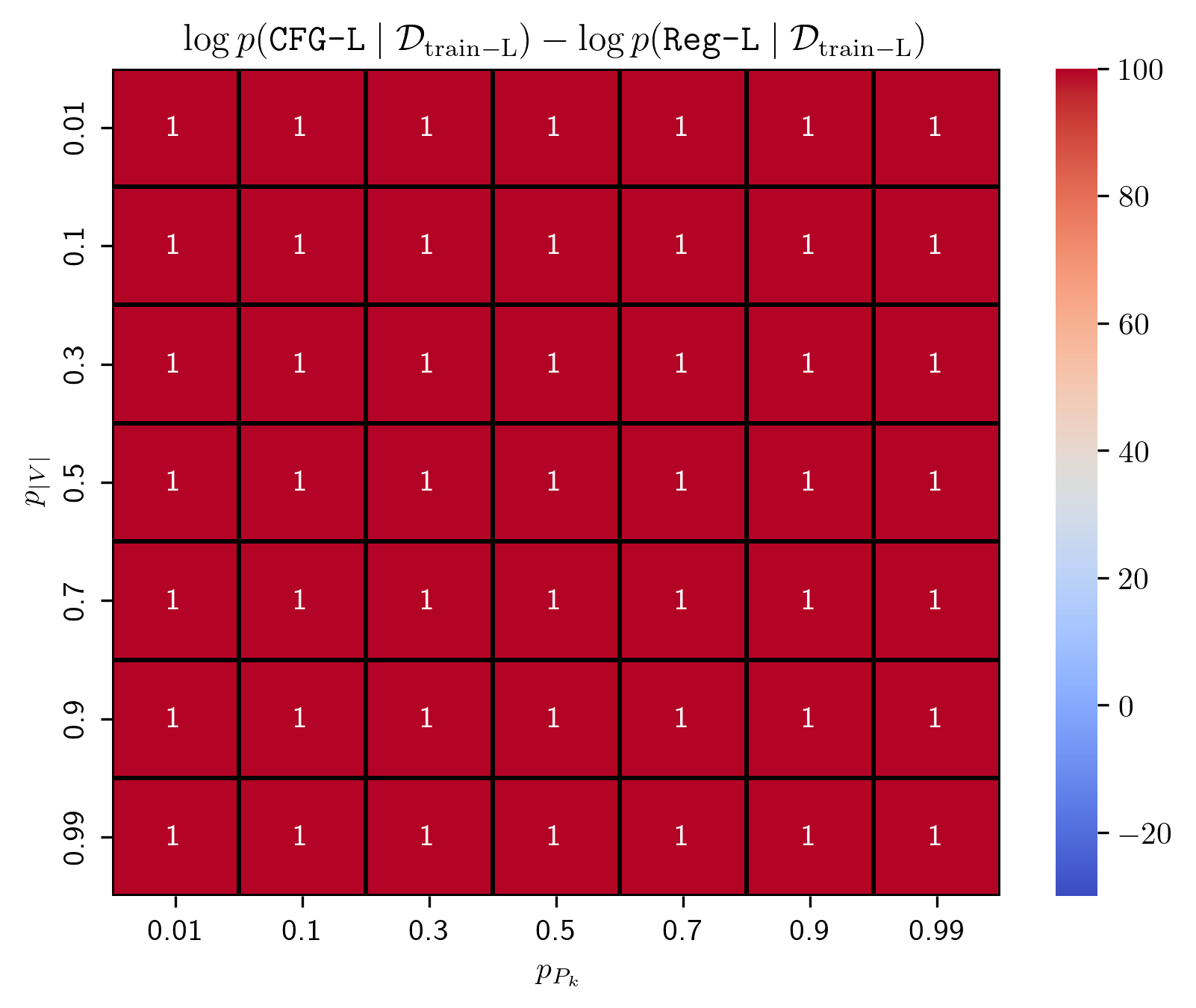}
    \caption{High diversity data case -- $\dtrainl$ for hand-constructed grammars.}
    \end{subfigure}\hfill
    \begin{subfigure}[t]{0.33\textwidth}
    \centering
    \includegraphics[width=0.99\textwidth]{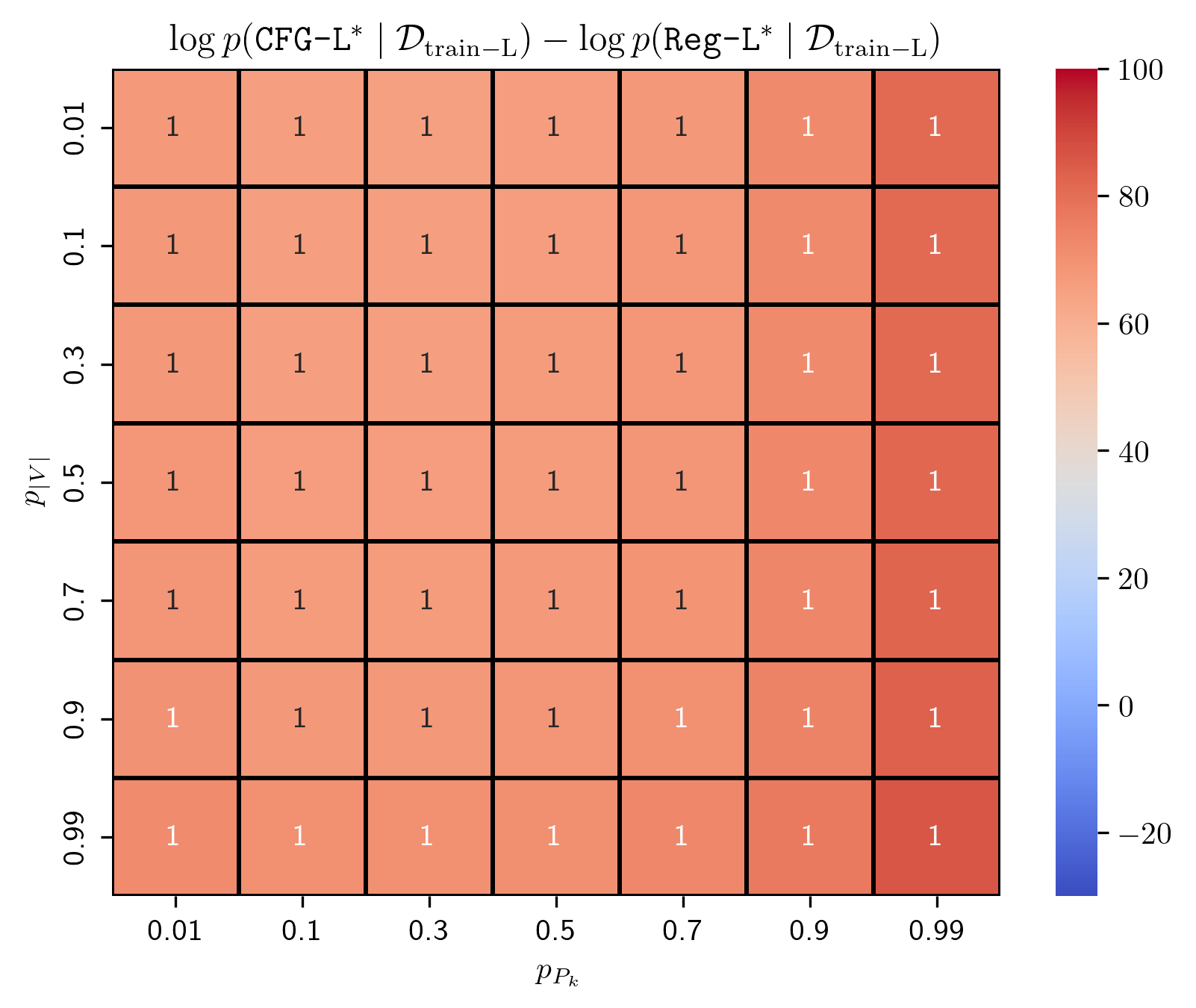}
    \caption{High diversity data case -- $\dtrainl$ for BMM minimized grammars.}
    \end{subfigure}

    \caption{Sensitivity analysis on varying the geometric distribution parameter $p_{|V|}$ for $p(|V|)$ and $p_{P_k}$ for $p(P_k)$. We plot the difference between the \lpost of the CFG and the other grammar with the highest posterior, which is $\onest{}$ for $\dtrains$ and $\regl$ (or $\regl^*$ for BMM minimized case) for $\dtrainl$. The values in the heatmaps correspond to the sign of the difference between the posteriors (1 for positive and -1 for negative). A positive sign implies that the CFG has the higher posterior than the alternate grammar and negative sign implies otherwise. }
    \label{fig:prior_senst_hm}
\end{figure}

\paragraph{Sensitivity Analysis.} We provide the plots for the sensitivity analysis o the choice of $p$ parameter of the geometric distributions used to define the prior in Figures \ref{fig:prior_senst} and \ref{fig:prior_senst_hm}. We provide the sensitivity analysis for both hand-constructed and BMM minimized grammars. For the latter case, we only provide the analysis on $\dtrainl$ dataset case, since after minimization on the smaller grammars ($\cfgs$ and $\regs$), we are left with no grammar obeying the hierarchical rule (see discussion in the previous paragraph for details).

\begin{figure}[!htbp]
    \centering
\begin{tikzpicture}[level distance=1.5cm,
  level 1/.style={sibling distance=4cm},
  level 2/.style={sibling distance=2cm}]
  \node {S}
    child {node {NP\_S}
      child {node {Det} 
        child {node {the}}}
      child {node {NPP\_S}
        child {node {N\_S} 
          child {node {\underline{\textbf{walrus}}}}}
        child {node {PP}
          child {node {Prep}
            child {node {near}}
          }
          child {node {N\_O}
            child {node {Det}
              child {node {my}}}
            child {node {N\_P}
              child {node {orangutans}}}}}}}
    child {node {VP\_S}
      child {node {V\_intrans\_s}
        child {node {\underline{\textbf{sings}}}}}};
\end{tikzpicture}
\caption{Example of production from the CFG with agreement following hierarchical rule}
\label{fig:cfg_prod_ex}
\end{figure}
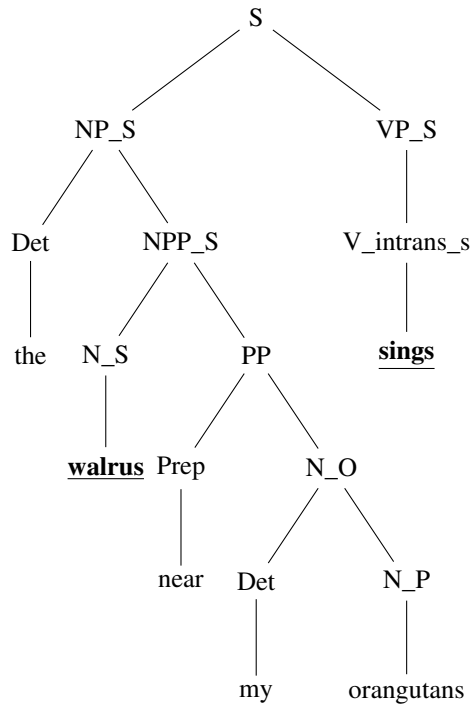

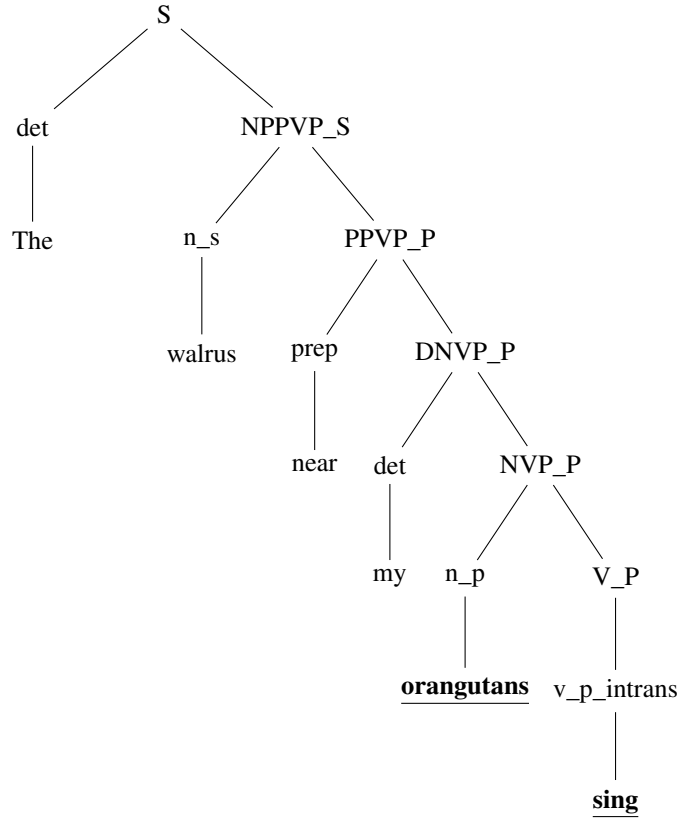
\begin{figure}[!htbp]
\centering
\begin{tikzpicture}[level distance=1.5cm,
  level 1/.style={sibling distance=3.5cm},
  level 2/.style={sibling distance=2.5cm},
  level 3/.style={sibling distance=2cm},
  level 4/.style={sibling distance=1.5cm}
  level 5/.style={sibling distance=1.5cm}
  level 6/.style={sibling distance=1.5cm}]
  \node {S}
    child {node {det}
        child {node {The}}}
    child {node {NPPVP\_S}
        child {node {n\_s}
            child {node {walrus}}}
        child {node {PPVP\_P}
            child {node {prep}
                child {node {near}}}
            child{node {DNVP\_P}
                child {node {det}
                    child {node {my}}}
                child {node {NVP\_P}
                    child {node {n\_p}
                        child {node {\underline{\textbf{orangutans}}}}}
                    child {node {V\_P}
                    child {node {v\_p\_intrans}
                        child {node {\underline{\textbf{sing}}}}}}}}}};
\end{tikzpicture}
\caption{Example of production from the linear grammar with agreement following order rule}
\label{fig:lg_prod_ex}
\end{figure}

\paragraph{Example parses}
Example parses from the CFG and regular grammars are provided in Figures \ref{fig:cfg_prod_ex} and \ref{fig:lg_prod_ex} respectively.

\begin{figure}[!htbp]
    \centering
    \begin{subfigure}[t]{0.33\textwidth}
    \centering
    \includegraphics[width=0.99\textwidth]{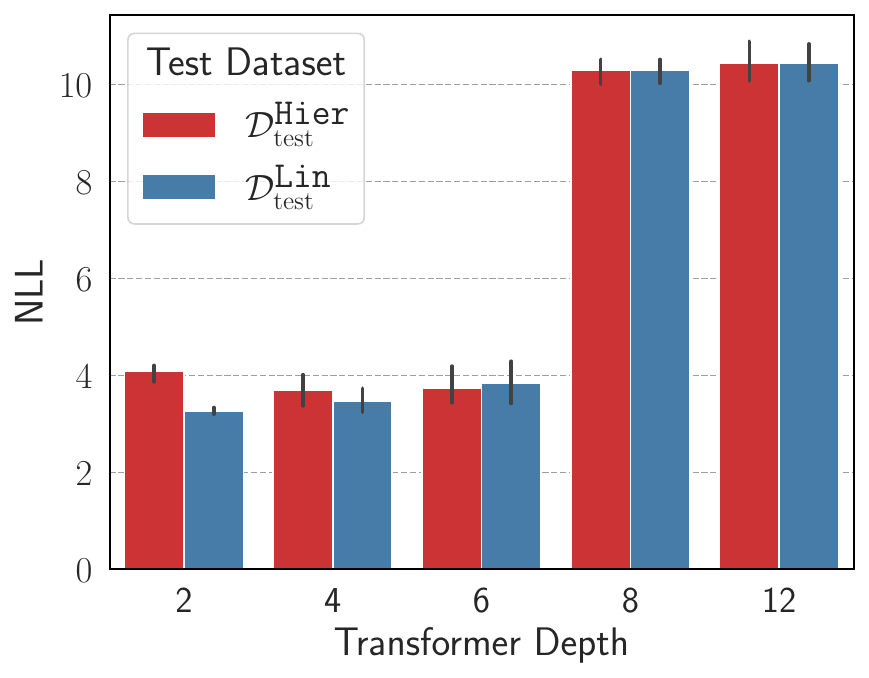}
    \caption{Negative \llike{} (NLL) on the $\dtestcfg$ and $\dtestreg$ test datasets.}
    \end{subfigure}\hfill
    \begin{subfigure}[t]{0.66\textwidth}
    \centering
    \includegraphics[width=0.99\textwidth]{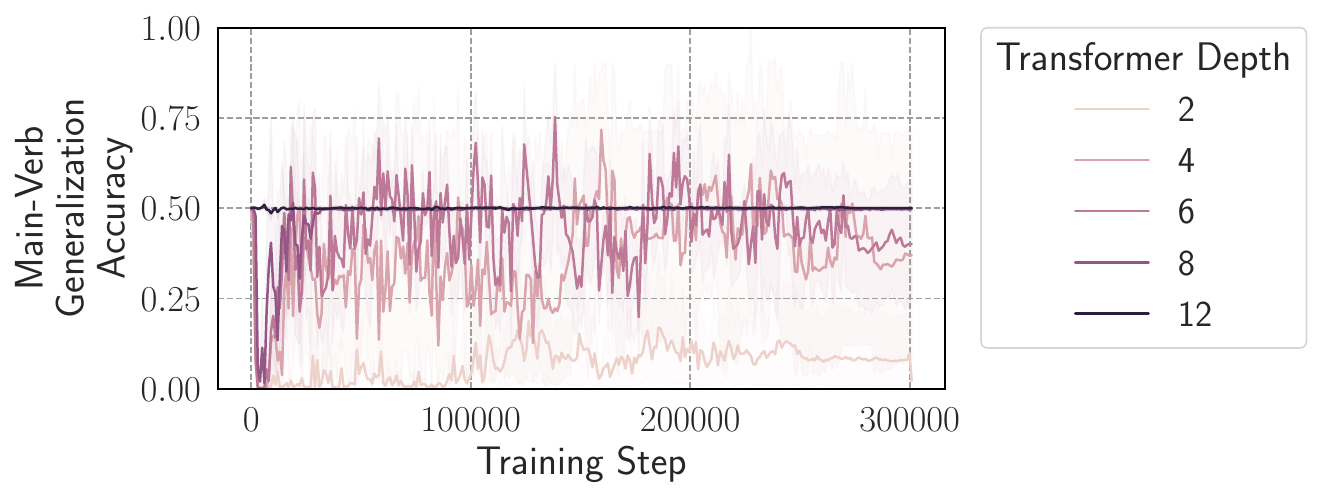}
    \caption{Main-verb generalization accuracy on $\dtestcfg$ test set.}
    \end{subfigure}
    \caption{Do transformer models trained on $\dtrains$ ever show hierarchical generalization? We vary the depth of the transformer-LM (number of decoder layers) and find in no case, transformer exhibiting hierarchical generalization. Interestingly, for smaller depths, we see the models generalizing according order rule, indicated by lower NLL on $\dtestreg$ than $\dtestcfg$ and a main-verb accuracy of roughly around 0\%/ when transformer depth is 2. For depths greater than 4, we observe starts to show no preference for either the linear or hierarchical rule.}
    \label{fig:bor_ld_diff_layers}
\end{figure}


\end{document}